\documentclass[preprint,12pt,authoryear,review]{elsarticle}

\usepackage{amssymb}
\usepackage{amsmath} 
\usepackage[dvipsnames]{xcolor}
\usepackage{dirtytalk}
\usepackage{float}
\usepackage{hyperref}
\usepackage{cleveref}
\usepackage{longtable} 
\usepackage{makecell} 

\begin{document}

\begin{frontmatter}


\title{From Words to Workflows: Automating Business Processes}

\affiliation[a]{
    organization={Novelis Research and Innovation Lab}, 
    address={40 av. des Terroirs de France},
    city={Paris},
    postcode={75012},
    country={France}
}

\author[a]{Laura Minkova}
\author[a]{Jessica López Espejel}
\author[a]{Taki Eddine Toufik Djaidja}
\author[a]{Walid Dahhane}
\author[a]{El Hassane Ettifouri}

\begin{abstract}

    As businesses increasingly rely on automation to streamline operations, the limitations of Robotic Process Automation (RPA) have become apparent, particularly its dependence on expert knowledge and inability to handle complex decision-making tasks. Recent advancements in Artificial Intelligence (AI), particularly Generative AI (GenAI) and Large Language Models (LLMs), have paved the way for Intelligent Automation (IA), which integrates cognitive capabilities to overcome the shortcomings of RPA. This paper introduces Text2Workflow, a novel method that automatically generates workflows from natural language user requests. Unlike traditional automation approaches, Text2Workflow offers a generalized solution for automating any business process, translating user inputs into a sequence of executable steps represented in JavaScript Object Notation (JSON) format. Leveraging the decision-making and instruction-following capabilities of LLMs, this method provides a scalable, adaptable framework that enables users to visualize and execute workflows with minimal manual intervention. This research outlines the Text2Workflow methodology and its broader implications for automating complex business processes.
    
\end{abstract}

\begin{graphicalabstract}
\begin{figure}[H]
    \centering
      \includegraphics[scale=0.45]{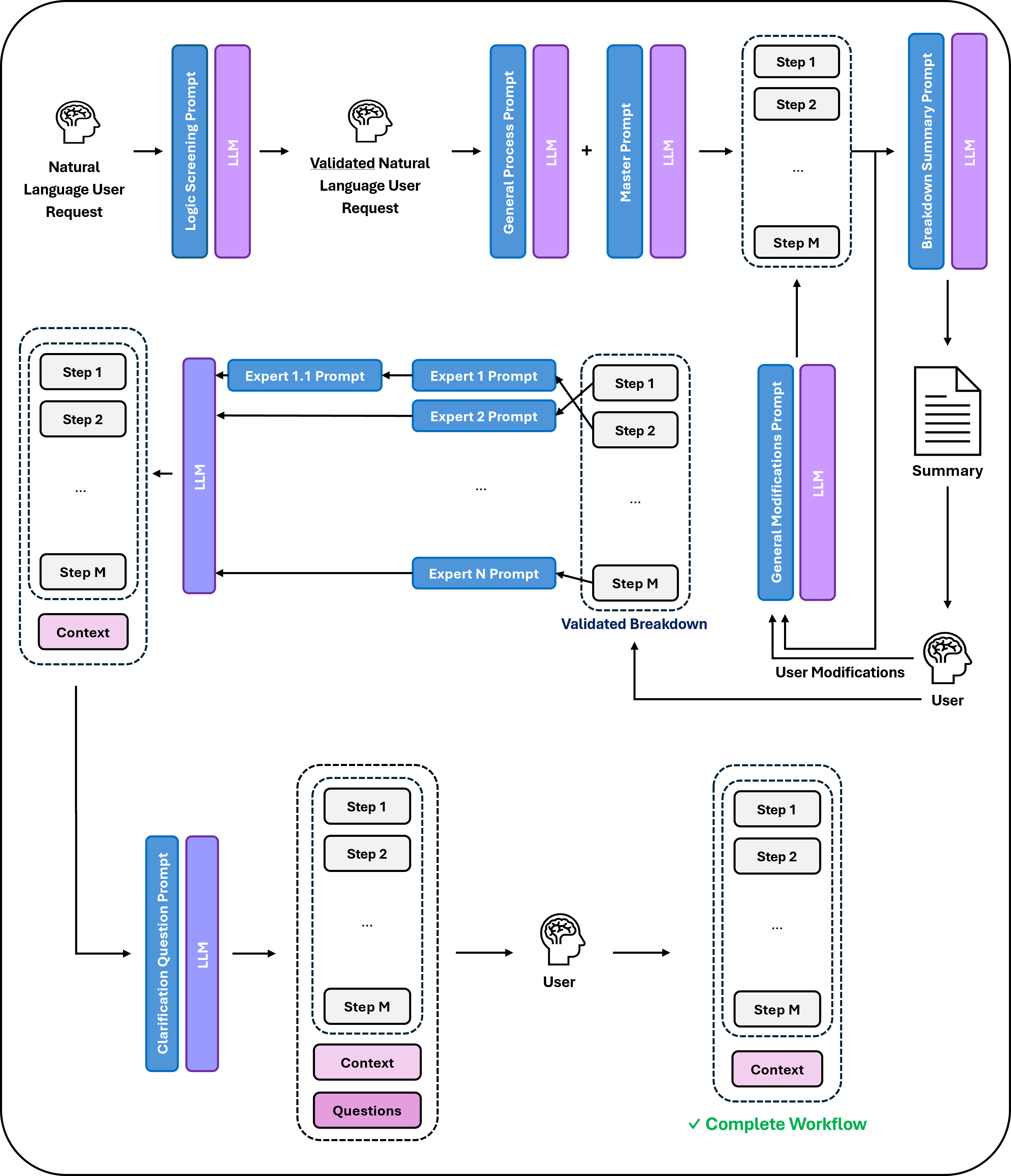}
        \label{fig:t2w-full-diagram}
\end{figure}

\end{graphicalabstract}

\begin{highlights}
    \item 
    We introduce Text2Workflow, a novel method for automating workflow generation. To our knowledge, no current open-source methods in the literature offer generalized automation across diverse business processes.
    \item 
    We propose Process2JSON, a dataset comprised of pairs of natural language user requests and corresponding workflows formatted in JavaScript Object Notation (JSON). It features a diverse array of user requests that span multiple tools and APIs with varying complexity, enabling a thorough evaluation of our approach's adaptability and performance.

    \item We conducted an ablation study to evaluate the impact of two user input mechanisms: (1) a general logic screening of the user request that the user must either accept or rewrite the request and (2) a user feedback loop inspired by the work done by \cite{FlowMind}.
        
    \item 
    We will publicly share our dataset and prompts to facilitate future research. 
    
\end{highlights}


\end{frontmatter}


\section{Introduction}
\label{sec:introduction}

    Robotic Process Automation (RPA) has long been a key enabler of business process automation, allowing organizations to streamline repetitive tasks and enhance operational efficiency. However, despite its widespread adoption, RPA has notable limitations. In particular, its reliance on expert knowledge, both in coding and business processes, makes its implementation and maintenance costly and complex \citep{Moreira2023_RPA, eulerich2024_darkRPA, FlowMind}. Additionally, RPA's cognitive capabilities fall short when handling tasks that require more advanced decision-making, leading to failure when faced with unforeseen circumstances or errors. As businesses evolve, these limitations underscore the need for more intelligent and adaptable automation solutions.

    The emergence of Artificial Intelligence (AI) has opened new avenues for overcoming the challenges inherent in RPA systems, leading to the emergence of systems known as Intelligent Process Automation (IPA). By integrating advanced AI technologies, including Machine Learning (ML) and  Generative Artificial Intelligence (GenAI), IPA extends automation beyond rule-based processes to encompass tasks requiring human-like cognitive abilities \citep{SiderskaAunimo}. Recent developments in GenAI, particularly with the advent of Large Language Models (LLMs) such as GPT (Generative Pre-trained Transformer)~\citep{gpt_1,gpt_2,GPT3-5}, have revolutionized the way automation systems can understand and generate natural language outputs, further expanding the scope of tasks that can be automated \citep{AdvancementsGAI}. These advancements have laid the groundwork for more dynamic and flexible automation frameworks capable of handling a broader range of business processes.

    While much research has been done on automating specific tasks through AI and LLM-based systems, most of this work has focused on physical robots and action planning for embedded AI, leaving gaps in the realm of generalized business process automation. However, there have been promising developments aimed at leveraging LLMs for workflow automation. For instance, \cite{ProcessGPT} introduced ProcessGPT, a fine-tuned GPT model trained on large datasets of business process information. This model has shown potential for automating repetitive business tasks by generating process flows from natural language inputs. More recently, frameworks such as FlowMind \citep{FlowMind} have combined the strengths of GPT-3.5 \citep{GPT3-5} with structured prompts to further enhance workflow generation.

    In this paper, we propose a novel approach, Text2Workflow, which automates the generation of workflows from natural language user requests. Unlike previous methods, which are often domain-specific or rely on predefined datasets, our method provides a generalized solution for converting user inputs into executable workflows across a wide range of business processes. We have chosen to represent our workflows in JSON format for its: (1) ease of readability, (2) efficient and lightweight format, and (3) wide usability. In addition, the JSON output of Text2Workflow enables seamless visualization and modification offering a scalable and adaptable tool for IPA and whose structure will, ultimately, serve as the blueprint for executing the business process. The key contributions of this paper are as follows:

    \begin{itemize}
        \item 
        We introduce a novel approach to automate workflow generation, called Text2Workflow. To the best of our knowledge, no existing open-source methods in the literature address the generalized automation of workflows across a wide variety of business processes.
        \item 
        We propose a dataset named Process2JSON, which includes a variety of user requests that cover a wide range of tools and APIs, varying in complexity to effectively assess the adaptability and performance of our approach.
        \item 
        We conducted an ablation study to evaluate the impact of two user input mechanisms: a general logic screening of the user request that the user must either accept or rewrite the request, and a user feedback loop inspired by the work done by \cite{FlowMind}.
        \item 
        We will publicly share our dataset and prompts to facilitate future research.
    \end{itemize}

    To date, there are no other open-source methods in the literature that tackle the automatic generation of any business process. There are, however, certain proprietary, closed-sourced products such as UiPath's AutoPilot \citep{uipath_autopilot}, and Microsoft's Copilot Studio \citep{ms_copilot_studio}.

    In the following sections, we will detail the methodology of Text2Workflow, discussing its structure, capabilities, and the potential for broad application in automating complex business processes.

\section{Related works}
\label{sec:related_works}

    In this section, we review related work in the fields of Robotic Process Automation, Generative Artificial Intelligence for Text Generation—highlighting developments from both the pre- and post-Transformers era—and workflow automation.

\subsection{Robotic Process Automation}
\label{subsec:RPA}

    RPA is a collection of technological tools designed to automate business processes by executing code that operates independently on software systems \citep{vanDerAalst2018Robotic,costa2022_RPA,Moreira2023_RPA}. Developing RPA solutions requires close collaboration between coding experts and business process experts~\citep{costa2022_RPA}, in order to create rule-based automation systems capable of operating with minimal human intervention. Many companies have turned to adopting such solutions or have outsourced their implementation to specialized firms such as Blue Prism~\footnote{\url{https://www.blueprism.com/}} and UiPath~\footnote{\url{https://www.uipath.com/}} \citep{vanDerAalst2018Robotic}.
    
    RPA offers several key advantages that help businesses remain competitive in a rapidly evolving landscape \citep{SiderskaAunimo,RPA_GAI}. First, RPA improves operations by accelerating processes and reducing human errors, leading to increased efficiency. Second, it is highly scalable, with RPA-equipped bots capable of handling heavy workloads without necessarily an increase in resources.  Additionally, as highlighted by \cite{RPA_GAI}, significant cost savings are achieved through reduced labor expenses and increased productivity. Perhaps most importantly, the time saved by RPA allows employees to focus on more strategic and creative tasks, leveraging their unique reasoning and cognitive skills to drive innovation and value within the organization.

    However, RPA also has notable limitations that must be considered \citep{eulerich2024_darkRPA,agostinelli2024_limitationsRPA}. For instance, its cognitive capabilities are inadequate for complex decision-making tasks that require human-like understanding. In situations involving unexpected errors or challenges, RPA agents can become immobilized, unable to determine the appropriate next steps. Moreover, RPA's reliance on expert knowledge in both coding and business processes is crucial for its implementation and ongoing maintenance, particularly as processes evolve \citep{Coombs2020_strategicIA,wewerka2023robotic,FlowMind}. This dependence can result in maintenance costs that exceed initial projections.

    In response to these challenges, recent advancements in AI and GenAI have given rise to a new domain known as Intelligent Process Automation (IPA) \citep{yakovenko2023intelligent,qordia2024_rpabenefits}. IPA enhances RPA by integrating AI-driven decision-making, machine learning, and data analytics, enabling more effective automation of a broader range of tasks, including those that require complex cognitive abilities \citep{SiderskaAunimo}. This evolution represents a significant step forward in overcoming some of the traditional limitations associated with RPA, ultimately leading to more sophisticated and capable automation solutions.

\subsection{Generative Artificial Intelligence for Text Generation}\label{subsec:GAI for Text and JSON}

    \cite{AdvancementsGAI} examines recent advancements in GenAI, highlighting the surge of interest in this field, especially following the release of OpenAI’s ChatGPT  \citep{chatGPT}. GenAI focuses on creating new data outputs, such as text, images, and audio, by generating rather than merely analyzing data. The launch of ChatGPT catalyzed rapid development in LLMs, inspiring other companies and research institutions—including Google’s Gemini \citep{GooglePalm,GoogleGemini}, Meta’s LLaMA \citep{MetaLlama,touvron2023_llama2,roziere2024_codeLLaMA}, and Anthropic’s Claude \citep{claude3}—to release their own LLMs, spanning both open-source and proprietary offerings. However, the field of GenAI has roots that extend back decades, with steady progress that accelerated with the introduction of the transformer~\citep{AttentionIsAllYouNeed} architecture. To contextualize these developments, we organize this subsection into two periods: pre-transformer and post-transformer.

    \subsubsection{Pre-Transformers Era}
    
    The field of generative artificial intelligence began long before the recent emergence of transformers, GPTs~\citep{gpt_1, gpt_2,few-shot-learning}, and other LLMs \citep{gordijn2023chatgpt,minaee2024large,AdvancementsGAI}. A significant early contribution to this field was the development of Recurrent Neural Networks (RNNs) by \cite{RNNs}, which opened up numerous applications for treating sequential data, including natural language processing (NLP)~\citep{farooq2023multi,kang2023bilingual} tasks and time series analysis~\citep{yu2021analysis,wang2022ngcu}. Subsequently, \cite{LSTM} introduced the Long Short-Term Memory (LSTM) model, addressing a major limitation of RNNs by enabling the processing of longer and more complex sequences. Similarly, \cite{GRU} introduced the Gated Recurrent Unit (GRU), which not only improved upon the RNN architecture but also enhanced \cite{LSTM} work by offering a simpler and more resource-efficient model.

    Around the same time as GRU, models more traditionally associated with GenAI began to emerge. For instance, \cite{GANs} presented Generative Adversarial Networks (GANs) where two models - a generator and a discriminator - compete to produce the most convincing phony outputs. In a similar vein, \cite{VAEs} proposed Variational Auto-Encoders (VAEs), a variation of the autoencoder model that learns to encode input data and then decode it to regenerate the original input.

    \subsubsection{Post-Transformers Era}

    The field of GenAI took a turn with the groundbreaking advancements introduced by the transformer architecture \citep{AttentionIsAllYouNeed}. The transformer model revolutionized NLP by leveraging self-attention mechanisms to improve efficiency and accuracy in handling sequential data \citep{planningabilitieslargelanguage}, leading to the eventual development of models such as BERT \citep{BERT} and more recently, GPT-4 \citep{openai2024gpt4technicalreport}. Other key advancements include diffusion models \citep{diffusion,peebles2023scalable,yang2024cogvideox}, which have achieved state-of-the-art performance in image generation, and hybrid models like Chinchilla~\citep{compute-opt-llm} that improve training efficiency and scalability. Additionally, methods such as Reinforcement Learning from Human Feedback (RLHF)~\citep{rlhf} have been introduced to align AI models with human intentions and improve the coherence of their generated outputs. These innovations have expanded the capabilities of GenAI, allowing it to tackle more complex tasks and opening doors to broader industrial and commercial applications.

    LLMs, which are based on the transformer architecture, have been central to the evolution of GenAI, particularly in NLP tasks. LLMs such as OpenAI’s GPT series \citep{gpt_1,gpt_2,few-shot-learning,GPT3-5,openai2024gpt4technicalreport}, Google’s PaLM \citep{GooglePalm}, and Meta’s LLaMA \citep{MetaLlama,touvron2023_llama2,dubey2024llama3herdmodels} have shown remarkable proficiency in generating human-like text, understanding complex instructions, and performing various tasks, from answering questions to composing coherent articles. These models excel at processing and generating text at scale while maintaining context over long sequences, revolutionizing tasks such as machine translation, summarizing, and conversational AI.

    LLMs are also capable of zero-shot and few-shot learning~\citep{few-shot-learning,kojima2023largelanguagemodelszeroshot, labrak2024zeroshotfewshotstudy}, allowing them to generalize to new tasks with minimal training. Their competencies now extend to reasoning~\citep{Bang_MultitaskChatgpt2023,espejel2023gpt}, problem-solving~\citep{liu2023improvinglargelanguagemodel,AlphaProof,AlphaGeometry}, and even coding~\citep{Li2022_AlphaCode, codegemmateam2024,pinnaparaju2024_stablecode}, broadening their applications beyond traditional NLP tasks.  However, despite these advancements, LLMs still encounter challenges such as hallucinations, ethical concerns, and biases in generated content, which continue to be active research areas~\citep{dangers-llm}. While their ability to handle long sequences has improved, their performance in complex, real-world scenarios remains limited~\citep{li2024long}. Nevertheless, with ongoing enhancements through  RLHF, LLMs are becoming more aligned with human expectations and practical needs~\citep{hou2024chatglmrlhfpracticesaligninglarge,banerjee2024reliablealignmentuncertaintyawarerlhf}. This progress is driving their adoption across various sectors, including customer service~\citep{pandya2023automating, shi2024chops}, content creation~\citep{moore2023empowering,leiker2023prototyping}, healthcare~\citep{mesko2023imperative,haltaufderheide2024ethics}, and legal analysis~\citep{nay2024large,cheong2024not}.

    Prompt engineering has also emerged as a critical aspect of optimizing LLMs~\citep{yang2024large,sahoo2024systematicsurveypromptengineering}, allowing users to guide the model's output in a more controlled and task-specific manner. Several well-established techniques have been developed to improve the quality and relevance of LLM responses. Few-shot prompting \citep{few-shot-learning}, for example, provides the model with a few examples of the desired output format before generating its own response, thus enhancing performance on tasks with minimal training data. Chain-of-Thought (CoT) prompting~\citep{CoT} encourages the model to generate intermediate reasoning steps, leading to more accurate results in tasks that require logical and mathematical reasoning. Furthermore, instruction-based prompting~\citep{feng2023sentencesimplificationLM,juseon2024instructcmp} explicitly defines the task or question in the prompt, proving especially effective for models fine-tuned with RLHF to follow complex instructions. These techniques not only improve output accuracy but also help reduce issues like ambiguity and irrelevant responses, making them essential tools for optimizing LLM performance across diverse applications.

    \subsection{Workflow Automation}
    \label{subsec:Workflow_Automation}

    The existing literature on the automation of generating workflows in JSON format from natural language requests remains limited. However, several relevant research developments address adjacent topics, including action planning and scheduling, code generation, JSON generation, and workflow generation. These areas provide a foundational context for exploring the automation of workflow creation from natural language inputs.

    Significant research has been conducted in the domain of action planning and scheduling~\citep{jokic2023framework,arora2024anticipate}. Much of this work focuses on physical robots and embodied artificial intelligence, as demonstrated by studies such as those by \cite{genRobots, SmartLLM, AnticipateAct, DoAsICanNotAsISay, ChatGPT4Robotics}.  Conversely, other research examines the planning capabilities of LLMs more broadly. For instance,~\cite{PlanningAbilitiesCriticalInvest} investigate the reasoning capabilities of LLMs in planning commonsense tasks, concluding that these capabilities remain limited; one of the most reputable models at the time, GPT-4 \citep{openai2024gpt4technicalreport}, achieved only a 12\% success rate on tasks, with some averages as low as 3\%. The study by \cite{TPTU} established a structured framework for evaluating the reasoning capabilities of LLMs in Task Planning and Tool Usage (TPTU). Additional frameworks, such as AgentGen \citep{AgentGen}, Text2Reaction \citep{Text2Reaction}, and ISR-LLM~\citep{ISR-LLM}, have also been developed to enhance the reasoning abilities of LLMs by leveraging their existing capabilities.

    Since workflow automation inherently involves the execution of code, the generation of code through artificial intelligence is particularly relevant to this research. An important work in the development of code generation via language models (LMs) and LLMs was done by \cite{Codex}. They produced a 12B parameter fine-tuned version of GPT on publicly available Python code on Github. Remarkably, Codex was able to solve 28\% of the problems from their HumanEval dataset, as opposed to GPT-3's 0\% success rate. Subsequent advancements include the introduction of the CodeGen family of models \citep{CodeGen}, which competes with state-of-the-art Python code generation capabilities on HumanEval.  This work also explores multi-turn code generation, allowing users to iteratively explain the code they wish to have generated. Moreover, substantial research has been dedicated to evaluating the code produced by these models. For example, \cite{codeReallyCorrect} introduced a framework called EvalPlus to rigorously assess the correctness of generated code and enhanced the original HumanEval dataset's test cases by eighty-fold to create HumanEval+, aiming to improve future code generation models. Together, these developments in planning, reasoning, and code generation highlight the intricate relationship between LLM capabilities and the automation of complex workflows.

    Ensuring that a LLM outputs consistently outputs the same structured format is highly desirable and has significant implications for automating business processes.  One of the most commonly used structured formats, JSON, is particularly relevant to Text2Workflow. OpenAI has previously released specific JSON response formats~\citep{GPTStructuredFormats} for both  GPT-3.5 and GPT-4. More recently, with the introduction of GPT-4o~\citep{GPT4o}, they have implemented a feature that facilitates the generation of custom structured data, which helps prevent the LLM from producing JSON outputs with missing keys or hallucinations \citep{GPTStructuredFormats}. Additionally, research by \cite{SynCode} presents a general framework called SynCode that efficiently constraints LLM output tokens to adhere to any defined context-free grammar. This approach offers the advantage of generating various types of structured data, particularly JSON structures, across different LLM models and tasks, thereby enhancing the flexibility and reliability of automated workflows.

    Lastly, considerable research has been conducted in the field of workflow automation. One notable advancement is the introduction of ProcessGPT~\citep{ProcessGPT}, a variant of the GPT model that has been fine-tuned on extensive datasets related to business processes. This model leverages GPT's robust natural language understanding and generation capabilities, enabling it to identify opportunities for business process improvements and automate repetitive tasks by generating process flows. More recently, \cite{FlowMind} introduced FlowMind,  a framework that leverages GPT's strengths with a structured prompt recipe comprising three key elements: context, a description of various tools, and a request for code generation. By integrating GPT with this specific prompt recipe and a human-in-the-loop system, FlowMind achieves remarkable accuracy in generating workflows described in Python code, approaching nearly 100\% precision. These developments represent significant strides toward enhancing the efficiency and effectiveness of workflow automation.

\section{Methodology}
\label{sec:methodology}

\subsection{JSON-Based Structure in Text2Workflow}
\label{sub:JSON-Based_Structured_Text2Workflow}

    The goal of Text2Workflow is to automatically transform a natural language user request into a structured workflow of actionable steps, represented in JSON format, with the ultimate aim of autonomously executing the entire process. In the following subsection, we outline the methodology of our approach. We begin by describing the structure of the output, followed by an explanation of how the proposed method generates this output.

\begin{figure}[H]
        \center
        \includegraphics[width=0.8\textwidth, height=0.8\textheight, keepaspectratio]{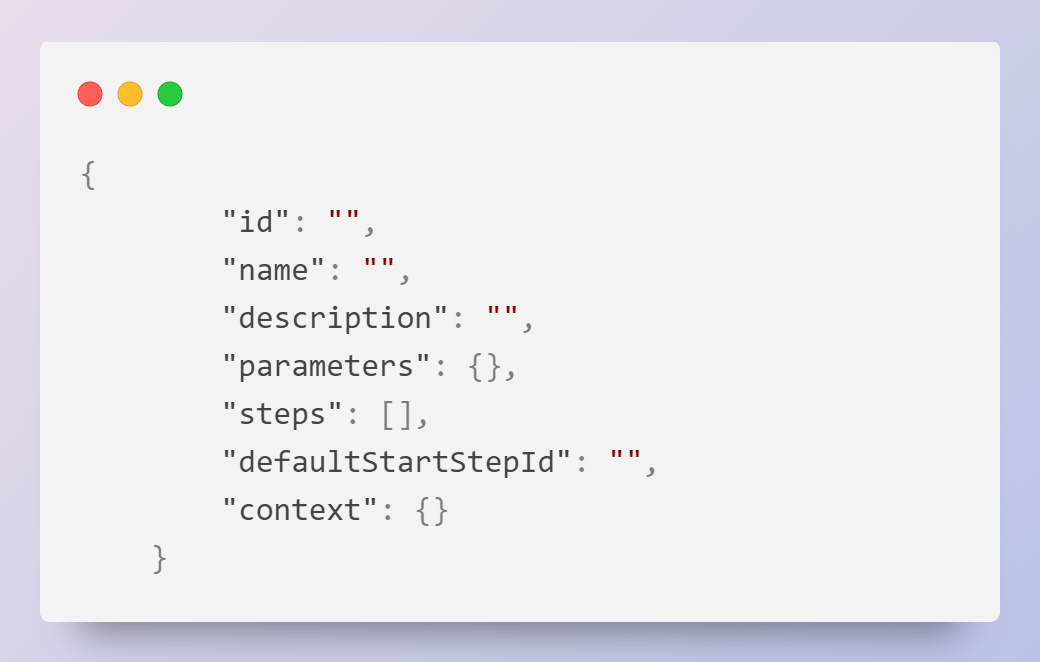}
            \caption{General Process JSON.}
        \label{fig:generalProcessJSON}
    \end{figure}

    We predefined a JSON output structure, designed with generalizability in mind. Specifically, the format was developed to accommodate a wide variety of business processes, ensuring adaptability across different applications. The JSON workflow is divided into two components: a general process component and a step-specific component. The general process component contains keys that store metadata and overarching information about the entire process. The structure of this component is shown in Figure~\ref{fig:generalProcessJSON}.

    The \textit{id} is a process-unique, random universal unique identifier (UUID). The \textit{name} key stores a short, couple-word name that summarizes the process. The \textit{description} is a longer, one to two sentence long summary of the process. The \textit{parameters} key stores process unique parameters, though it is often left empty. The \textit{steps} key stores a list of individual, unitary steps required to accomplish the process. The \textit{defaultStartStepId} stores the ID of the first step in the \textit{steps} list to begin the execution by. Finally, the \textit{context} is a dictionary structure that stores the names of the variables used throughout the workflow, wherein every key is a variable name and its value is a dictionary with keys: type, value, and description.

    The structure of each unitary step in the \textit{steps} section of the workflow is determined by the step's \textit{type}. Currently, the workflow includes the following defined step types: Decision, Loop, Calculation, DataExtraction, API-type (which covers interactions with services such as Outlook, Excel, File, Web, and Desktop), Exception, and Unknown. Each type has a unique JSON structure, outlined in \ref{app:json-struct}.

    The step-specific component, stored under the \textit{steps} key, outlines the individual actions within the workflow. This component contrasts with the overarching process component, which manages general metadata for the workflow. Together, they enable a flexible and detailed execution framework that can be adapted to various business processes and contexts.
    
    \subsection{Enhancing Generation in Text2Workflow through Master and Experts}
    
    To construct the aforementioned JSON structures, Text2Workflow can be broken down into seven distinct layers of prompts that we detail in this section:
    
    \begin{enumerate}
        \item \textbf{User Request Ambiguity Screening}. User Request Screening Prompt.
        \item \textbf{Building the Workflow Skeleton}. General Process Prompt \& Master Prompt.
        \item \textbf{Human Feedback Loop}. High-level Summary Prompt \& Workflow Modification Prompt.
        \item \textbf{Workflow Details}. Expert Prompts.
        \item \textbf{*Special Case* Further Workflow Details}. Parameter Expert Prompt.
        \item \textbf{Verifying Missing Parameters}. Questions Prompt.
        \item \textbf{Final Modifications}. Workflow Modification Prompt.
    \end{enumerate}

    The remainder of this subsection provides a detailed analysis of each layer within Text2Workflow. The prompts used in each layer are listed in~\ref{app:prompts}. For a visual overview of how each prompt layer functions within our approach, please refer to \Cref{subsec:mechanisms}.

\subsubsection{User Request Clarification}
\label{subsub:user_request_clarification}

    The first step in Text2Workflow involves verifying that the user request is, at a minimum, not blatantly missing any information, logically structured, and relevant to business processes. This is achieved through the \textit{User Request Screening Prompt} (see \Cref{fig:logic-screening-prompt}), which either returns an empty response if the request is valid or provides follow-up questions if clarification is needed. The user can then choose to re-write their request to clarify specific points raised by the Text2Workflow agent or ignore the suggestions if preferred. This initial layer serves as a preliminary check, alerting the user to potential issues without blocking further progress in Text2Workflow.

\subsubsection{Building the Workflow Skeleton}
\label{subsub:building_workflow_skeleton}

    We define a workflow skeleton as a nearly complete version of the business process' workflow JSON, lacking only the specific parameters for each step. In order to construct this skeleton, we concatenate the results from two prompts: the \textit{General Process Prompt} (see \Cref{fig:gen-process-prompt}), and the \textit{Master Prompt} (see \Cref{fig:master-prompt-1,fig:master-prompt-2,fig:master-prompt-3,fig:master-prompt-4,fig:master-prompt-5}).

    The \textit{General Process Prompt} generates values for each key in the general process JSON, except for the \textit{steps} key, as shown in \Cref{fig:generalProcessJSON}.

    The \textit{Master Prompt}, on the other hand,  is responsible for filling out the \textit{steps} key of the final workflow. Leveraging its understanding of various step types, it breaks down the business process into unitary steps, classifies each step into an appropriate type whereby it assigns it to the \textit{type} key, and provides each step with an \textit{id}, \textit{name}, and \textit{description}. The steps are then organized in chronological order according to the original user request by assigning a \textit{nextStepId} value to each step, ensuring a single subsequent ID except in Decision steps, where the next step is contingent on a certain condition.

\subsubsection{User Feedback Loop}

    The workflow skeleton, as described in \Cref{subsub:building_workflow_skeleton}, is subsequently reviewed by a human user. This approach is inspired by the findings of \cite{FlowMind}, which demonstrated that incorporating a human-in-the-loop mechanism for confirming or editing a workflow resulted in an accuracy improvement of over 6.5\% in more complex scenarios.

    Initially, the workflow skeleton is summarized using the \textit{High-level Summary Prompt} (see \Cref{fig:brkdwn-summary}). This prompt provides a natural language summary of the skeleton without presuming any coding knowledge of the user that will be analyzing it. In other words, it articulates each step as presented in the \textit{steps} key of the workflow.

    Afterwards, the summary is then presented to the user, upon which they have one of two options: approve the summary and move on to the next stage of Text2Workflow, or suggest edits to be made to the workflow.

    In the latter case, the \textit{Workflow Modification Prompt} takes the previous workflow skeleton along with the user’s edits and modifies the workflow accordingly. This revised workflow is then reintroduced to the \textit{High-level Summary Prompt} for the user to review the modifications. This iterative process continues indefinitely until the user validates the workflow.

\subsubsection{Workflow Details}

    Once the workflow has been validated by the user, Text2Workflow iteratively traverses the \textit{steps}. For each step, a specific \textit{Expert Prompt} is invoked based on the step's \textit{type}. For a comprehensive list of the \textit{Expert Prompts}, please refer to \Cref{fig:api-function-prompt,fig:writein-params,fig:click-select-params,fig:calculation-params-1,fig:calculation-params-2,fig:data-extract-params-1,fig:data-extract-params-2,fig:loop-params-1,fig:loop-params-2,fig:trycatch-desc}. Each \textit{Expert Prompt} returns the correct parameters, with the exception of step types that require an API. In such cases, the \textit{Expert Prompt }is a generic prompt that simply returns the necessary function based on the tool being used—such as Outlook, Excel, File, Web, or Desktop.

    The \textit{Expert Prompt} is also responsible for creating and keeping track of the workflow's \textit{context}, which is a dictionary that tracks all of the variables throughout the workflow's execution. This context includes variables that store extracted or read text, calculated entities, data tables, and more. The \textit{Expert Prompt} populates each step's parameters, which may involve using previously instantiated variables or updating the context with a newly instantiated variable.

\subsubsection{Special Case -- Further Workflow Details}
    
    This applies to steps requiring additional parameter details, such as steps of type Outlook, Excel, File, Web, Desktop, or Exception (TryBlock). In such cases, an additional layer, called the \textit{Parameter Expert Prompt}, is invoked (\Cref{fig:api-params-1,fig:api-params-2,fig:tryblock-params,fig:throw-except-params}). This extra layer of a \textit{Parameter Expert Prompt} has nearly identical functionality to the previous layer, simply returning the parameters of certain steps, and updating the context if needs be.

\subsubsection{Verifying Missing Parameters}
\label{subsub:verifying_missing_parameters}

    Once the business process workflow is fully defined with all steps and their corresponding parameters, a final check is done to identify any missing essential parameters that may not have been filled out by the \textit{Expert Prompts}. Essential parameters are prerequisites for executing the API function and are predetermined. For example, in sending an email, the ``to'' and ``body'' fields are essential.  For every step, a list of missing essential parameters is manually created and is verified by checking for empty strings or null values. This list is then sent as input to the \textit{Questions Prompt} (see \Cref{fig:questions-prompt}), which  formulates a natural sounding question that will be displayed to the user. The user can then manually add the missing information they may have not provided initially or that the LLM may not have caught.

\subsubsection{Final Modifications}
\label{subsub:final_modifications}

    Finally, if the user chooses to make changes to the generated workflow after everything has been done, they can do so by providing certain modifications they would like to be made. These modifications, along with the complete, generated workflow are then provided as input to the  \textit{Workflow Modification Prompt} (see \Cref{fig:workflow-mod-prompt-1,fig:workflow-mod-prompt-2}). This prompt will handle any modifications requested. If a step has been added, or a previous step's parameters have been modified, the corresponding \textit{Expert Prompt} may be invoked to complete or adjust the parameters as needed.

\subsection{Prompt Engineering to Improve the Text2Workflow pipeline}
\label{sub:prompt_engineering}

    To design precise prompts, we applied prompt engineering research strategies, including few-shot prompting, where a small set of examples demonstrates the desired output. Following~\cite{chen2023_few1shot}, we found that a single example is often sufficient to guide the model effectively and produce favorable results. For this reason, nearly all Expert prompts have a single example. The only exceptions are the \textit{Loop Expert Prompt} with 2 examples (see \Cref{fig:loop-params-1,fig:loop-params-2}) and the \textit{Master Prompt} with 4 examples-- two in English and two in French (see \Cref{fig:master-prompt-1,fig:master-prompt-2,fig:master-prompt-3,fig:master-prompt-4,fig:master-prompt-5}). We felt the latter two prompts required more than one example given the complexity of the task at hand. Additionally, we applied the Chain-of-Thought~\citep{CoT} technique, prompting the model to break down complex tasks into logical steps to improve its responses on specific tasks. Finally, we used role-based prompting~\citep{zhao2021calibrateuseimprovingfewshot} to align responses from both gpt-3.5-0125 and gpt-4o-mini with the intended context and style.

    After designing the prompts, we found that the average baseline prompt length is as high as 9058 tokens. Given the context window capacities of gpt-3.5-0125 (16,385 tokens) and gpt-4o-mini (128K tokens), we hypothesize that gpt-4o-mini may perform better due to its larger window. However, existing studies~\citep{li2024long,hosseini2024efficient} suggest that an extended context window does not inherently ensure accurate comprehension of lengthy sequences. Specifically,~\cite{hosseini2024efficient} emphasizes that current LMs still face challenges in reasoning and understanding within long contexts.

    Based on prompt engineering techniques and insights about long context windows, we developed the methodology outlined in Section \ref{sub:JSON-Based_Structured_Text2Workflow}. This Text2Workflow approach breaks down the task of generating JSON workflows from natural language requests into modular steps with distinct prompts. We also assess the performance of gpt-3.5-0125 and gpt-4o-mini using extended prompts to evaluate their comprehension in longer contexts.

\subsection{General Architecture}
\label{subsec:mechanisms}

\begin{figure}[H]
        \centering
        \includegraphics[width=1.0\textwidth, height=1.0\textheight, keepaspectratio]{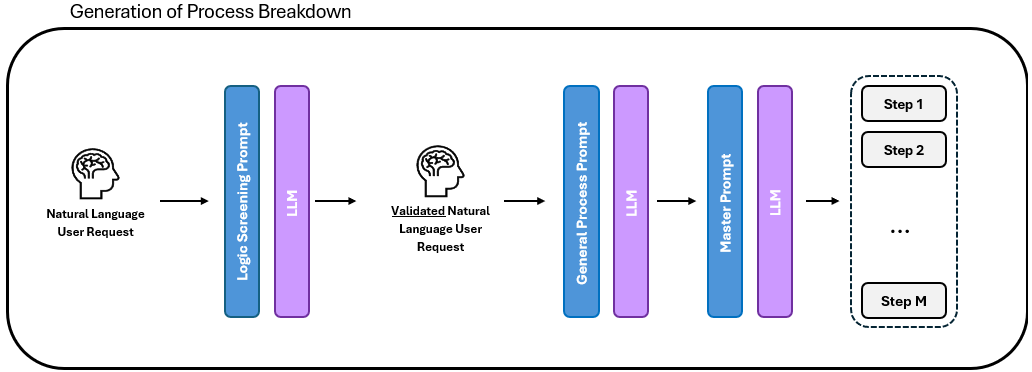}
        \caption{The process of breaking down a natural language user request into a workflow skeleton.}
        \label{fig:breakdown-generation-1}
    \end{figure}

\begin{figure}[H]
        \centering
        \includegraphics[width=1.0\textwidth, height=1.0\textheight, keepaspectratio]{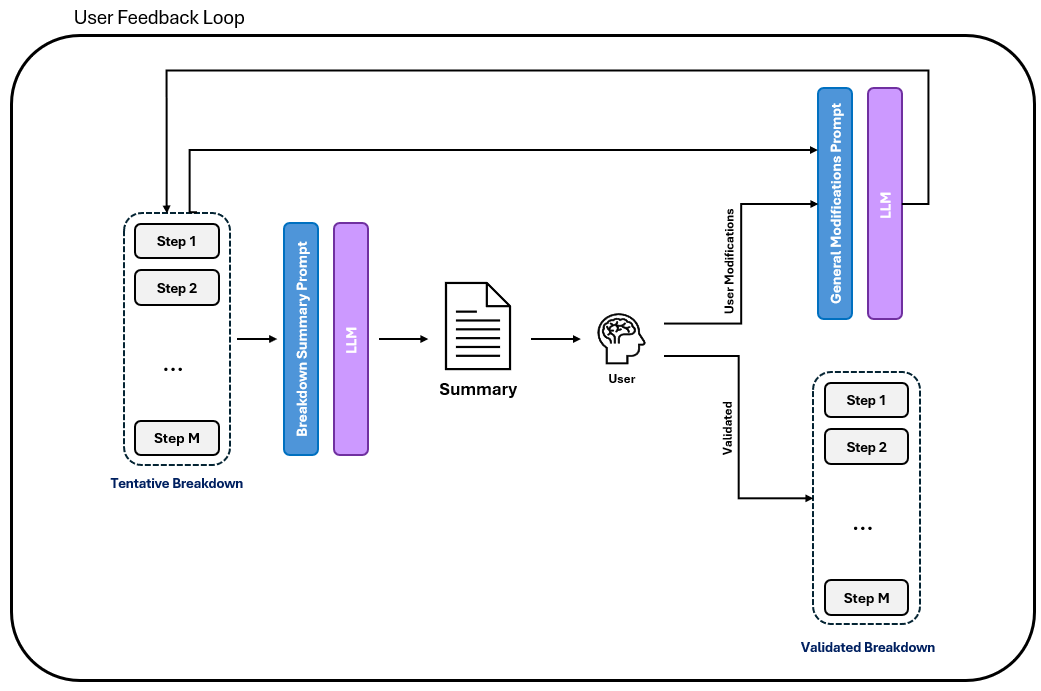} 
        \caption{The user feedback loop mechanism, enabling the user to validate and/or modify the workflow skeleton, if needs be.}
        \label{fig:user-feedback-loop-2}
    \end{figure}

\begin{figure}[H]
        \centering
        \includegraphics[width=1.0\textwidth, height=1.0\textheight, keepaspectratio]{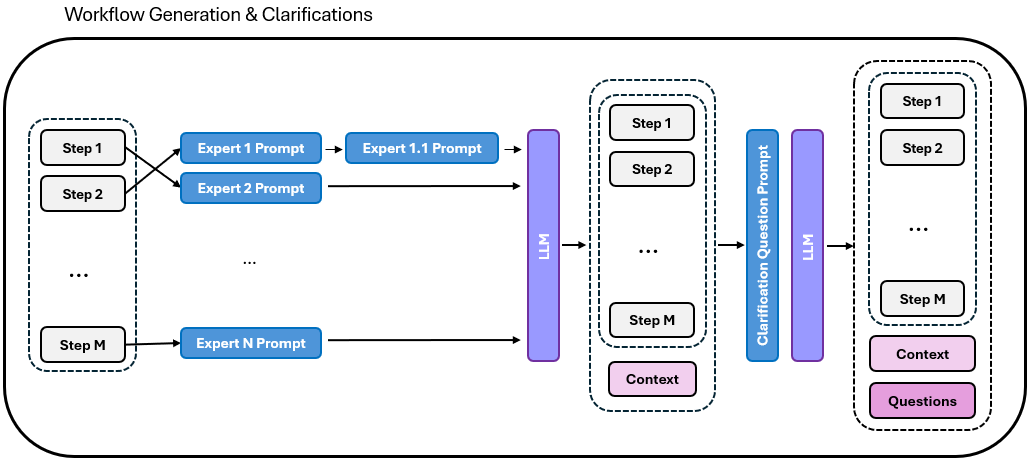}
        \caption{Complete workflow generation, including verifying and creating natural sounding questions for missing parameters.}
        \label{fig:workflow-n-questions}
    \end{figure}

\begin{figure}[H]
        \centering
        \includegraphics[width=1.0\textwidth, height=1.0\textheight, keepaspectratio]{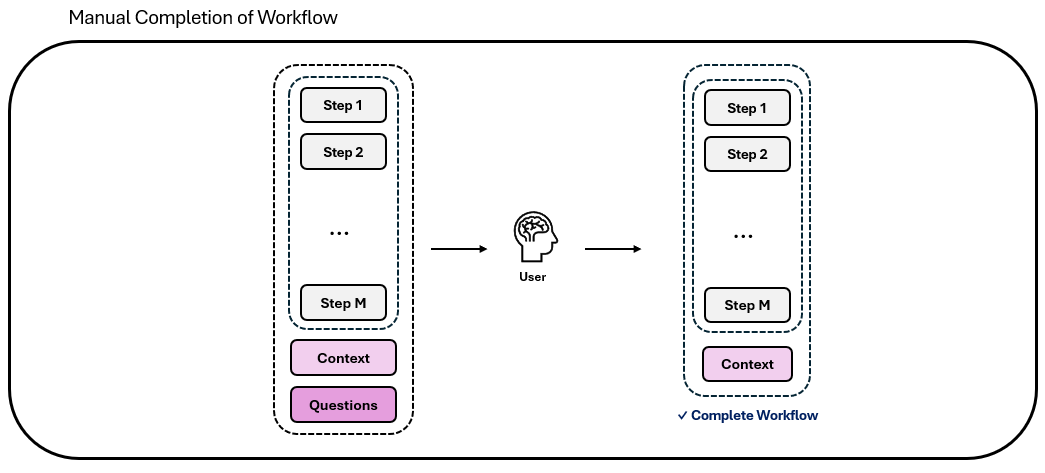} 
        \caption{Manually completing the workflow by allowing the user to fill in the missing parameters.}
        \label{fig:manual-completion-workflow}
    \end{figure}

\begin{figure}[H]
        \centering
        \includegraphics[width=1.0\textwidth, height=1.0\textheight, keepaspectratio]{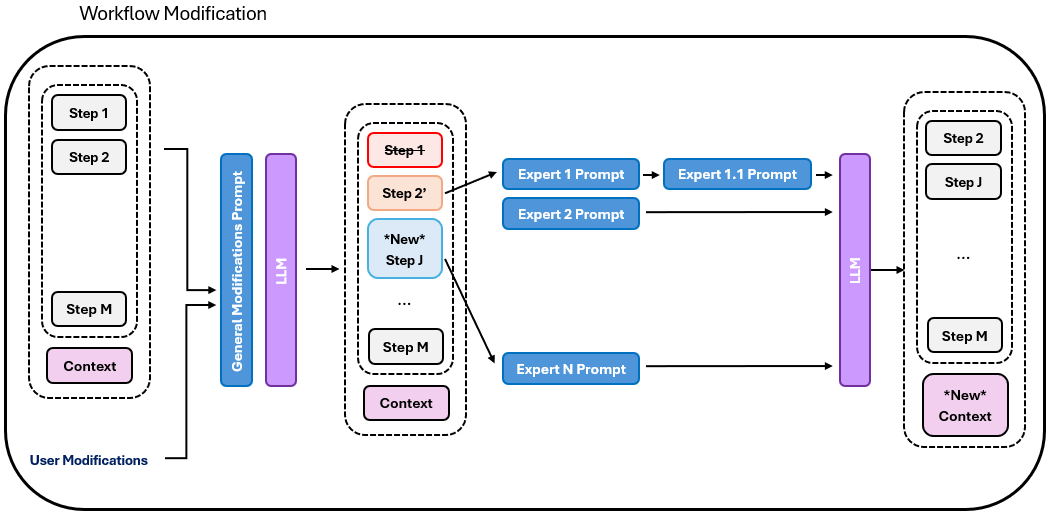}
        \caption{Workflow modification mechanism.}
        \label{fig:workflow-mod}
    \end{figure}

    Consequently, through the use of the seven distinct layers of prompts outlined in \Cref{sub:JSON-Based_Structured_Text2Workflow}, Text2Workflow can be summarized further into five main mechanisms, as listed below:

    \begin{enumerate}
        \item Breakdown Creation (\Cref{fig:breakdown-generation-1})
        \item User Feedback Loop (\Cref{fig:user-feedback-loop-2})
        \item Workflow Creation and Question Generation (\Cref{fig:workflow-n-questions})
        \item Manual Completion of Workflow (\Cref{fig:manual-completion-workflow})
        \item Workflow Modification (\Cref{fig:workflow-mod})
    \end{enumerate}

\section{Evaluation methodology}

    This section presents a thorough analysis of the metrics and evaluation criteria applied to our dataset. 

\subsection{Proposed Dataset}
\label{subsec:dataset}

    We constructed our own dataset, Process2JSON, consisting of $60$ concise example user requests. We define a short user request as being anything less than half a page. The dataset is categorized into three levels of complexity: $20$ easy examples, $20$ medium examples, and $20$ complex examples.

    Easy examples are deemed as examples that do not require Decision, Loop or Exception steps, and that are very clearly communicated, and having no ambiguity. Medium level examples include Decision and Loop steps but are logical and well-explained. Finally, complex examples are characterized by including all step types, nested Loops, consecutive Decisions, TryBlock steps and/or can contain implicit or ambiguous steps within the request. 
    
\begin{figure}[H]
        \center
        \includegraphics[width=0.9\textwidth, height=0.9\textheight, keepaspectratio]{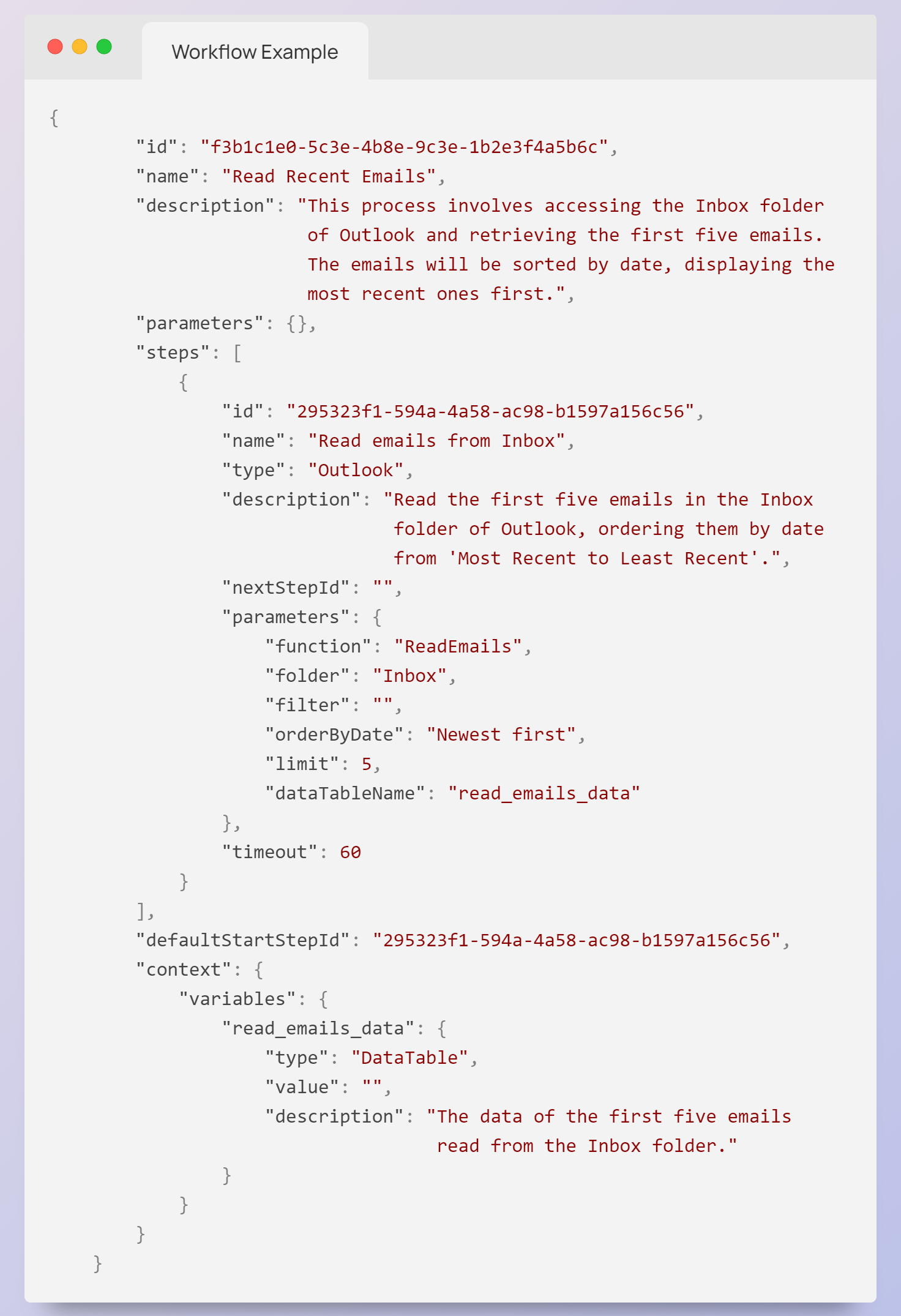}
        \caption{Expected Output JSON for Easy Sample.}
        \label{fig:expect_output_easy_sample}
    \end{figure}

    \Cref{tab:dataset-examples}  provides an example from each problem category within the Process2JSON dataset, along with the expected output for an easy sample, as shown in  \Cref{fig:expect_output_easy_sample}. 
    
    \begin{longtable}{|p{1.2cm}|p{11.8cm}|}
     \hline
     \multicolumn{1}{|c|}{\textbf{Difficulty}} & \multicolumn{1}{c|}{\textbf{Example}} \\
     \hline
     \centering Easy & Read the first five emails in the Inbox folder of Outlook, by ordering the dates from ‘Most Recent to Least Recent’.\\ 
     \hline
     \centering Medium & Read the file 'user/Downloads/Medical/Doctor\_Prescription.txt' and extract the patient name, doctor name, medication name, and the date of the letter. If the medication name is MEDEX and the date is before the 14 of April, 2020, send an email with the extracted data in body to 'report@recall.com' using Outlook.\\
     \hline
     \centering Complex & You have a folder 'C:/Feedback' containing text files with customer feedback from various product lines. Each text file is named according to the product line (e.g., 'ProductA\_Feedback.txt'). Get all the text files. For every file, extract the product name from its name. If a folder with this product name does not already exist, create it, then move the text file in question to this folder. If the folder already exists, move the text file to this folder. \\
     \hline
    \caption{An example from each difficulty level of Process2JSON: easy, medium and complex.}
    \label{tab:dataset-examples}
    \end{longtable}

\subsection{Evaluation and Metrics}
\label{subsec:eval-metrics}

    To evaluate Text2Workflow, we focused on three key criteria: (1) generation time, (2) token usage, and (3) JSON accuracy. These metrics provide a comprehensive view of both the tool’s efficiency and its output quality, allowing us to assess how well Text2Workflow performs in real-world applications.  Each of these metrics plays a crucial role in understanding the overall effectiveness of the system.

    For the first criterion, generation time is essential in assessing the efficiency of Text2Workflow, as the goal of automating workflow creation is to reduce the time required for building, modifying, and maintaining workflows. All tests were conducted using OpenAI’s API, meaning that generation time is influenced not only by Text2Workflow itself but also by the performance of the OpenAI platform, which may vary under different conditions.

    Token usage is another critical metric. Since the experimentation relied on OpenAI’s API models—specifically gpt-3.5-0125 and gpt-4o-mini — understanding token consumption provides insights into the potential costs of deploying Text2Workflow in real-world applications. However, if the underlying language model is hosted locally, token usage becomes less of a concern.

    Lastly, JSON accuracy evaluates more than just literal accuracy; it considers semantic consistency and contextual coherence throughout the JSON structure. Our evaluation differentiates between minor and major errors to provide a nuanced measure of quality. The scoring criteria for each level of accuracy are detailed in \Cref{tab:scoring-schema}, allowing a more granular assessment of JSON output quality.

    \begin{longtable}{|p{1.2cm}|p{11.8cm}|}
         \hline
         \multicolumn{1}{|c|}{\textbf{Score}} & \multicolumn{1}{c|}{\textbf{Description}} \\
         \hline
         \centering 1 & Perfect  JSON output. Coherent and semantically similar to the gold standard. Minor syntactical differences are permitted.\\ 
         \hline
         \centering 0.75 & The JSON output has expected structure, though minor reasoning errors are made. e.g erroneous API parameters, variable names are not prefixed with \$\{\} in strings. \\
         \hline
         \centering 0.5 & The JSON output is logical, but has considerable weaknesses in its structure.
        e.g missing Loop structure, hallucinating a non-existent \textit{nextStepId}, misuse of data extraction, incorrect function for API type step. \\
         \hline
         \centering 0.25 & The JSON output is comprehensible
        given the user request but there are several mistakes or a single big mistake. 
        e.g combination of mistakes such as those outlined above, hallucinating parameter keys or function names. \\
         \hline
        \centering 0 & The JSON output is not representative
        of the user request, lacks a great deal of information, or is nonsensical. \\
         \hline
        \caption{Scoring schema for Text2Workflow's JSON output.}
        \label{tab:scoring-schema}
    \end{longtable}








\section{Results and Discussion}
\label{sec:results}

    To evaluate the effectiveness of our Text2Workflow approach, which utilizes distinct prompts across multiple pipeline layers, we established two primary baseline models that rely on a single comprehensive prompt (see \ref{app:baseline-prompt}). This consolidated prompt includes all the instructions to generate a JSON workflow from a natural language user request. The first baseline uses the gpt-3.5-turbo-0125 \citep{GPT3-5} model, which was the latest available at the start of this project. The second baseline uses the same prompt with gpt-4o-mini (\texttt{gpt-4o-mini-2024-07-18}) \citep{GPT4o}, OpenAI's recent model optimized for both speed and cost, which reportedly offers improved performance over gpt-3.5. Due to the superior performance of gpt-4o-mini, Text2Workflow leverages this model as its foundation.

    We further performed an ablation study to examine the effects of each user input mechanism —specifically, general user clarifications and the user feedback loop. For the purpose of this paper, in all experiments involving the feedback loop, we have limited its use to two loops maximum.

    To summarize, six separate experiments were conducted:
    
    \begin{enumerate}
        \item \underline{\textbf{Baseline-gpt-3.5-0125}} : A single prompt run on OpenAI's model gpt-3.5-turbo-0125 \citep{GPT3-5}. 
        \item \underline{\textbf{Baseline-gpt-4o-mini}} : A single prompt run on OpenAI's model gpt-4o-mini-2024-07-18.
        \item \textbf{Text2Workflow-NUA (\underline{NUA})}: Text2Workflow with \textbf{N}o \textbf{U}ser \textbf{A}id.
        \item \textbf{Text2Workflow-GC (\underline{GC})}: Text2Workflow with the initial \textbf{Ge}neral \textbf{C}larifications verification (logic screening).
        \item \textbf{Text2Workflow-HFL (\underline{HFL})}: Text2Workflow with a \textbf{H}uman \textbf{F}eedback \textbf{L}oop at the process breakdown level. For the purpose of this research paper, we have limited the number of times the feedback loop can be used to two, though in practice this loop could be infinite.
        \item \underline{\textbf{Text2Workflow}}: The complete Text2Workflow pipeline with both user input mechanisms (GC, HFL). Again, for the purpose of this research paper, we have limited the number of times the feedback loop can be used to two.
    \end{enumerate}

\begin{figure}[H]
        \centering
        \makebox[\textwidth][c]{
        \includegraphics[width=1.0\textwidth, height=0.7\textheight, keepaspectratio]{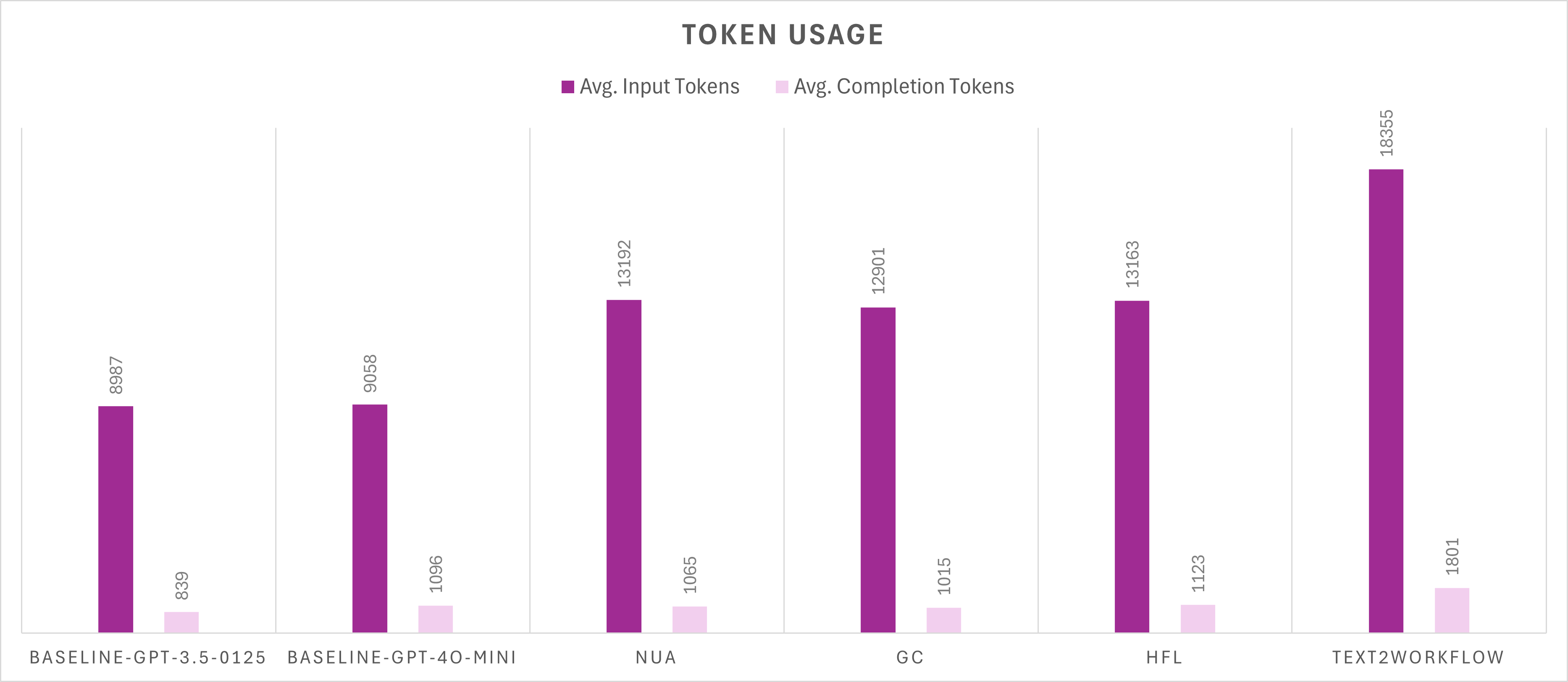}
        }
        \caption{Average token usage, across all experiments.}
        \label{fig:token_usage}
    \end{figure}

    We initiate our study of the Text2Workflow approach by evaluating its token consumption and generation efficiency. As discussed in \Cref{subsec:eval-metrics}, this metric provides insight into the potential cost of the solution. Figure~\ref{fig:token_usage} shows that Text2Workflow consumes, on average, significantly more tokens for both input and completion than the two baselines and the three ablation models (NUA, GC, HLF). The difference in token consumption is especially notable for input tokens. For instance, the baseline-gpt-3.5-0125 uses an average of $8987$ input tokens, closely followed by baseline-gpt-4o-mini with $9058$ tokens. In contrast, the Text2Workflow approach averages $18 355$ tokens, more than double that of both baselines. This higher token usage is expected, given the numerous prompts in Text2Workflow, especially with the human feedback loop adding two additional prompts each time it is re-run.

    We observe, nonetheless, that the baseline solutions also require a substantial number of tokens due to the length of their prompts. In comparison, the Text2Workflow version without user assistance, NUA, uses approximately 45\% more input tokens than the baseline-gpt-4o-mini. This indicates that although the \textit{Expert Prompts} contribute to the increased token consumption, a substantial number of tokens are inherently necessary to thoroughly explain and complete the task.

\begin{figure}[H]
        \centering
         \makebox[\textwidth][c]{
        \includegraphics[width=1.0\textwidth, height=0.7\textheight, keepaspectratio]{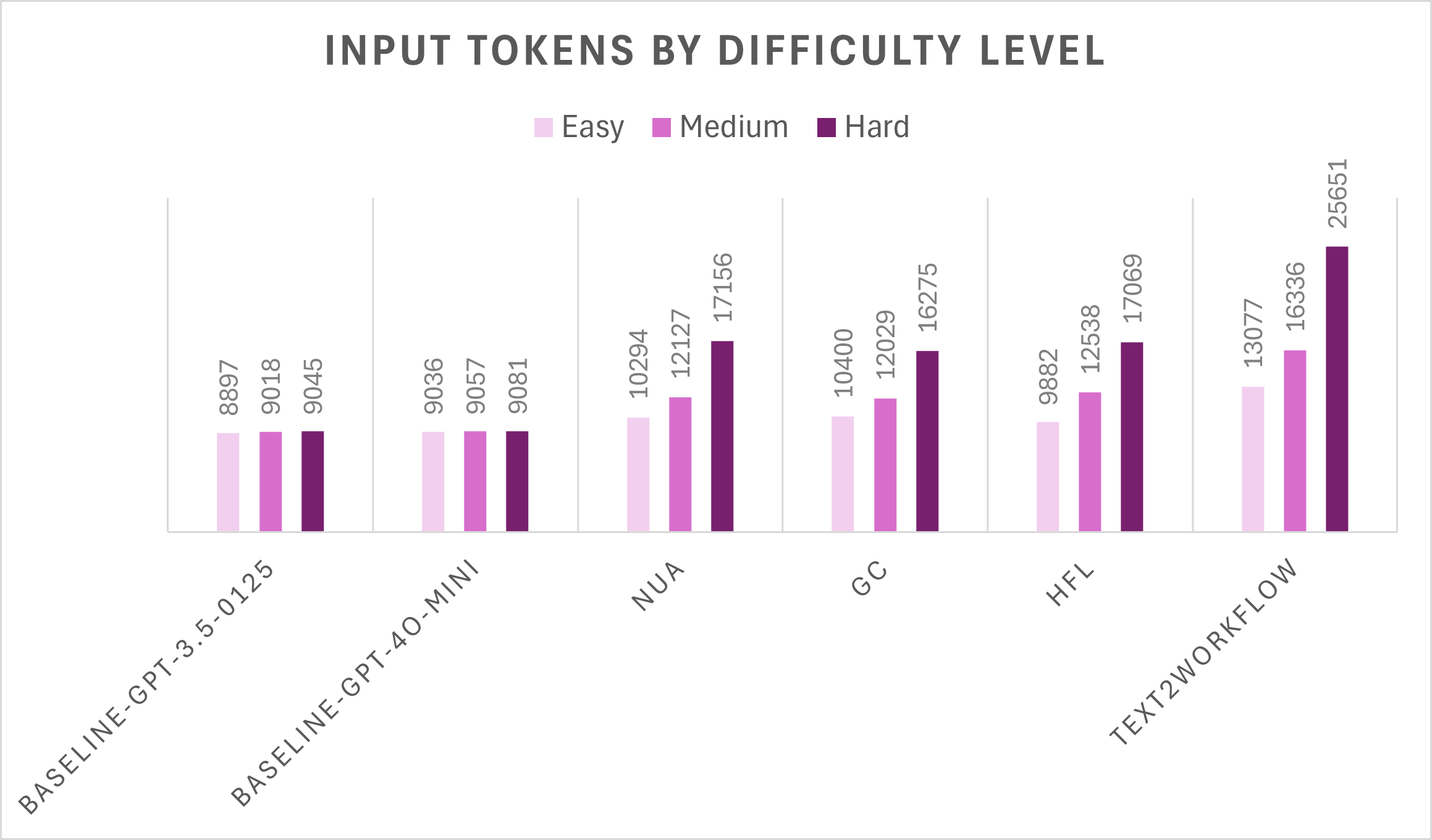}
        }
        \caption{Input token usage, across all experiments and by difficulty level.}
         \label{fig:input-token-usage}
    \end{figure}

    We further analyzed the average number of input and completion tokens per experiment, categorized by the difficulty of user requests. As shown in Figure~\ref{fig:input-token-usage}, the more difficult user requests unsurprisingly consume the highest number of input tokens across all experiments. This trend is especially pronounced in the NUA, GC, HFL, and Text2Workflow approaches. While both baselines maintain a relatively consistent token consumption across the three difficulty levels (with less than a 150-token difference), Text2Workflow’s token consumption for the hard-level samples is significantly higher, using approximately 183\% more tokens than baseline-gpt-3.5, 182\% more than baseline-gpt-4o-mini, 49\% more than NUA, 57\% more than GC, and 50\%\ more than HFL.

\begin{figure}[H]
        \centering
         \makebox[\textwidth][c]{
        \includegraphics[width=1.0\textwidth, height=0.7\textheight, keepaspectratio]{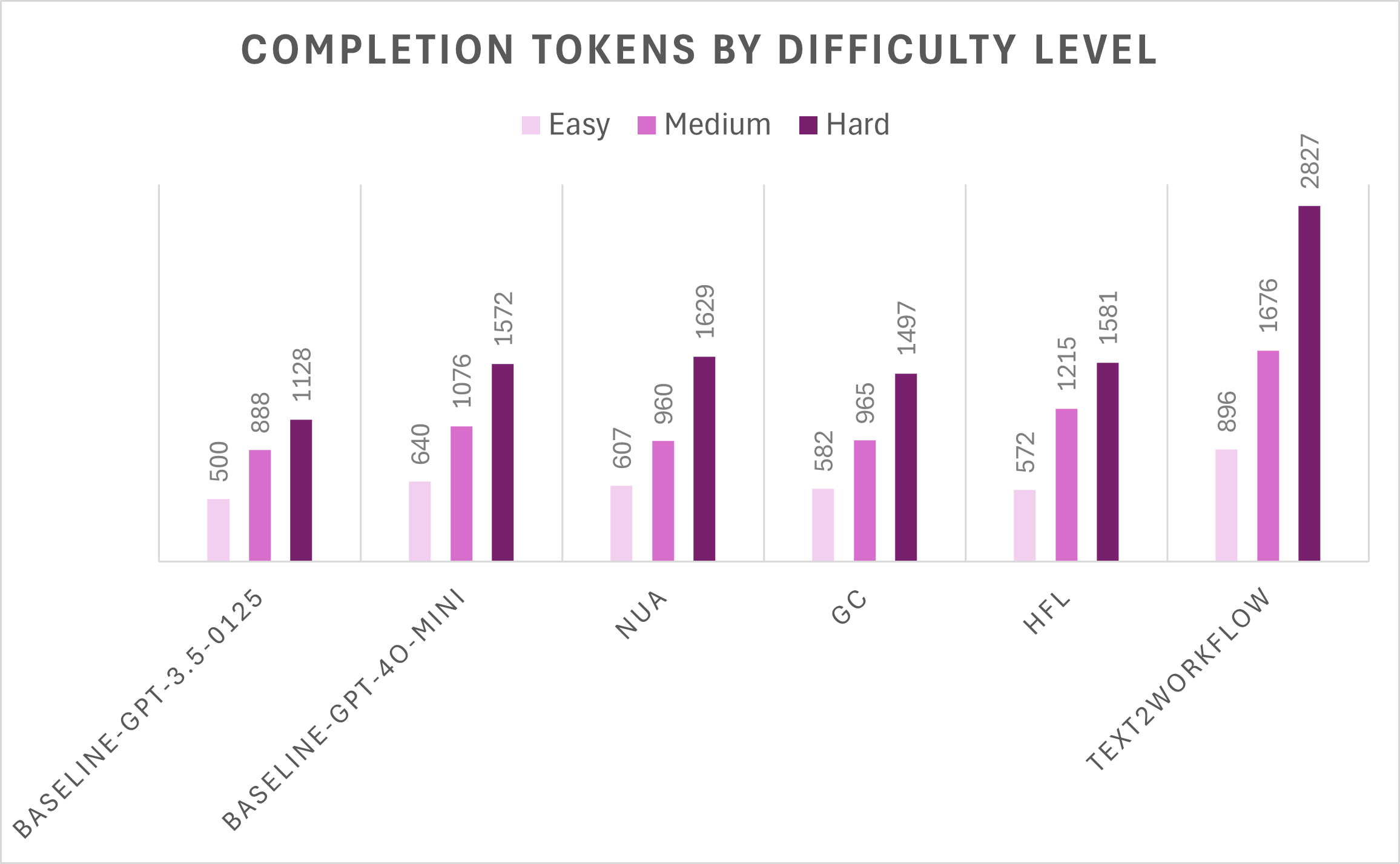}
        }
        \caption{Completion token generation, across all experiments and by difficulty level.}
        \label{fig:completion-token-usage}
    \end{figure}

    Moreover, Figure~\ref{fig:completion-token-usage} presents the completion token usage categorized by difficulty level. In contrast to input tokens, the variation in completion tokens across different difficulty levels is more significant, with easy samples generating fewer tokens and hard samples yielding a considerably higher number. This discrepancy primarily arises because the JSON output for easy samples involves fewer steps and less extensive context. Notably, Text2Workflow exhibits the highest completion token usage, particularly at the hard level—consuming 79\% more tokens than baseline-gpt-4o-mini, 73\% more than NUA, 88\% more than GC, and 78\% more than HFL. In comparison, baseline-gpt-3.5-0125 has the lowest completion token usage for hard tasks, with $1128$ tokens compared to Text2Workflow's $2827$ tokens.

\begin{figure}[H]
        \centering
        \makebox[\textwidth][c]{
        \includegraphics[width=1.0\textwidth, height=1.0\textheight, keepaspectratio]{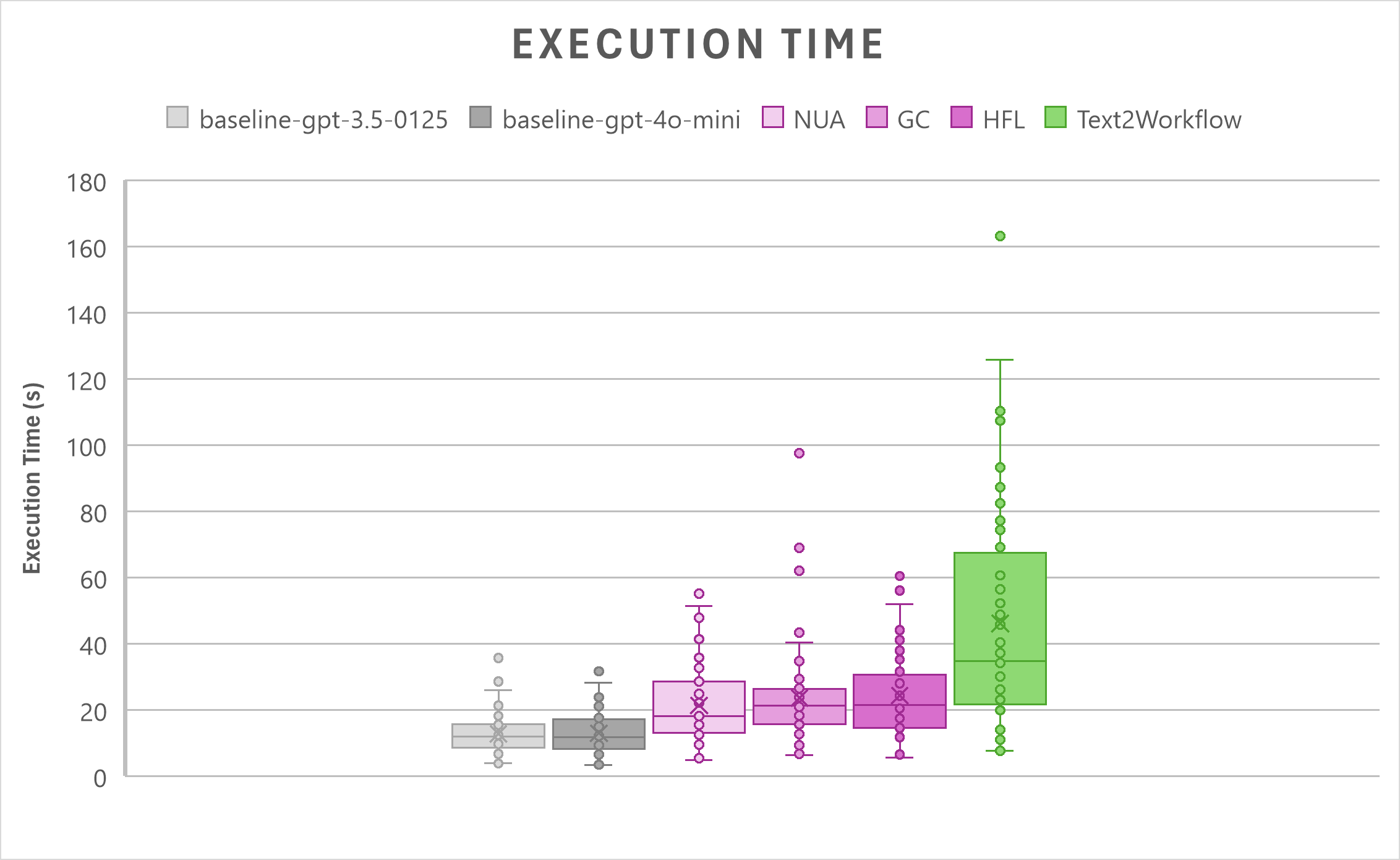}
        }
        \caption{Execution time (in seconds), across all experiments.}
         \label{fig:time-analysis}
    \end{figure}

     The second metric we examine is execution time, with results summarized in Table \ref{tab:time-results}. As expected, the baseline methods, due to their simplicity, demonstrated the shortest average execution times and the least variance compared to other methods. Text2Workflow, with its multi-step structure, required a longer duration (Figure \ref{fig:time-analysis}). A key observation is that the maximum execution time for Text2Workflow can extend up to $163$ seconds when substantial modifications are necessary. It is important to note, however, that while Text2Workflow may have a relatively extended execution time, this duration is segmented into shorter intervals due to sequential user interactions, making the overall experience more manageable.

    \begin{table}[H]
    \centering
    \resizebox{0.99\textwidth}{!}{
        \begin{tabular}{ccccccc}
         \hline
           & \centering \makecell{\textbf{Baseline} \\ \textbf{(gpt-3.5-0125)}} & \centering \makecell{\textbf{Baseline} \\\textbf{(gpt-4o-mini)}} & \textbf{NUA}  & \textbf{GC} & \textbf{HFL} & \textbf{Text2Workflow} \\
         \hline
         \hline
         \centering \textbf{Average Time $\downarrow$} & \centering 12.8 &  \centering 13.0 & \centering 21.4 & 23.6 &  24.3 & 46.2 \\
         \hline
         \centering \textbf{Median Time $\downarrow$} & \centering 11.9 & \centering 11.7 & \centering 18.2 & 21.3 &  21.5 & 34.8\\ 
         \hline
        \centering \textbf{Max Time $\downarrow$} & \centering 35.6 & \centering 31.7 & \centering 55.1 & 97.5 &  60.5 &  163.1\\  
        \hline
        \centering \textbf{Min Time $\downarrow$} & \centering 3.8 &  \centering 3.3 & \centering 4.8 &  6.3 &  5.7 &  7.6 \\ 
         \hline
        \end{tabular}
        }
    \caption{Summary statistics for the execution times, in seconds, of each method.}
    \label{tab:time-results}
    \end{table}

    For the third metric, Table \ref{tab:results} presents the accuracy achieved across all experiments. Among the baselines, baseline-gpt-4o-mini demonstrates superior performance compared to baseline-gpt-3.5-0125, achieving 92.5\% accuracy for easy-level user requests, 70\% for medium level, and 30\% for hard level. In contrast, baseline-gpt-3.5-0125 scores lower, with 53.8\% accuracy for both easy and medium levels and only 8.8\% for hard-level requests. Furthermore, baseline-gpt-4o-mini outperforms the NUA (78.8\%), GC (85\%), HFL (88.8\%), and Text2Workflow (87.5\%) pipelines in easy-level user requests. However, at the hard level, Text2Workflow surpasses baseline-gpt-3.5-0125 by 48.7\%, and equally outperforming baseline-gpt-4o-mini (by 27.5\%), NUA (by 31.2\%), GC (by 30\%), and HFL (by 25\%).

    \begin{table}[H]
        \centering
        \resizebox{0.99\textwidth}{!}{
        \begin{tabular}[\textwidth]{ccccccc}
         \hline
          \centering \textbf{Accuracy $\uparrow$} & \makecell{\textbf{Baseline} \\ \textbf{(gpt-3.5-0125)}} & \centering \makecell{\textbf{Baseline} \\\textbf{(gpt-4o-mini)}} & \textbf{NUA}  & \textbf{GC} & \textbf{HFL} & \textbf{Text2Workflow} \\
         \hline
         \hline
         \centering Easy & \centering 53.8 & \centering \textbf{92.5} & 78.8 &  85 & 88.8 & 87.5 \\ 
         \hline
        \centering Medium & \centering 53.8 & \centering \textbf{70} & 48.8 &  47.5 &  65 & 68.8 \\  
        \hline
        \centering Hard & \centering 8.8 & \centering 30 &  26.3 &  27.5 &  32.5 & \textbf{57.5} \\ 
         \hline
        \end{tabular}
        }
        \caption{Average JSON accuracy results (in \%), by difficulty level.}
        \label{tab:results}
    \end{table}

\begin{figure}[H]
        \centering
        \includegraphics[width=0.9\textwidth, height=0.7\textheight, keepaspectratio]{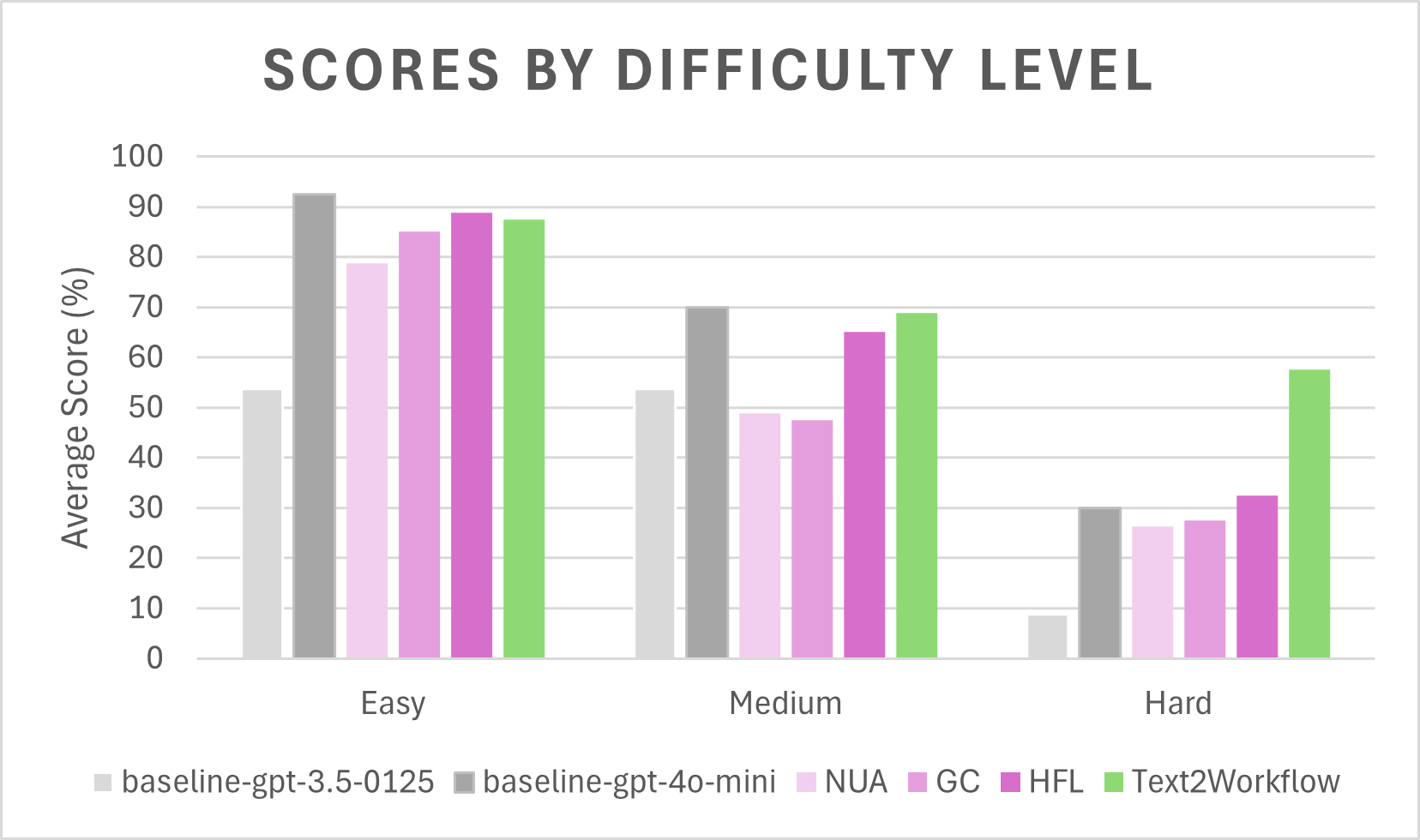}
        \caption{The JSON accuracy results (in \%), by difficulty level.}
        \label{fig:scores-by-diff}
    \end{figure}

    Figure~\ref{fig:scores-by-diff}  highlights the advantages of the HFL approach for medium and hard levels; HFL improves the accuracy of baseline-4o-mini, NUA, and GC by 2.5\%, 6.2\%, and 5\%, respectively. For user requests ranging from easy to medium difficulty, baseline-gpt-4o-mini marginally surpasses Text2Workflow, though never by more than 3\%. However, with increasing complexity, Text2Workflow consistently exceeds all other approaches by a margin of over 20\%.

\begin{figure}[H]
        \centering
        \makebox[\textwidth][c]{
        \includegraphics[width=1.0\textwidth, height=0.7\textheight, keepaspectratio]{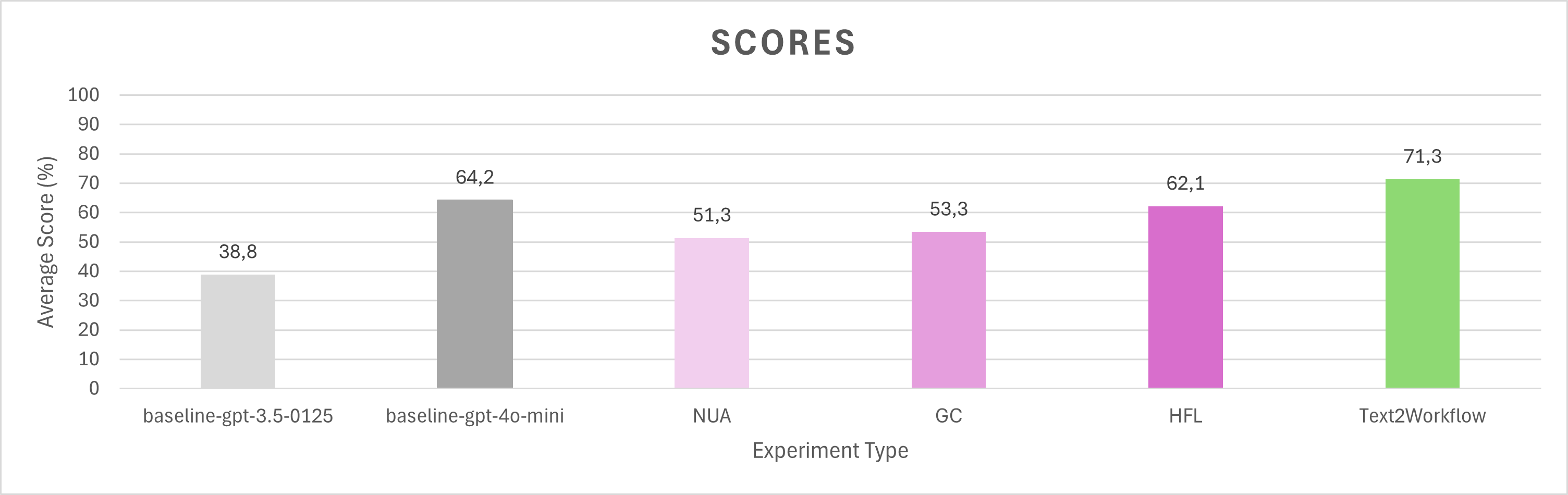}
        }
        \caption{The overall performances of every method (in \%).}
        \label{fig:scores-overall}
    \end{figure}

   Overall, Figure~\ref{fig:scores-overall} shows that Text2Workflow achieves the highest performance across experiments, with a 71.3\% accuracy, followed by baseline-gpt-4o-mini (64.2\%), HFL (62.1\%), and GC (53.3\%), while baseline-gpt-3.5-0125 trails at 38.8\%. Additionally, user input mechanisms, including general clarifications and the human feedback loop, contributed significantly to improved accuracy. However, there was minimal improvement observed between the NUA and GC mechanisms.

   The main drawbacks of the Text2Workflow approach include several issues that impact the accuracy and functionality of the workflows it generates. First, there are frequent mistakes with nextStepIds, especially when handling Decision and Loop steps, which can disrupt the flow of the workflow. Another issue is the misuse, or complete lack of use, of context variables, which affects the workflow's ability to maintain state and pass necessary data between steps. Additionally, the logic screening isn’t robust enough to identify requests that could be clarified for better accuracy, resulting in workflows that may not align closely with the intended actions. Another more technical error is that DataExtraction cannot be used to read information, but rather can only extract from text that is already stored in a variable - a subtlety that the agent has trouble with. The most significant issue is its pronounced difficulty with Exception steps, particularly when the function is set to TryBlock, as it struggles to handle these cases properly. Despite these challenges, there are no structural errors in the JSON format, and the output is consistently verbose as expected. Refer to \ref{app:example_failures} for an example of failures encountered in the JSON workflow generation process.

\section{Limitations}

   According to \cite{TPTU}, LLMs demonstrate four main limitations: inconsistency in adhering to specified formats, challenges in fully interpreting task instructions, over-dependence on single tools, and limited summarization capabilities. Given Text2Workflow's strong reliance on LLMs, it similarly encounters these issues. As demonstrated in Section \ref{sec:results}, although infrequent, Text2Workflow sometimes generates incorrect keys in the JSON structure that should not exist (e.g. Exception steps with Try-Blocks tend to have keys that are not in the predefined structure).

    Another significant limitation is the specificity of the prompts. While the Master and Experts strategy helps break down the task into smaller, more manageable problems, the prompts themselves remain highly specialized. This specificity poses challenges for maintenance, especially if the JSON structure, step types, or function/API lists are updated, requiring substantial rework to adapt the prompts accordingly. Moreover, the use of multiple prompts may result in better performance but comes at the cost of consuming more tokens and increasing the complexity of maintenance.

    Additionally, the current solution exclusively uses GPT, a cloud-based service, which could introduce potential security concerns for businesses.

    Finally, the evaluation method in this study requires further research, as it is biased by reliance on a single evaluator. A more robust approach would use multiple evaluators with averaged or consensus scoring, or develop a metric to capture the system's performance more objectively.

    \section{Conclusions and Future Work}
    \label{sec:conclusions}

    This paper presents Text2Workflow, a novel approach designed for generalized business workflow automation. Additionally, we introduce a dataset of general user requests translated into JSON workflows, named Process2JSON. To the best of our knowledge, no existing methods in the literature have addressed the generalized automation of workflows spanning a diverse range of business processes.

    The results of our experiments demonstrated that Text2Workflow significantly outperforms OpenAI's gpt-4o-mini model used with a single long prompt, specifically when treating complex user requests. Our ablation study confirmed that the user input mechanisms integrated into the Text2Workflow pipeline substantially enhance performance, with the user feedback loop alone contributing to an improvement of over 10\%. While the initial logic screening of user requests also enhanced results, the increase was modest at 2\%, likely due to its infrequent invocation. Further exploration into optimizing the \textit{Logic Screening Prompt} could yield additional benefits for the Text2Workflow approach.

    Despite these advancements, we acknowledge several limitations, including inconsistencies, difficulties in fully interpreting task instructions, over-reliance on specific tools, and constrained summarization capabilities. Nonetheless, we believe that this approach marks a significant step forward in the field of automated workflow generation across various business processes and industries.

    Looking ahead, future work will focus on addressing these limitations by strengthening the robustness of the prompts to minimize the occurrence of incorrect keys in the JSON structure. This may involve creating a more flexible prompting framework capable of adapting to changes in the JSON structure, step types, or function/API lists without requiring extensive rework. Additionally, exploring alternatives to GPT could help alleviate security concerns for businesses using cloud-based services. We also intend to enhance our evaluation methodology by incorporating multiple evaluators, providing a more objective assessment of the system’s performance and improving the reliability of our findings. Finally, ongoing research will target improvements in summarization capabilities and task interpretation to ensure that Text2Workflow can function effectively across a broader range of scenarios.



\newpage
\textbf{Acknowledgments}

We would like to acknowledge Novelis for their support in publishing this article. We are especially grateful for the assistance and contributions of their research team.



\bibliographystyle{elsarticle-harv} 
\bibliography{bibliography}


\newpage

\appendix

\section{Text2Workflow Step-Type Specific JSON Structures}
\label{app:json-struct}

\begin{figure}[H]
    \centering
    \includegraphics[width=0.7\textwidth, height=0.7\textheight, keepaspectratio]{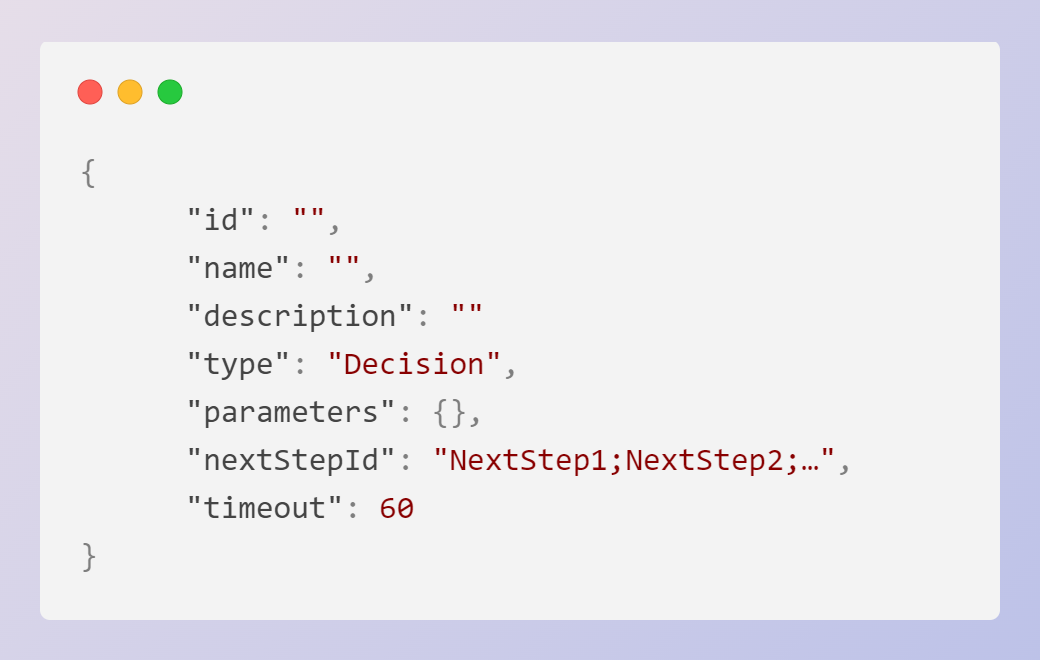}
    \caption{JSON structure for Decision type step.} 
    \label{fig:decisionJSON}
\end{figure}

\begin{figure}[H]
    \centering
    \includegraphics[width=0.67\textwidth, height=0.67\textheight, keepaspectratio]{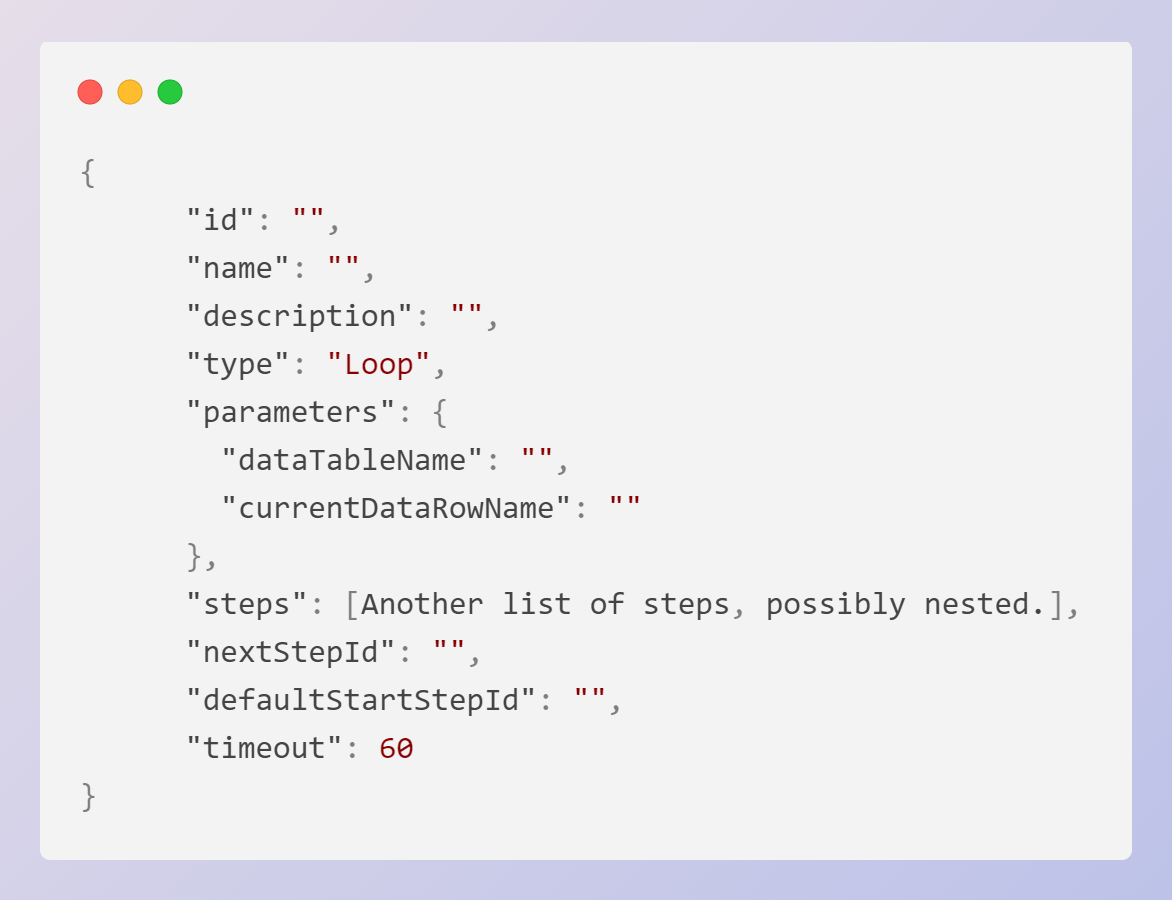}
    \caption{JSON structure for Loop type step.} 
    \label{fig:loopJSON}
\end{figure}

\begin{figure}[H]
    \centering
    \includegraphics[width=0.67\textwidth, height=0.67\textheight, keepaspectratio]{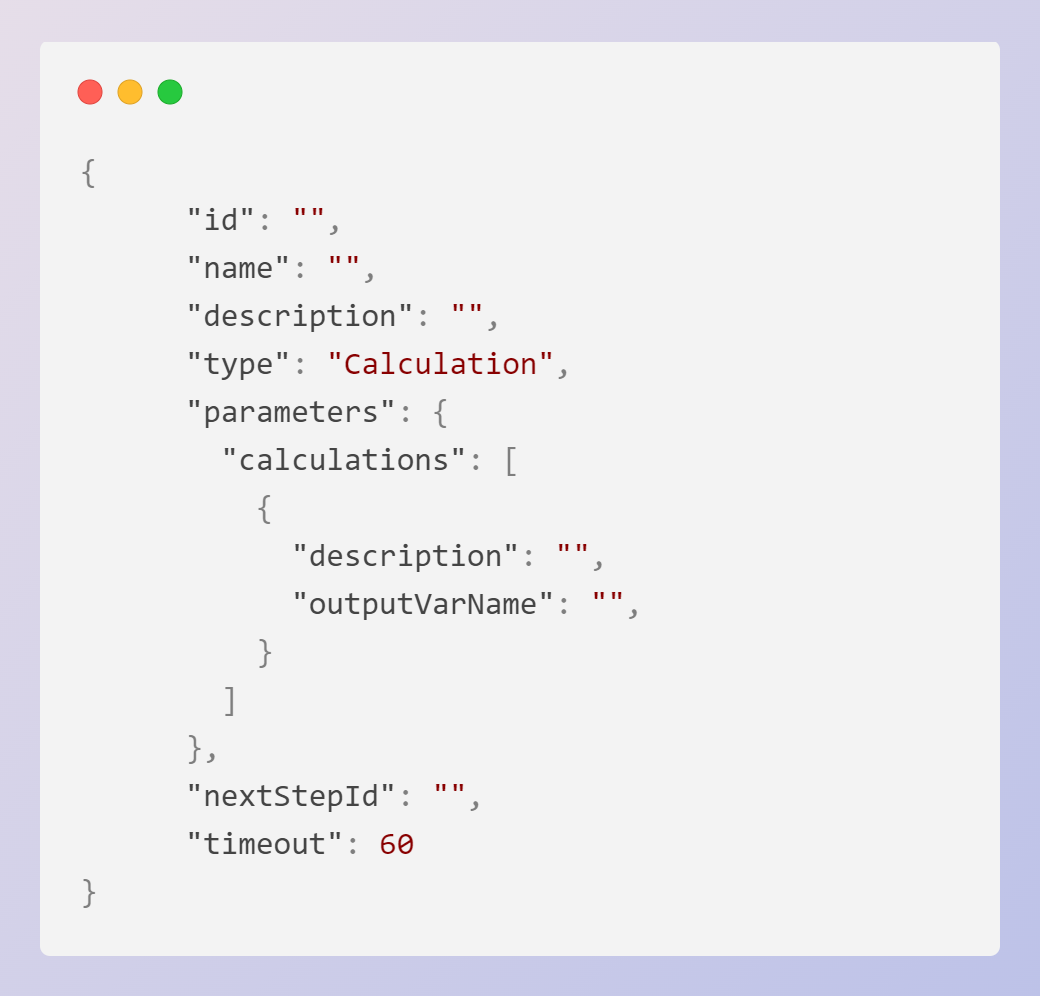}
    \caption{JSON structure for Calculation type step.} 
    \label{fig:calcJSON}
\end{figure}

\begin{figure}[H]
    \centering
    \includegraphics[width=0.7\textwidth, height=0.7\textheight, keepaspectratio]{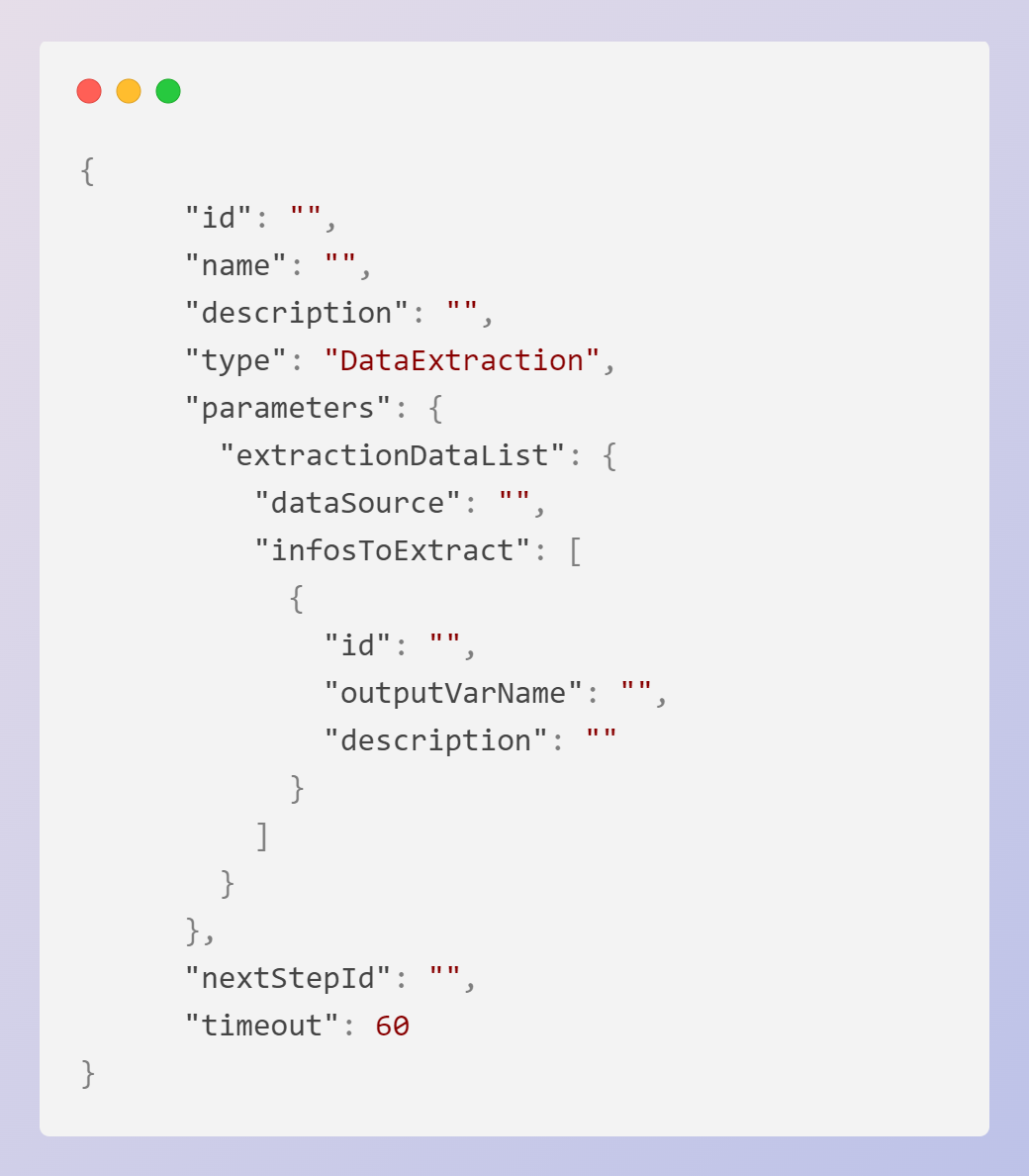}
    \caption{JSON structure for DataExtraction type step.} 
    \label{fig:extractJSON}
\end{figure}

\begin{figure}[H]
    \centering
    \includegraphics[width=0.67\textwidth, height=0.67\textheight, keepaspectratio]{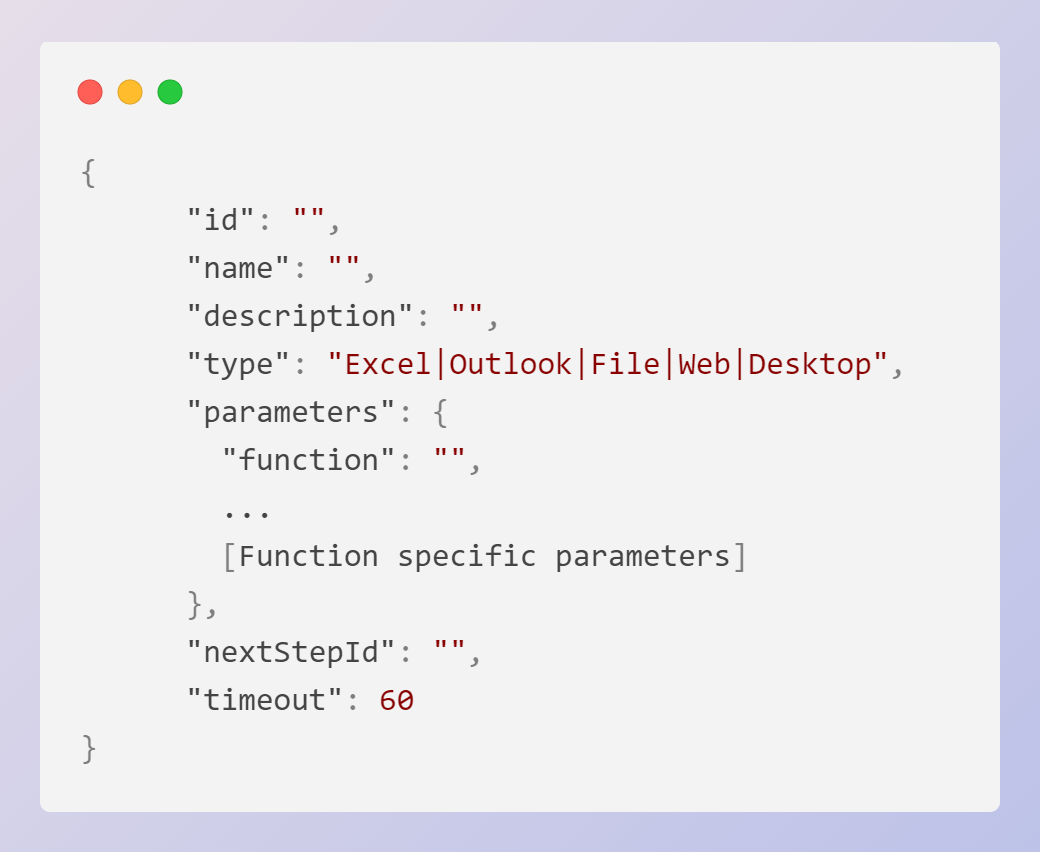}
    \caption{JSON structure for an API, Web or Desktop type step.} 
    \label{fig:taskJSON}
\end{figure}

\begin{figure}[H]
    \centering
    \includegraphics[width=0.67\textwidth, height=0.67\textheight, keepaspectratio]{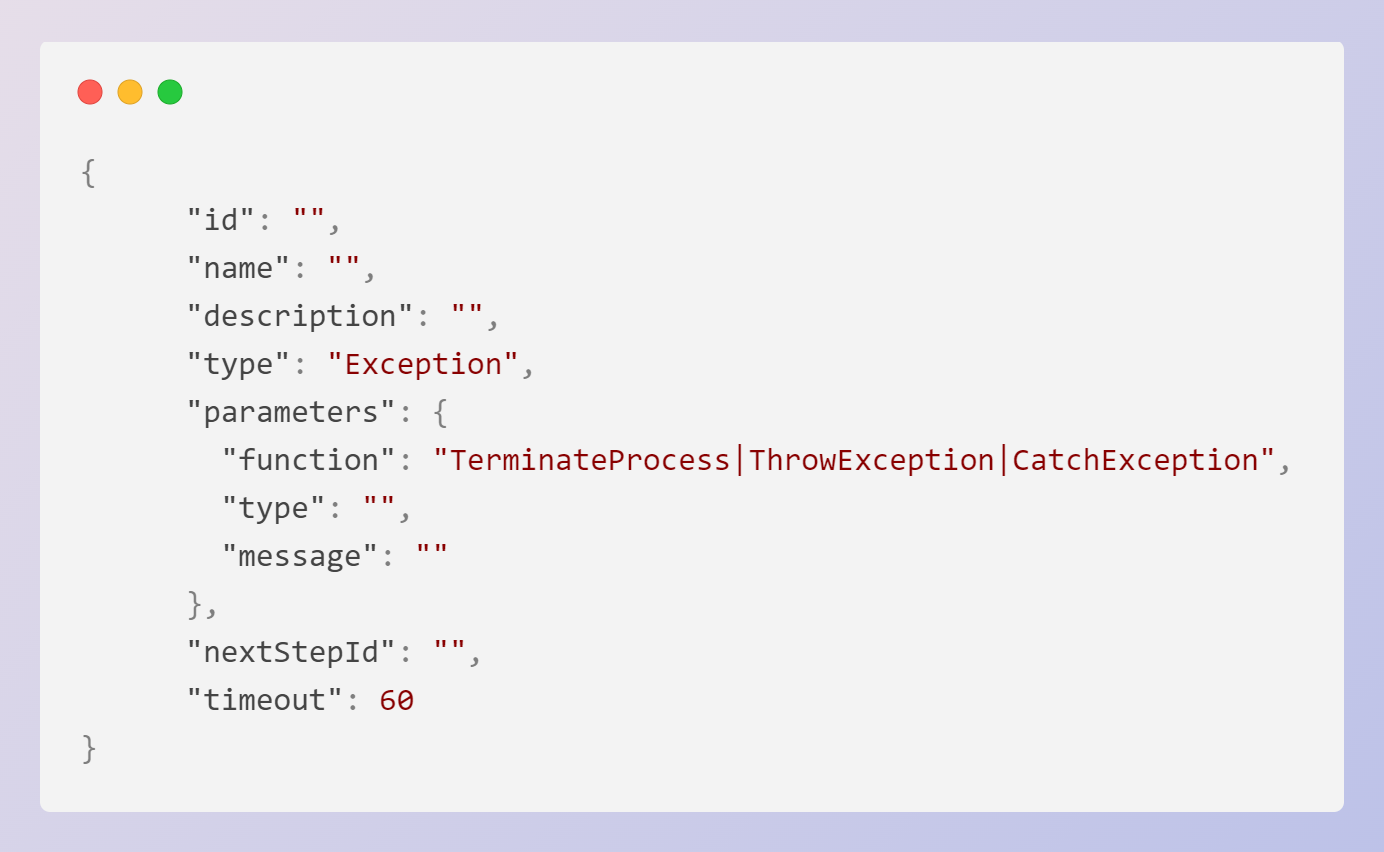} 
    \caption{JSON structure for an Exception (TerminateProcess, ThrowException, CatchException) type step.} 
    \label{fig:exceptionJSON}
\end{figure}

\begin{figure}[H]
    \centering
    \includegraphics[width=0.7\textwidth, height=0.7\textheight, keepaspectratio]{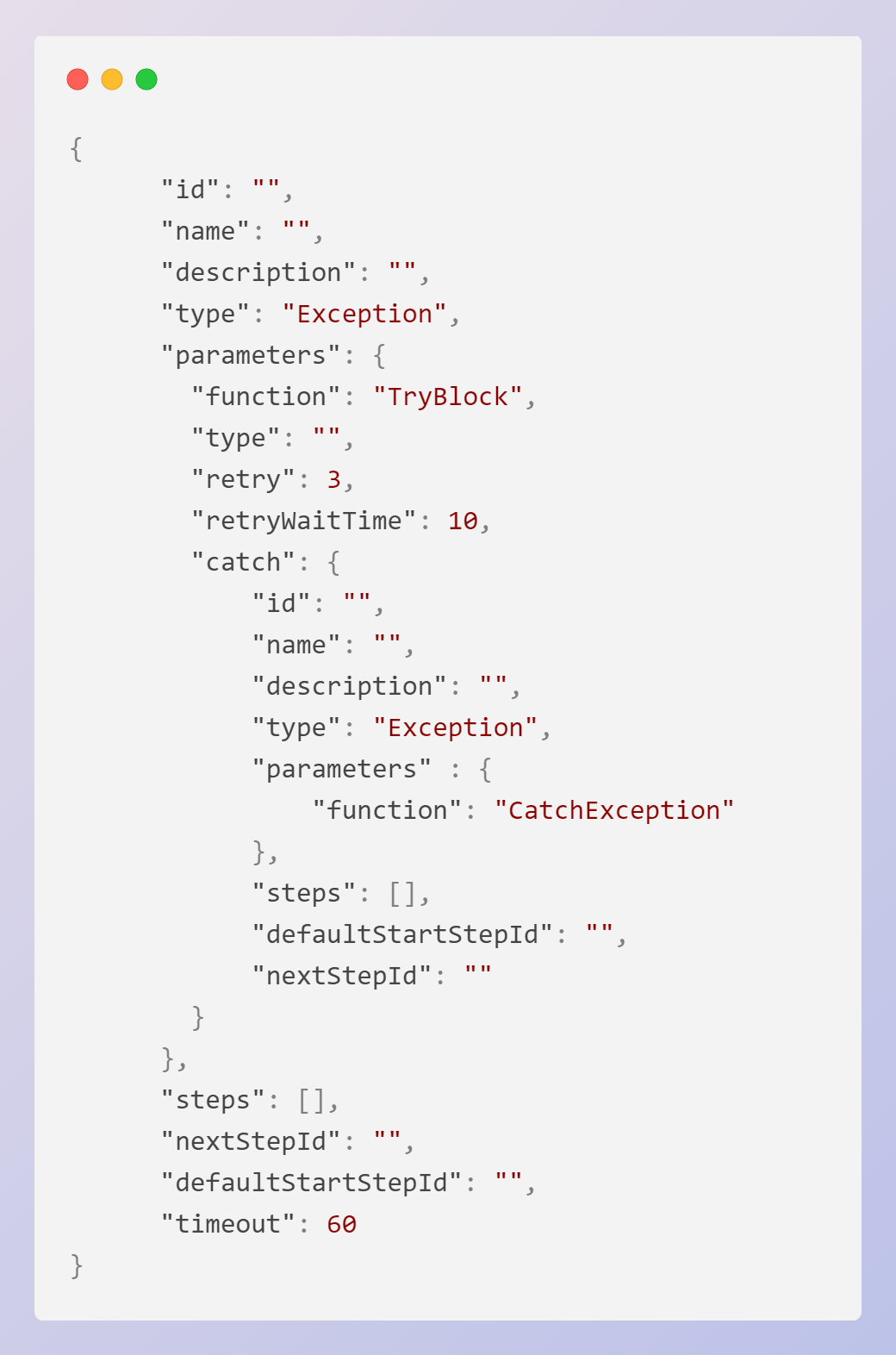} 
    \caption{JSON structure for an Exception-TryBlock type step.}
    \label{fig:TryBlockJSON}
\end{figure}

In this appendix, we present a detailed JSON structure for each of the seven step types: Decision (Figure~\ref{fig:decisionJSON}), Loop (Figure~\ref{fig:loopJSON}), Calculation (Figure~\ref{fig:calcJSON}), Data Extraction (Figure~\ref{fig:extractJSON}), API (Figure~\ref{fig:taskJSON}), Exception (Figure~\ref{fig:exceptionJSON}), and Unknown (Figure~\ref{fig:TryBlockJSON}).

\newpage
\section{Text2Workflow Prompts}
\label{app:prompts}

\begin{figure}[H]
    \centering
    \includegraphics[width=0.8\textwidth, height=0.8\textheight, keepaspectratio]{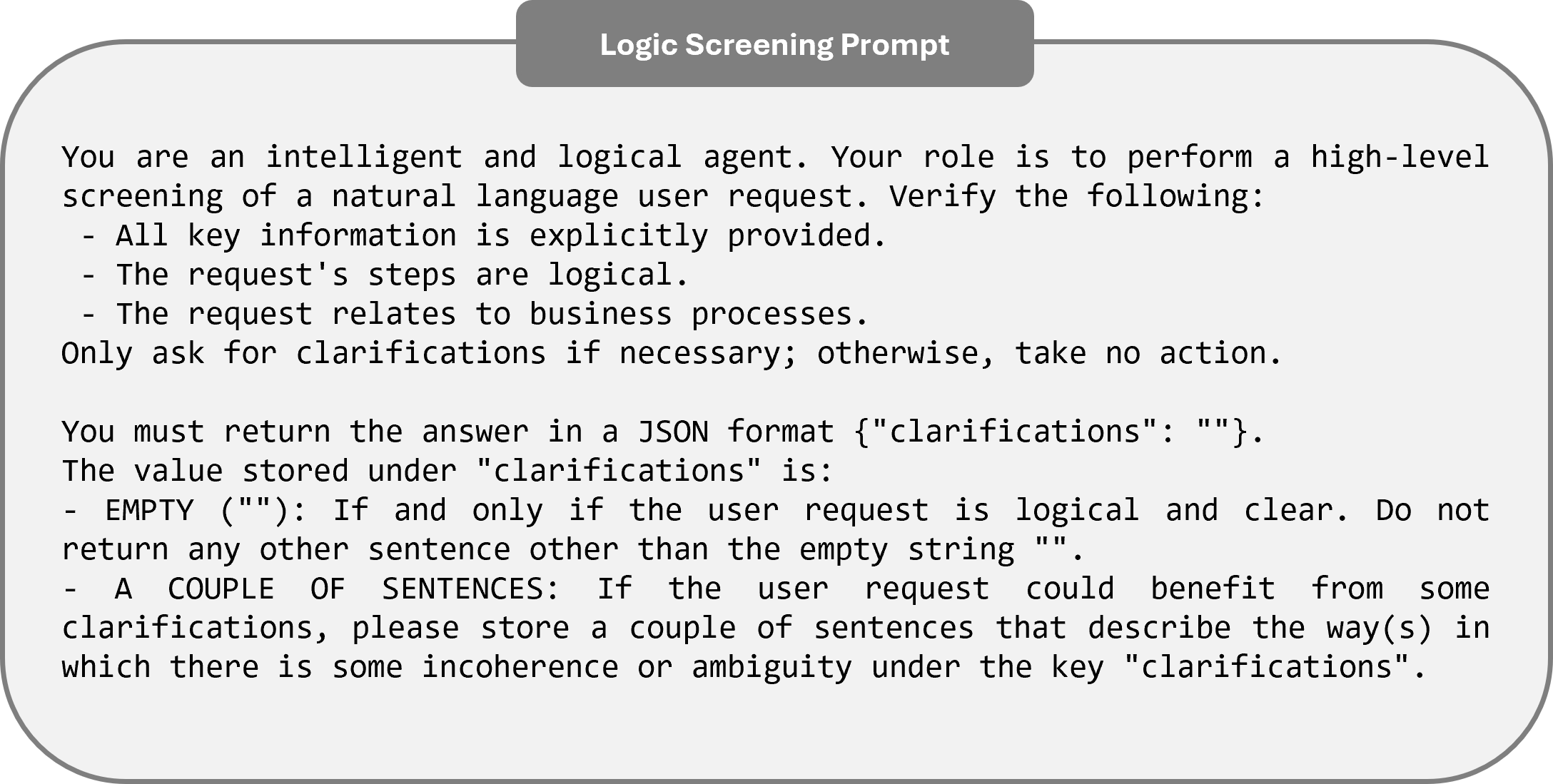} 
    \caption{The \textit{User Request Ambiguity Screening Prompt}.} 
    \label{fig:logic-screening-prompt}
\end{figure}

\begin{figure}[H]
    \centering
    \includegraphics[width=0.8\textwidth, height=0.8\textheight, keepaspectratio]{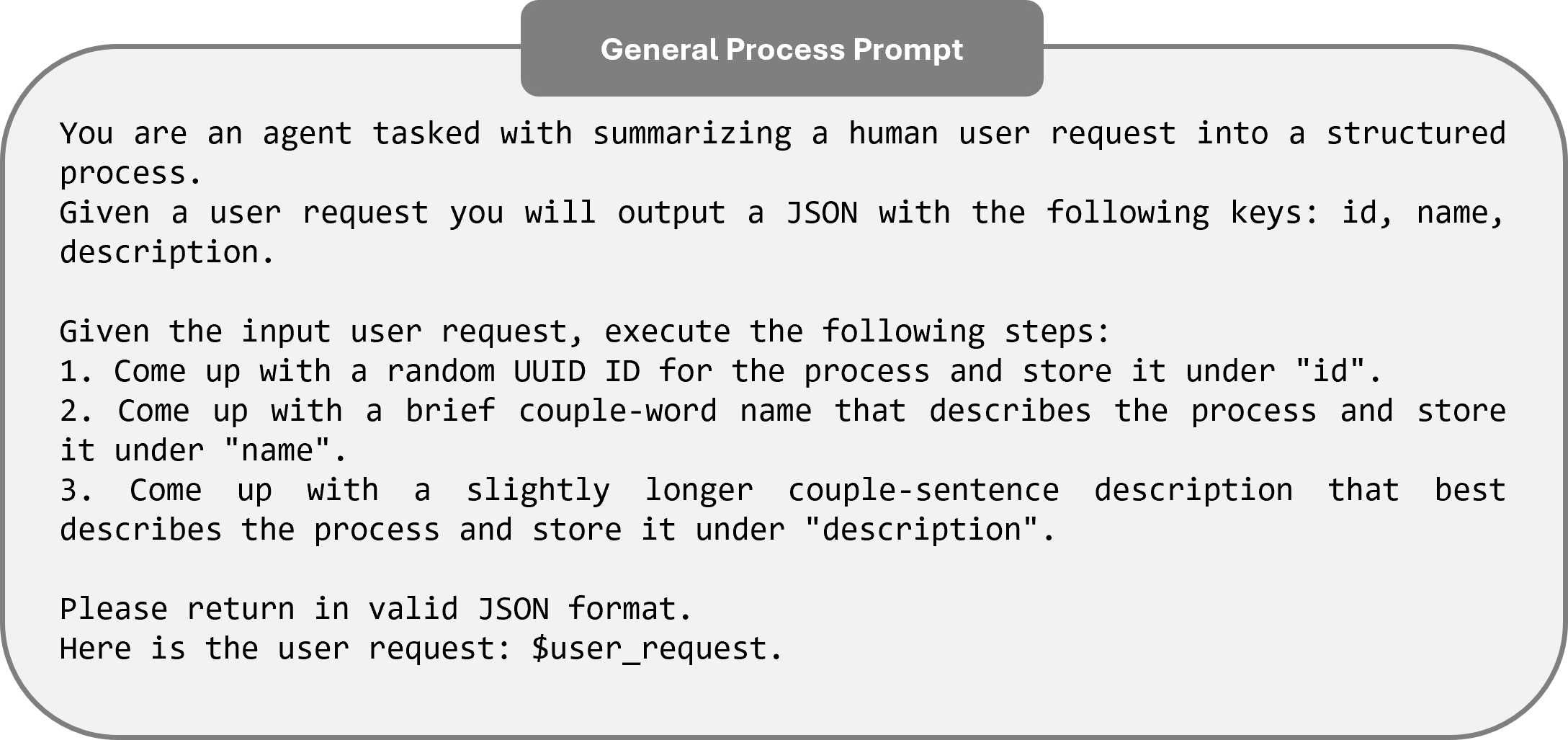}
    \caption{The \textit{General Process Prompt}.} 
    \label{fig:gen-process-prompt}
\end{figure}

\begin{figure}[H]
    \centering
    \includegraphics[width=\textwidth, height=\textheight, keepaspectratio]{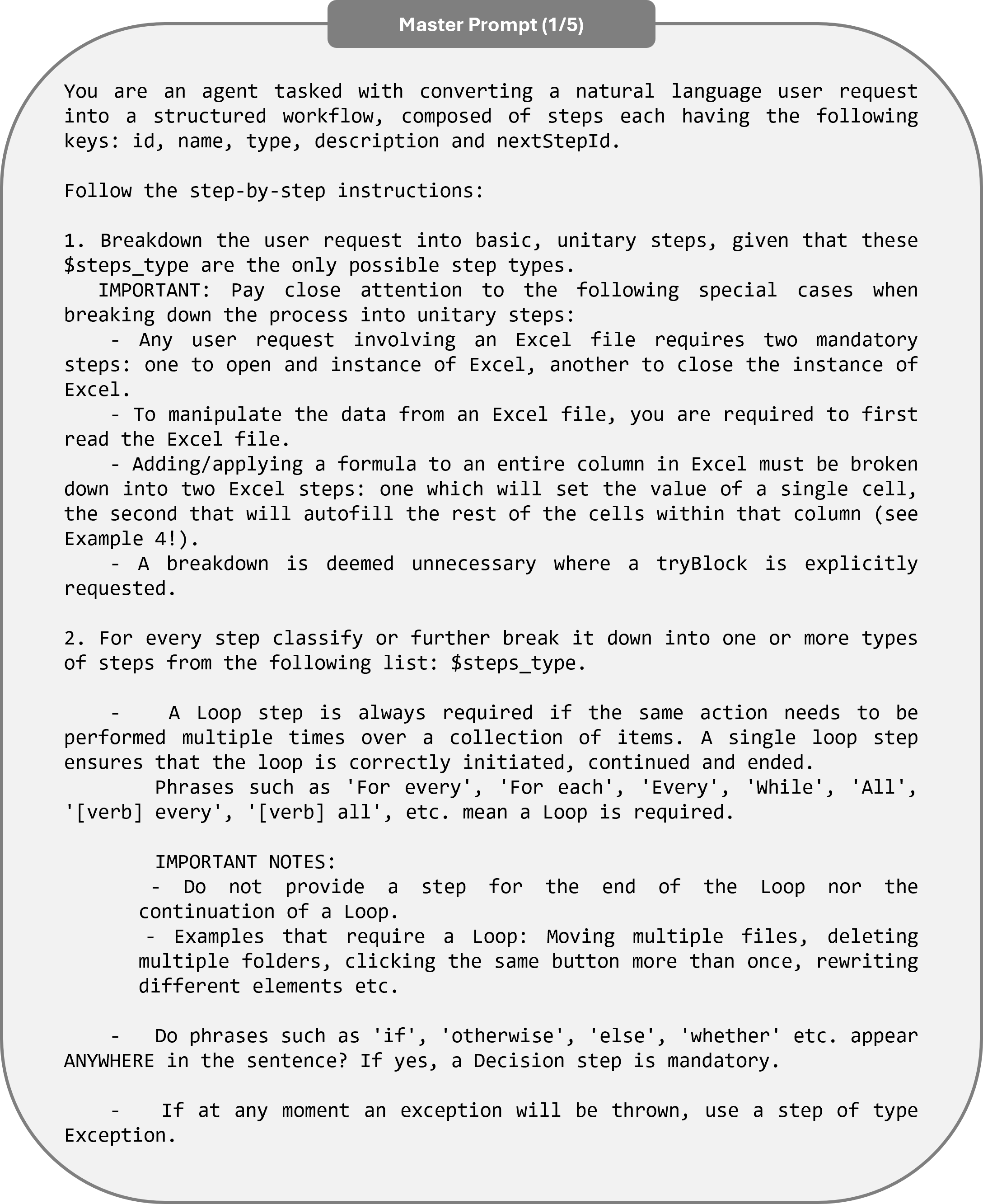} 
    \caption{\textit{Master Prompt} (Part 1 of 5).}
    \label{fig:master-prompt-1}
\end{figure}

\begin{figure}[H]
    \centering
    \includegraphics[width=\textwidth, height=\textheight, keepaspectratio]{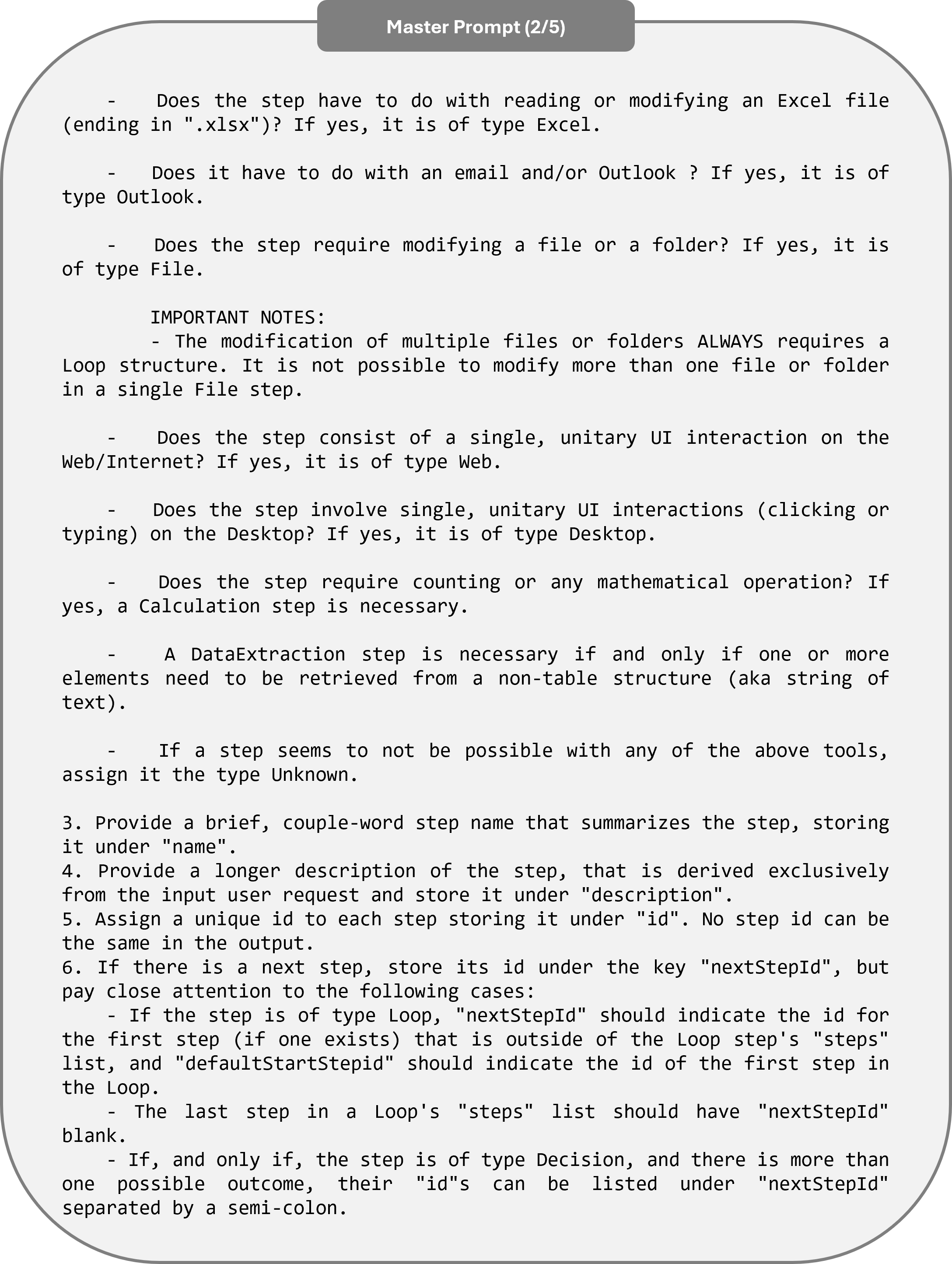}
    \caption{\textit{Master Prompt} (Part 2 of 5).}
    \label{fig:master-prompt-2}
\end{figure}

\begin{figure}[H]
    \centering
    \includegraphics[width=\textwidth, height=\textheight, keepaspectratio]{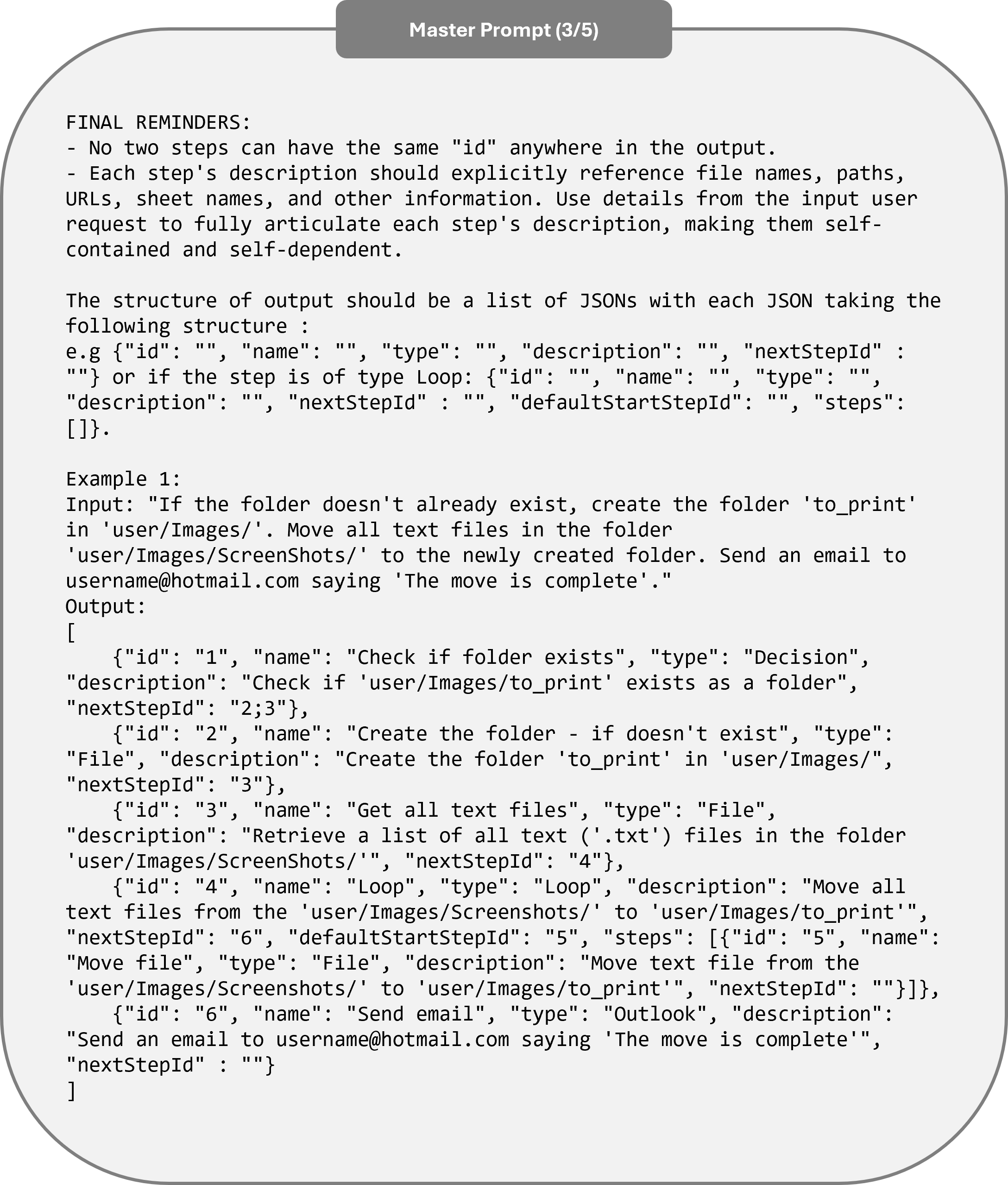}
    \caption{\textit{Master Prompt} (Part 3 of 5).}
    \label{fig:master-prompt-3}
\end{figure}

\begin{figure}[H]
    \centering
    \includegraphics[width=\textwidth, height=\textheight, keepaspectratio]{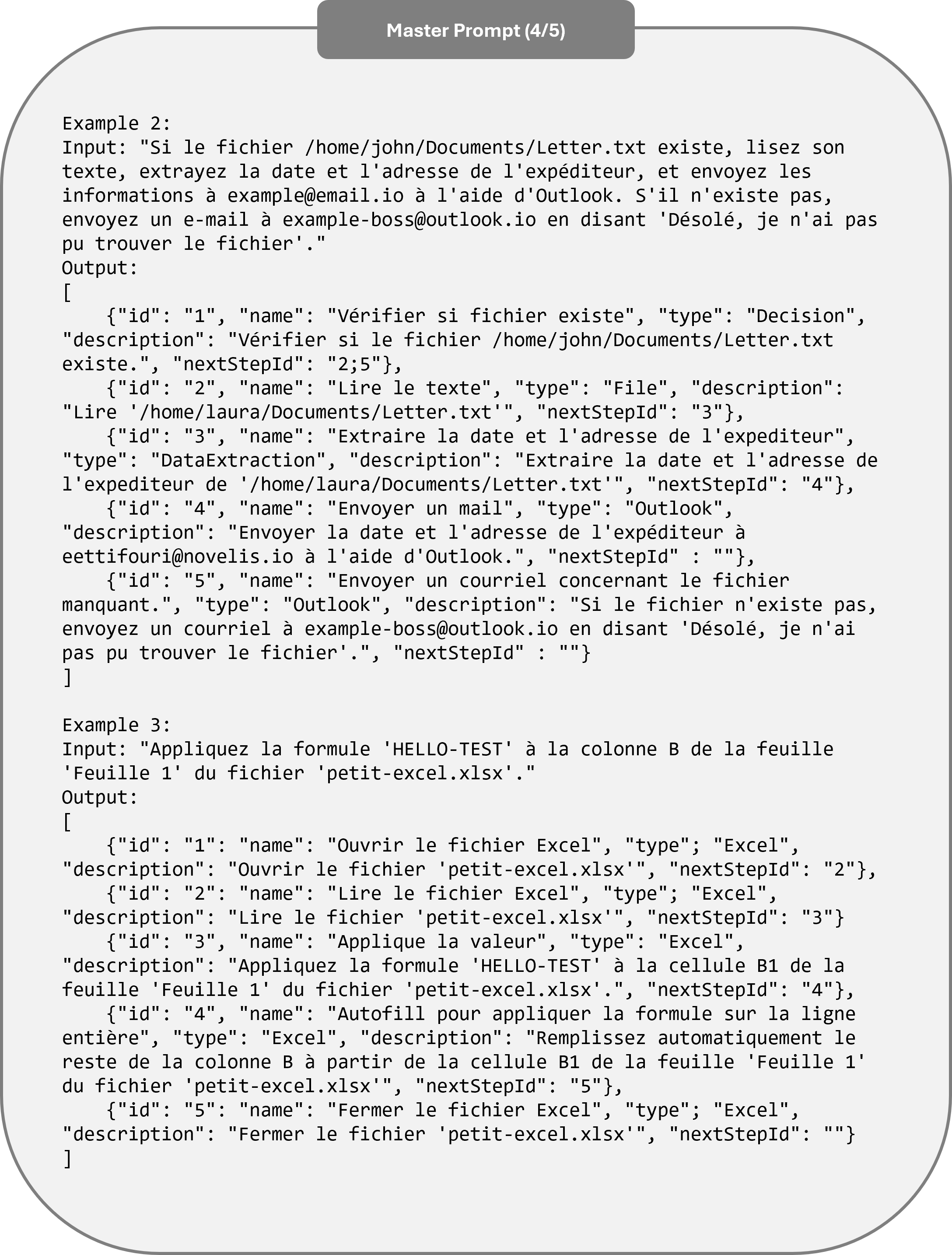}
    \caption{\textit{Master Prompt} (Part 4 of 5).}
    \label{fig:master-prompt-4}
\end{figure}

\begin{figure}[H]
    \centering
    \includegraphics[width=\textwidth, height=\textheight, keepaspectratio]{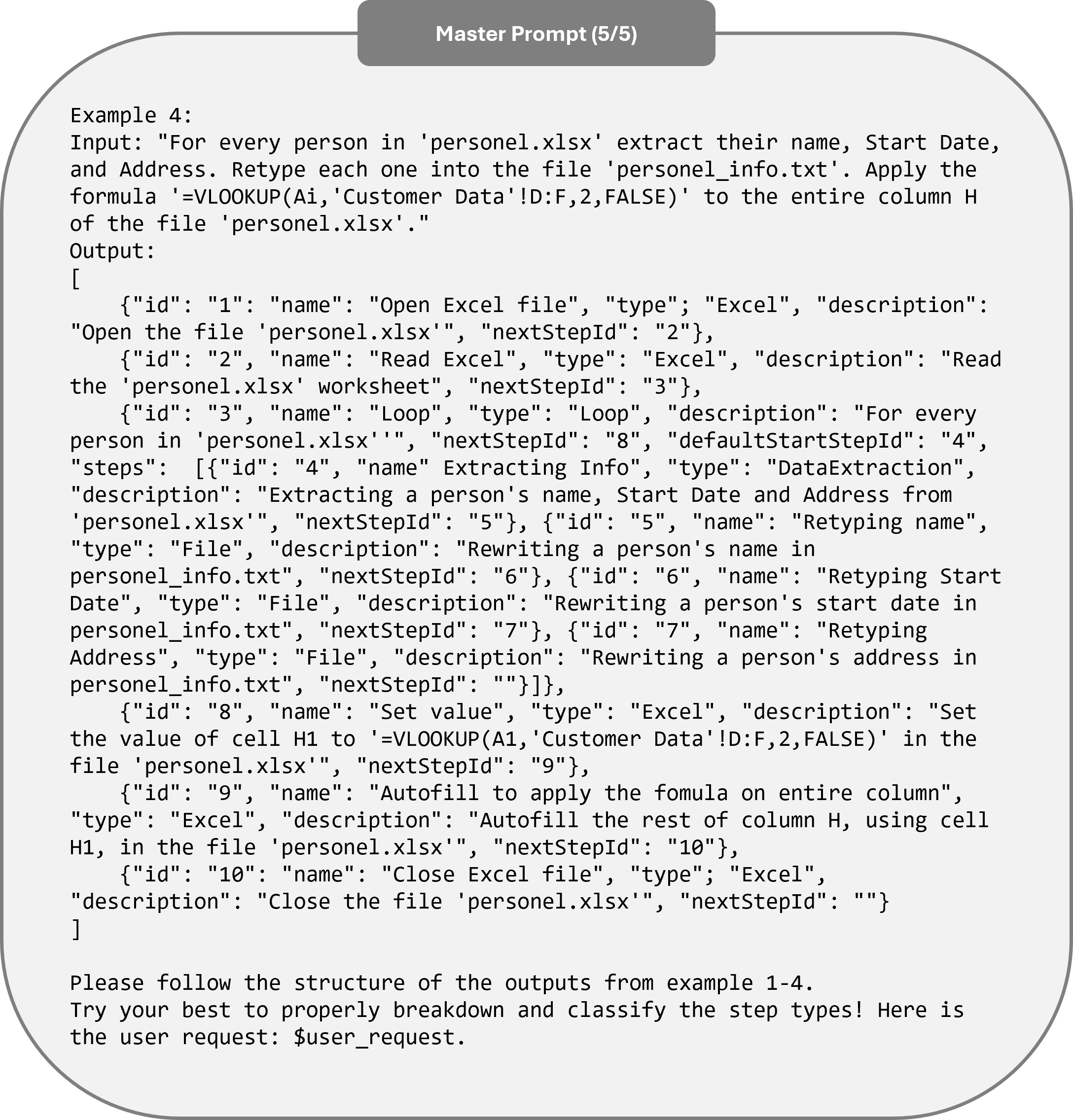}
    \caption{\textit{Master Prompt} (Part 5 of 5).}
    \label{fig:master-prompt-5}
\end{figure}

\begin{figure}[H]
    \centering
    \includegraphics[width=\textwidth, height=\textheight, keepaspectratio]{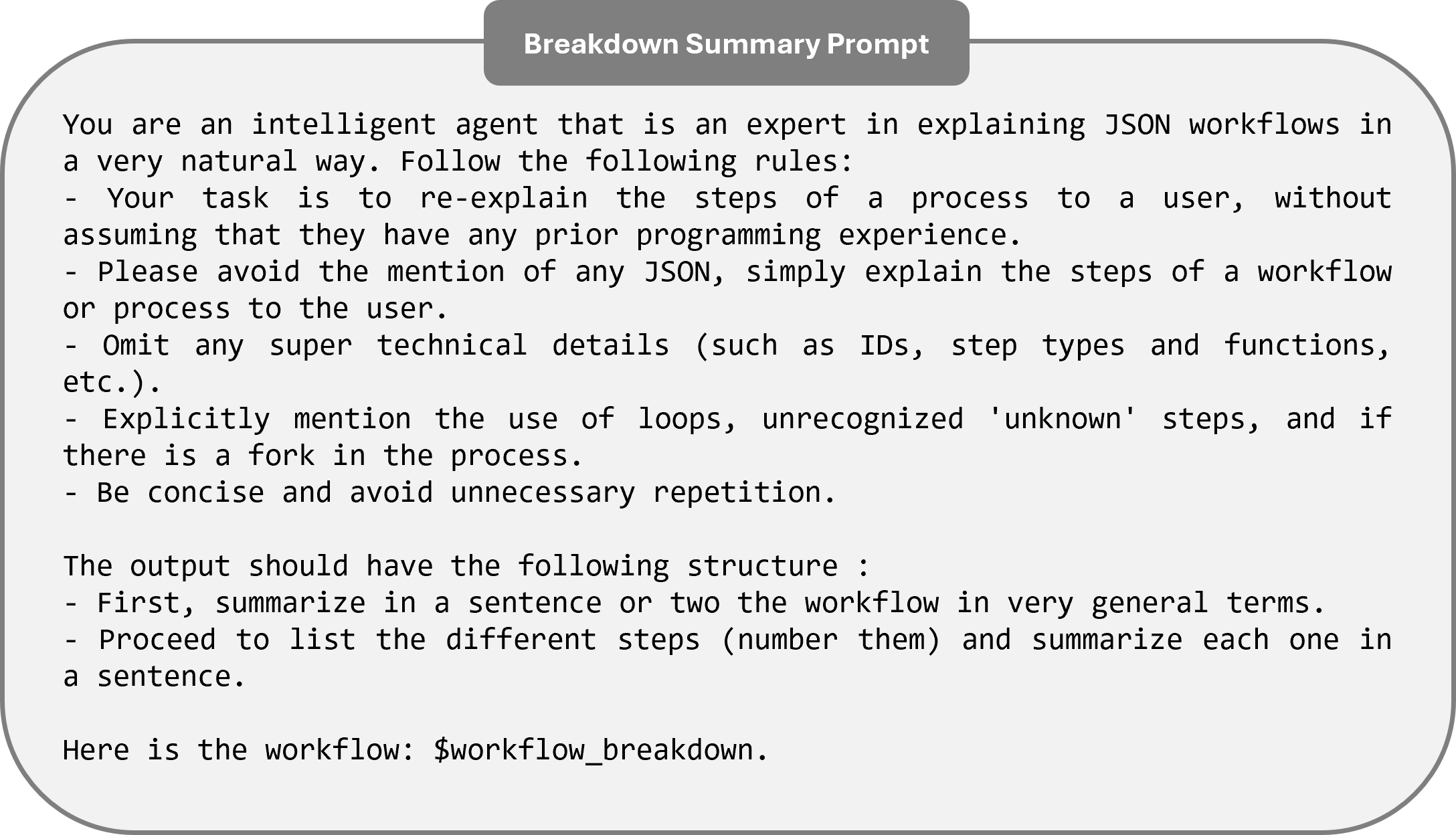}
    \caption{The \textit{High-level Breakdown Summary Prompt}.}
    \label{fig:brkdwn-summary}
\end{figure}

\begin{figure}[H]
    \centering
    \includegraphics[width=\textwidth, height=\textheight, keepaspectratio]{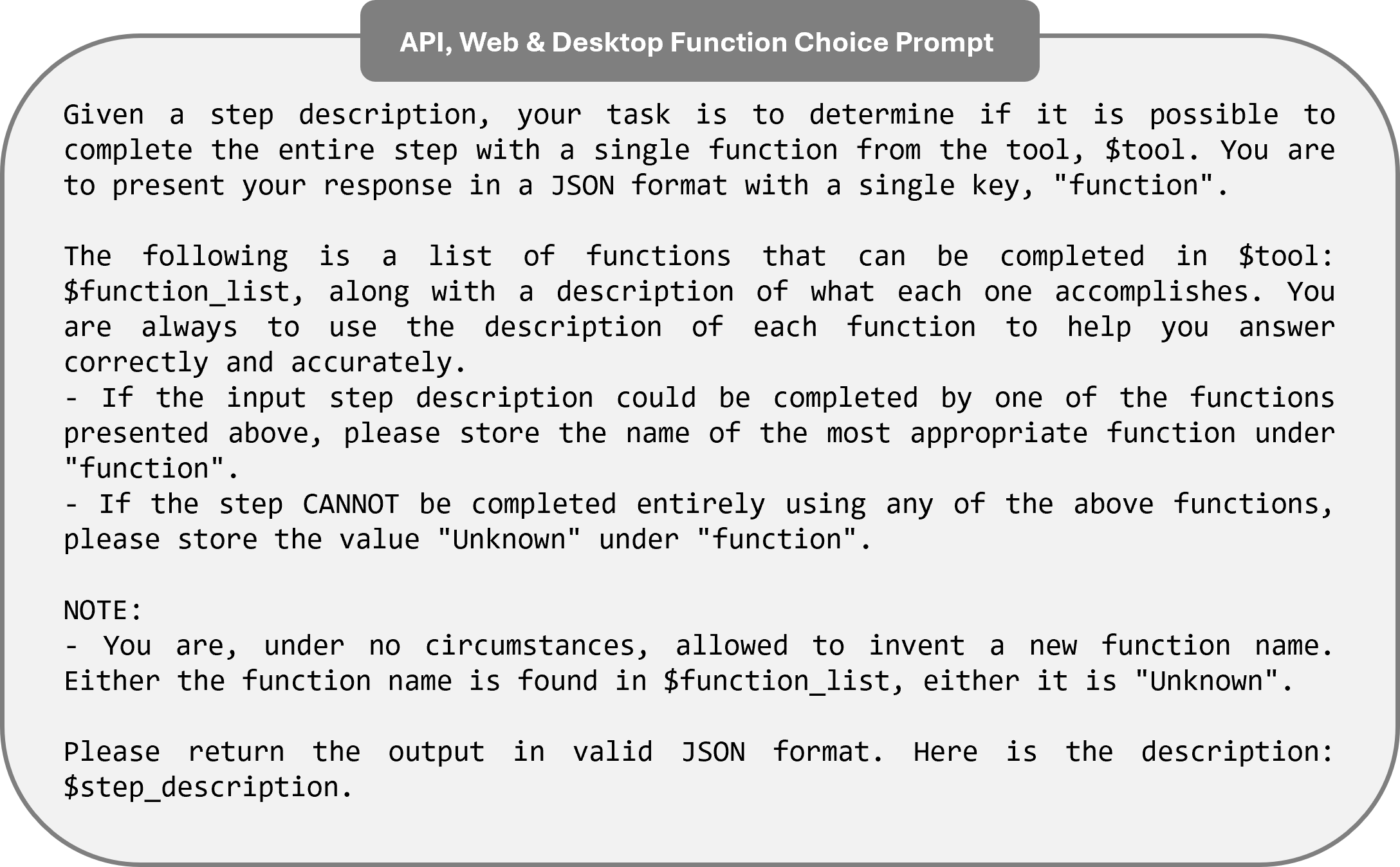}
    \caption{The \textit{API, Web and Desktop function/actionType choice Prompt}.}
    \label{fig:api-function-prompt}
\end{figure}

\begin{figure}[H]
    \centering
    \includegraphics[width=\textwidth, height=\textheight, keepaspectratio]{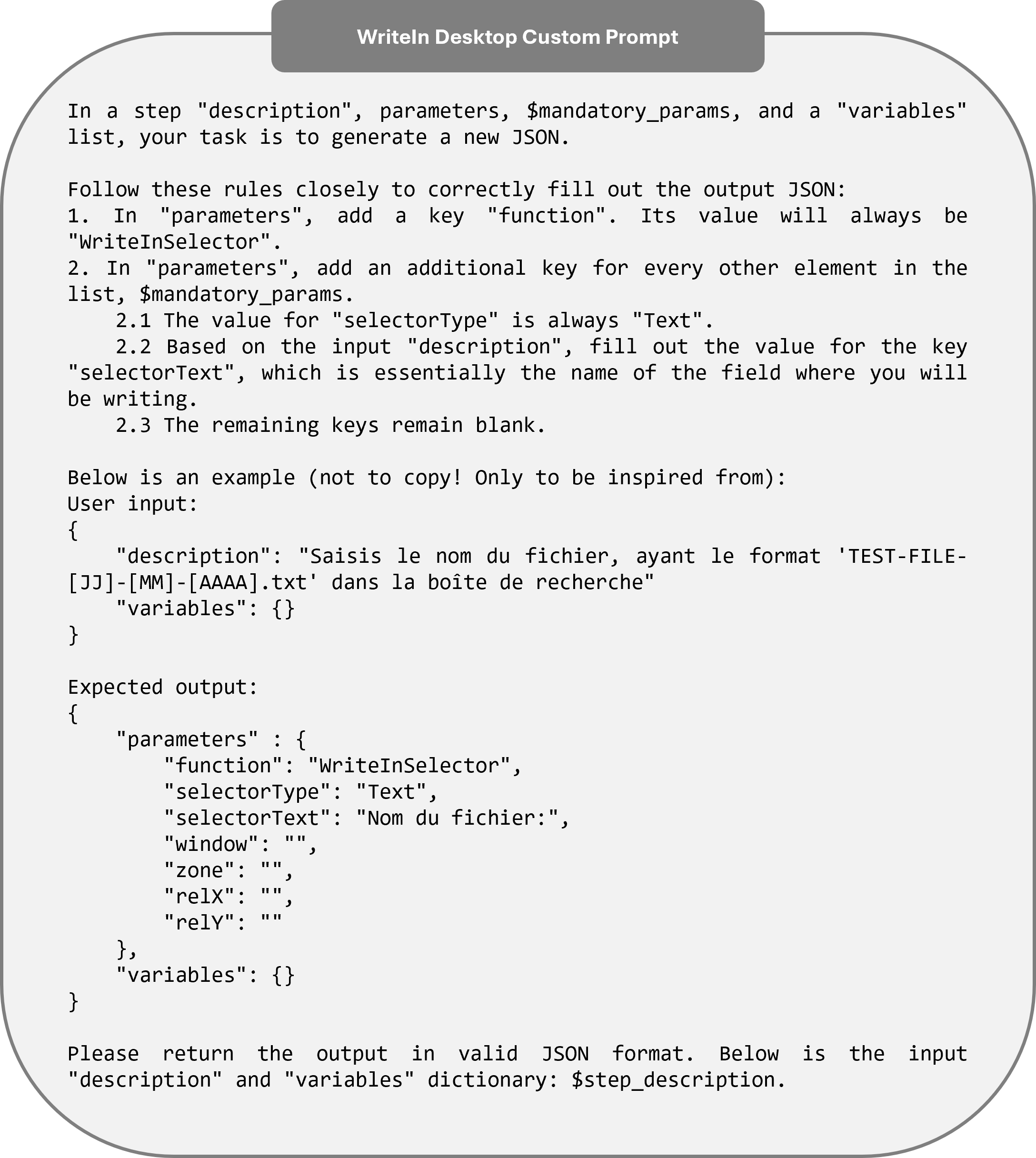}
    \caption{The \textit{Custom WriteIn Parameter Prompt}.}
    \label{fig:writein-params}
\end{figure}

\begin{figure}[H]
    \centering
    \includegraphics[width=\textwidth, height=\textheight, keepaspectratio]{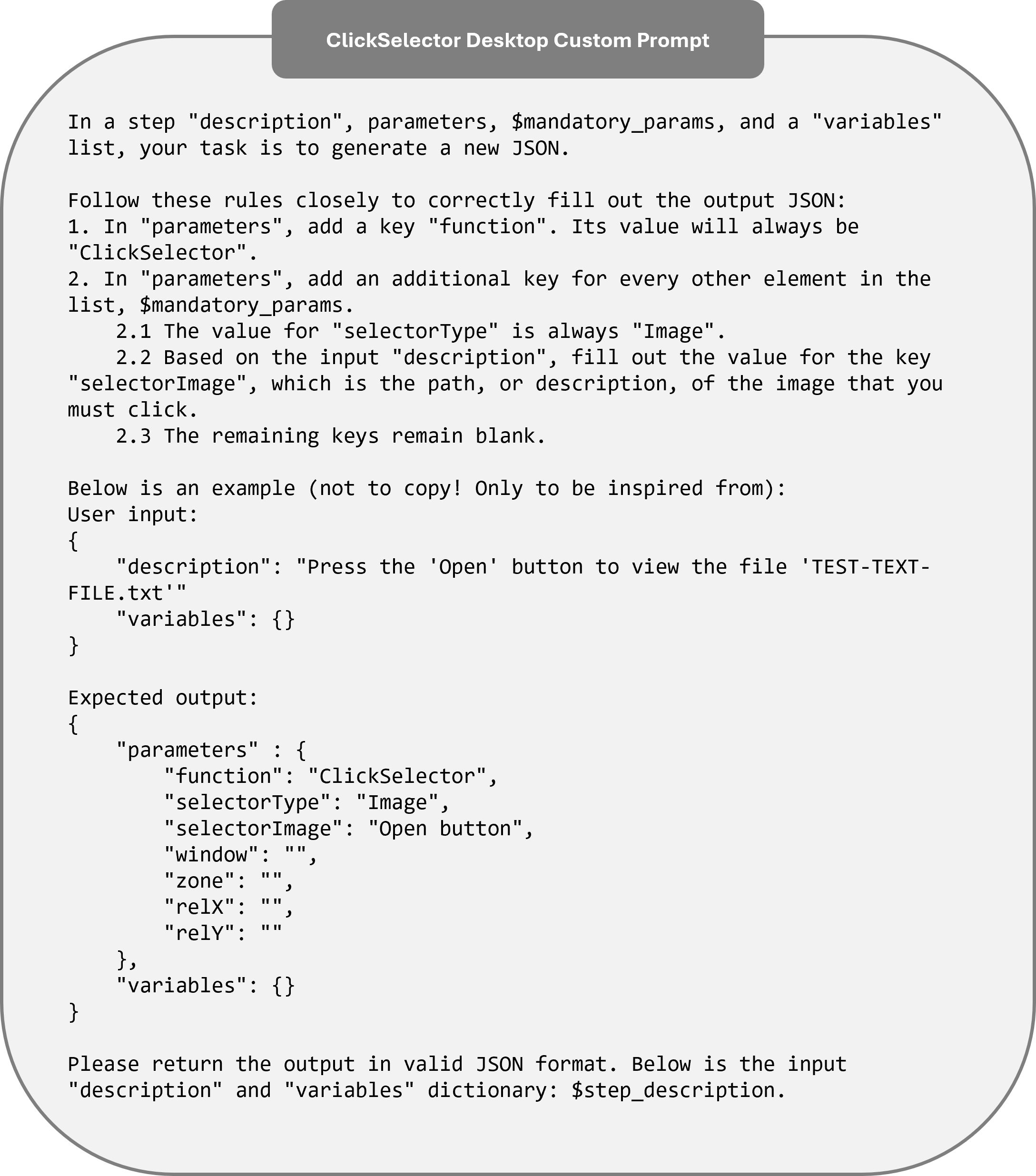}
    \caption{The \textit{Custom ClickSelector Parameter Prompt}.}
    \label{fig:click-select-params}
\end{figure}

\begin{figure}[H]
    \centering
    \includegraphics[width=0.95\textwidth, height=0.95\textheight, keepaspectratio]{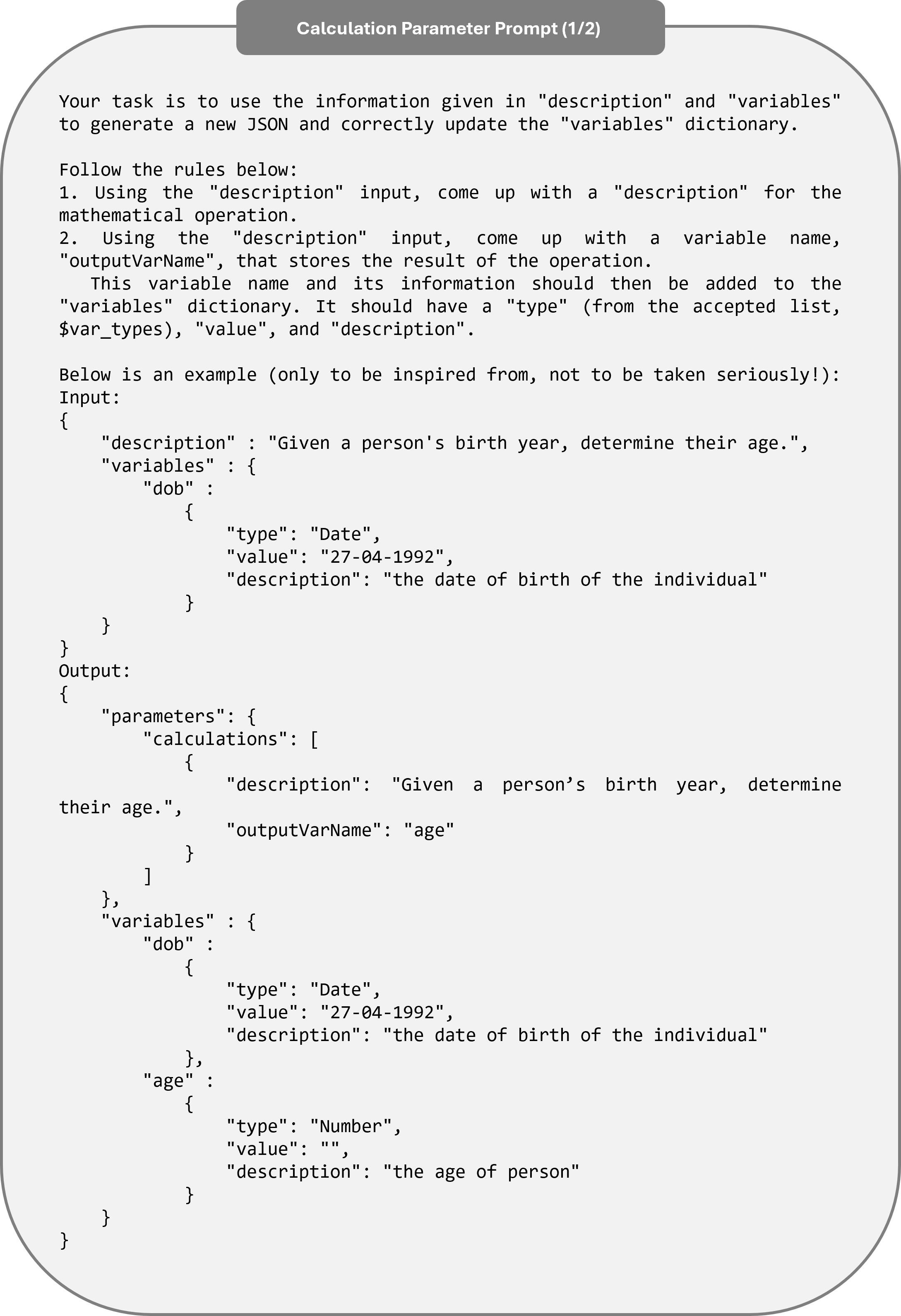}
    \caption{The \textit{Calculation Parameter Prompt} (Part 1 of 2).}
    \label{fig:calculation-params-1}
\end{figure}

\begin{figure}[H]
    \centering
    \includegraphics[width=\textwidth, height=\textheight, keepaspectratio]{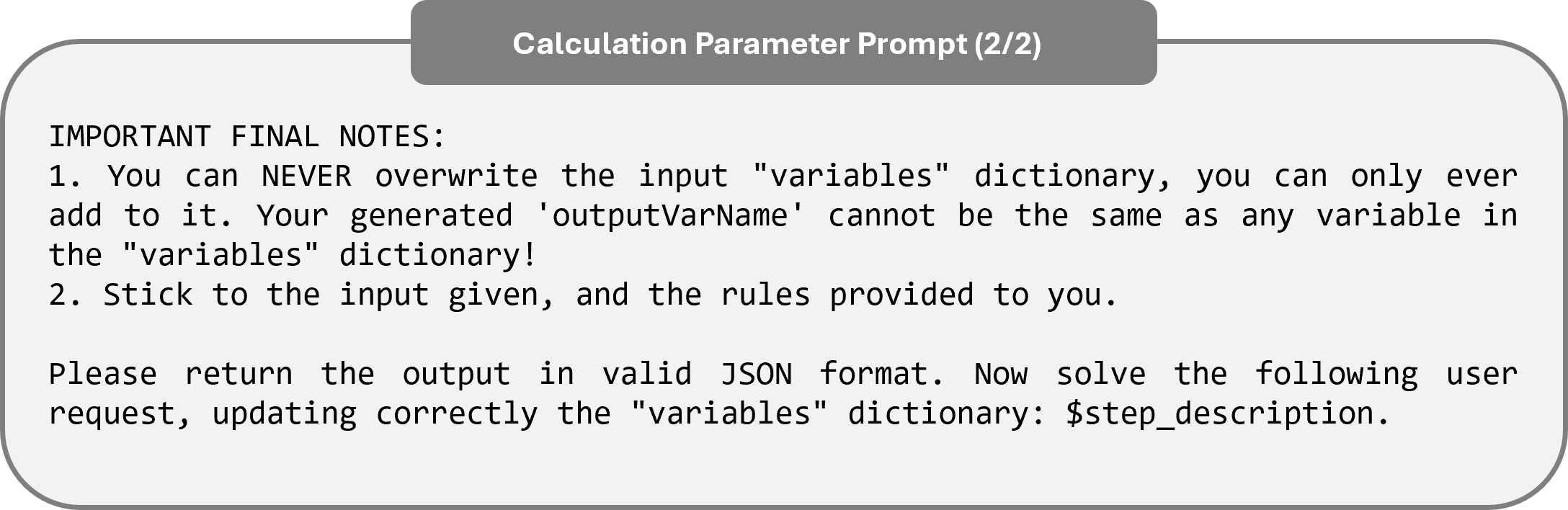}
    \caption{The \textit{Calculation Parameter Prompt} (Part 2 of 2).}
    \label{fig:calculation-params-2}
\end{figure}

\begin{figure}[H]
    \centering
    \includegraphics[width=\textwidth, height=\textheight, keepaspectratio]{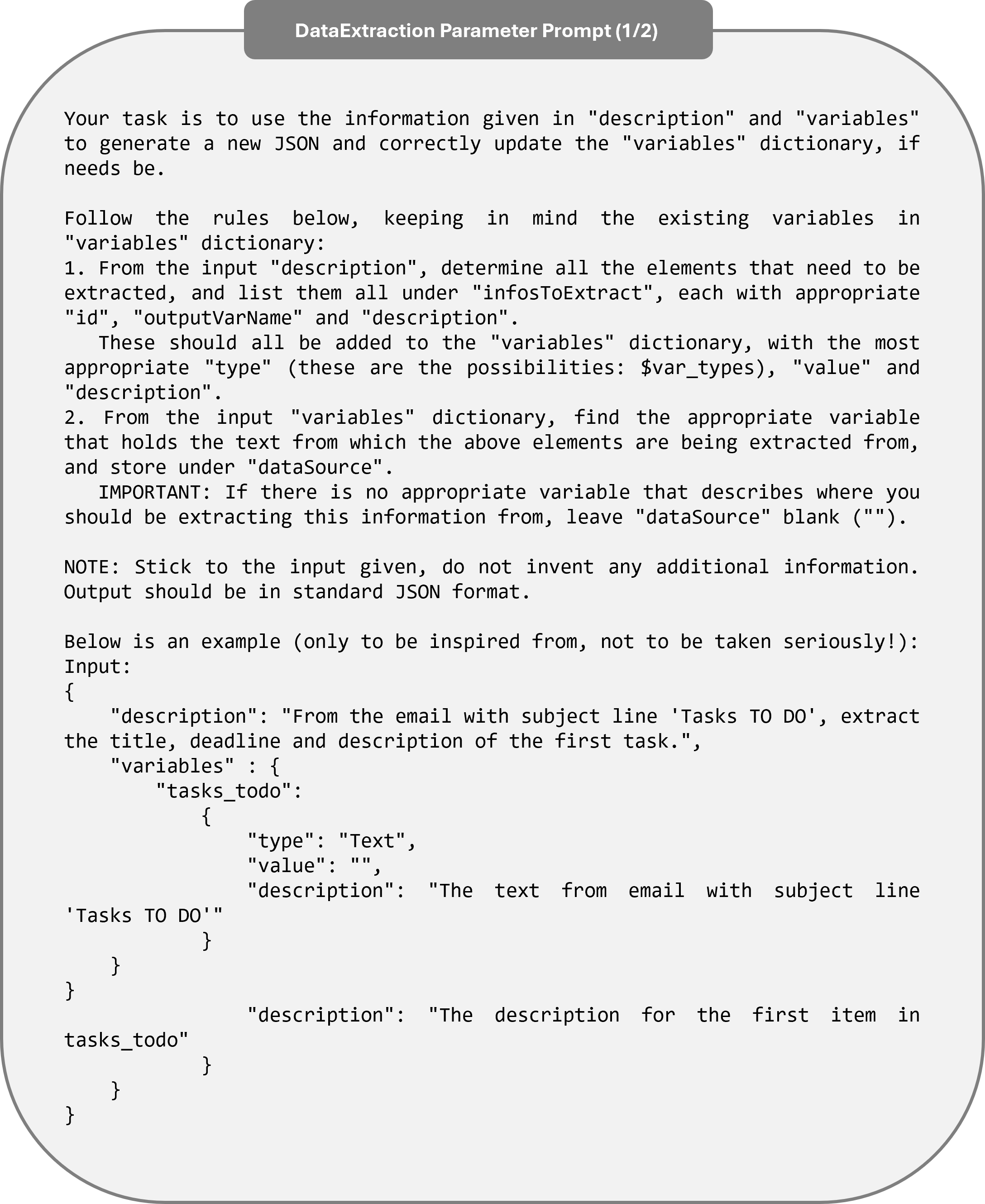}
    \caption{The \textit{DataExtraction Parameter Prompt} (Part 1 of 2).}
    \label{fig:data-extract-params-1}
\end{figure}

\begin{figure}[H]
    \centering
    \includegraphics[width=0.95\textwidth, height=0.95\textheight, keepaspectratio]{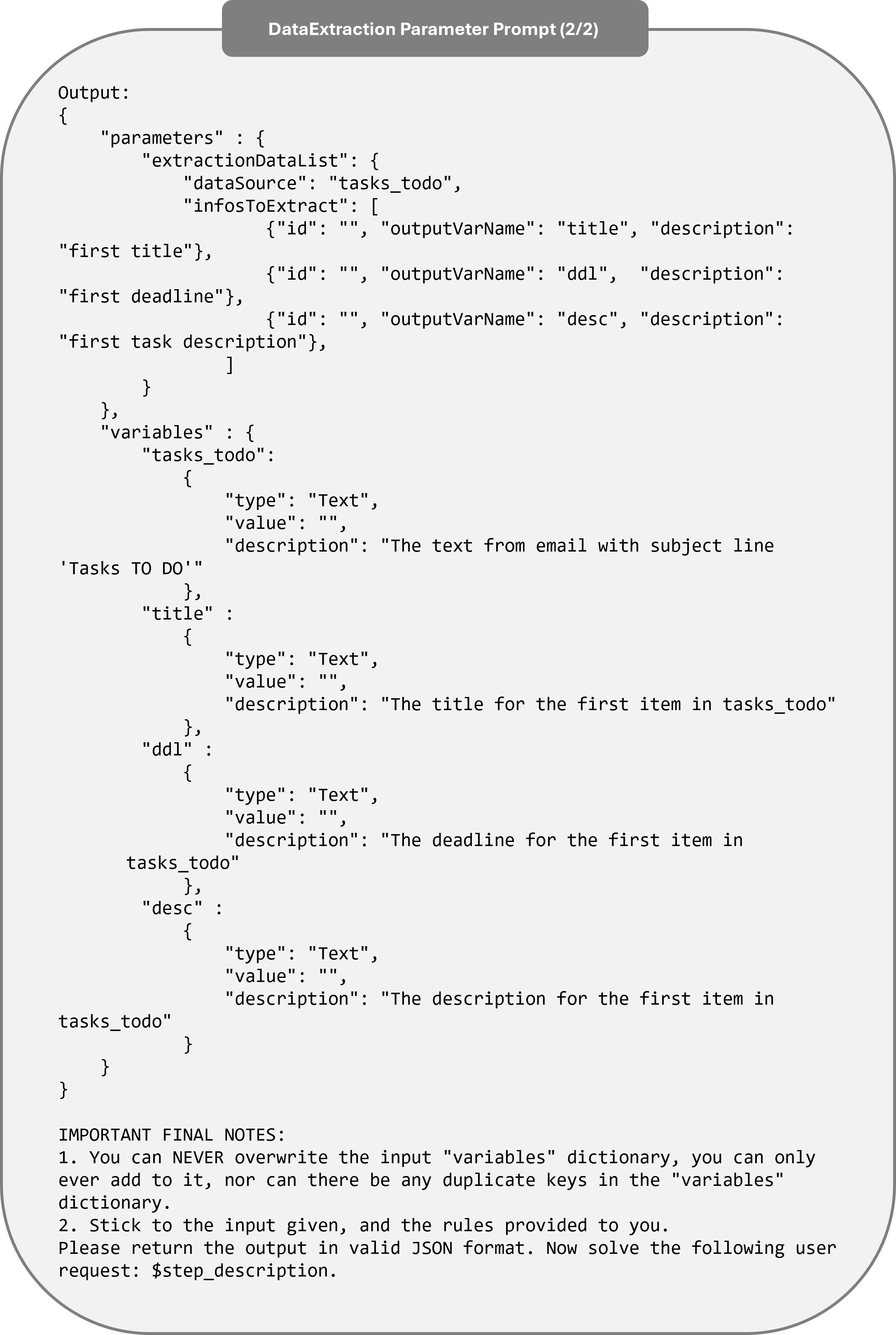}
    \caption{The \textit{DataExtraction Parameter Prompt} (Part 2 of 2).}
    \label{fig:data-extract-params-2}
\end{figure}

\begin{figure}[H]
    \centering
    \includegraphics[width=\textwidth, height=\textheight, keepaspectratio]{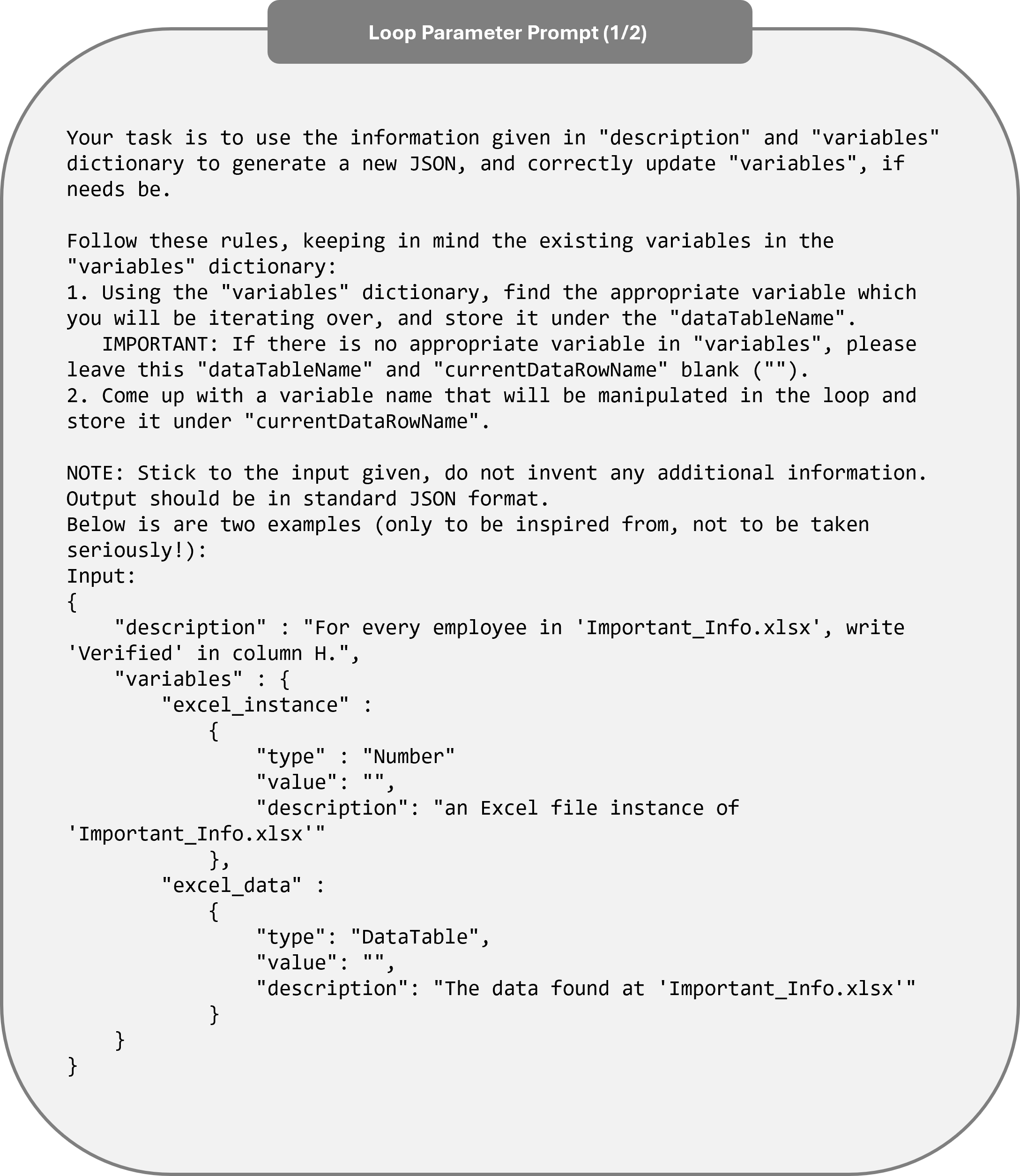}
    \caption{The \textit{Loop Parameter Prompt} (Part 1 of 2).}
    \label{fig:loop-params-1}
\end{figure}

\begin{figure}[H]
    \centering
    \includegraphics[width=0.95\textwidth, height=0.95\textheight, keepaspectratio]{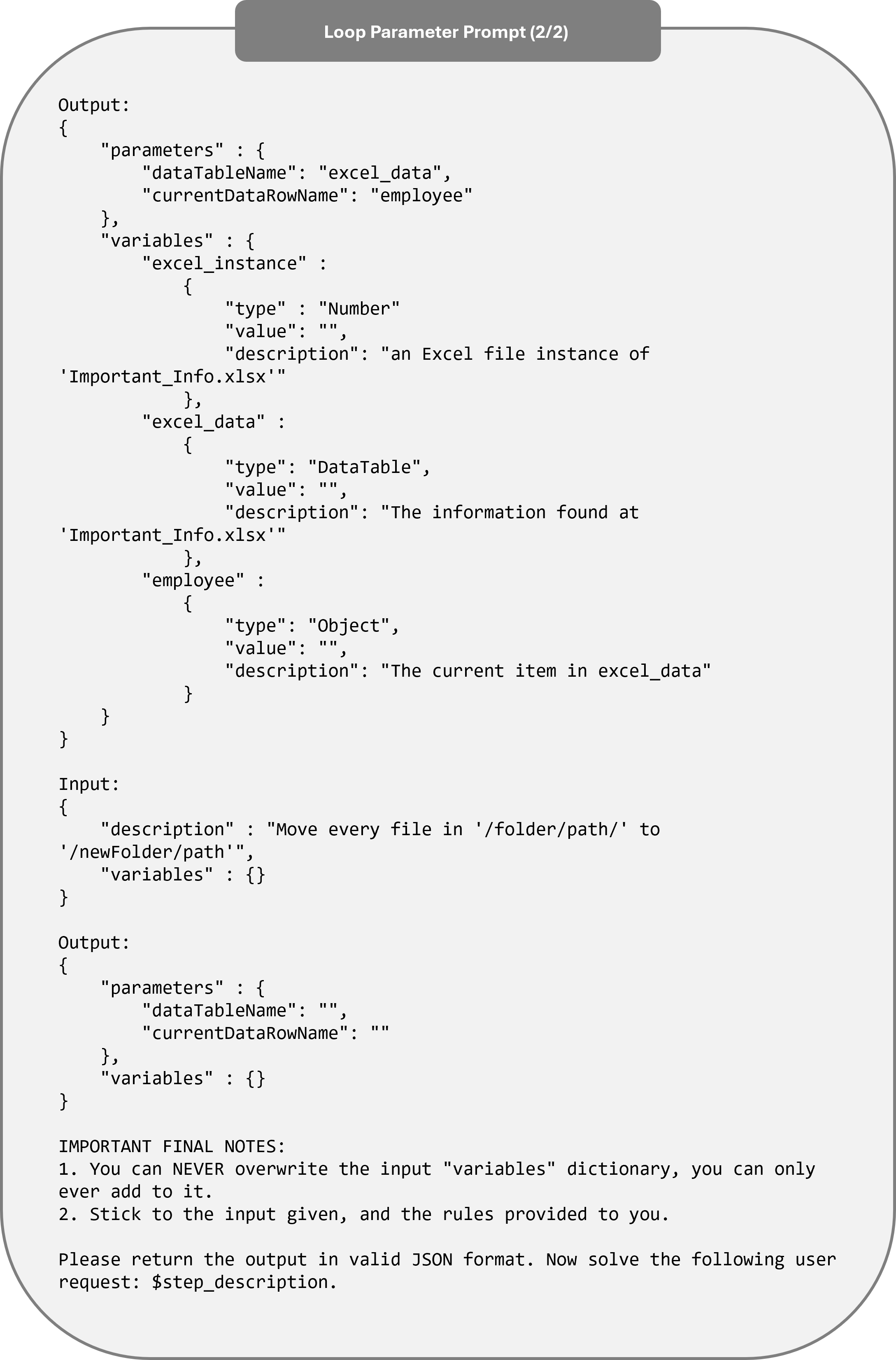}
    \caption{The \textit{Loop Parameter Prompt} (Part 2 of 2).}
    \label{fig:loop-params-2}
\end{figure}

\begin{figure}[H]
    \centering
    \includegraphics[width=\textwidth, height=\textheight, keepaspectratio]{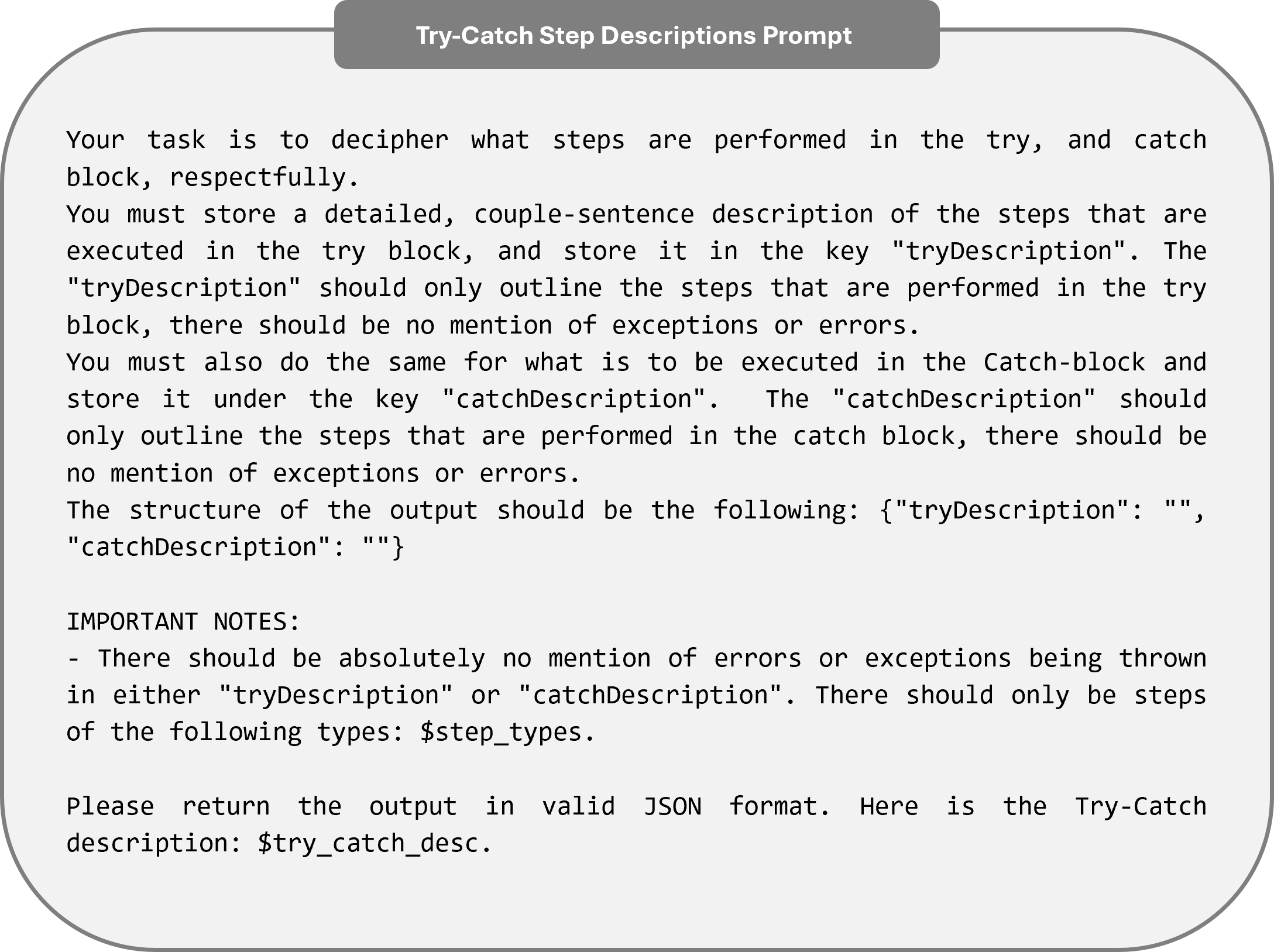}
    \caption{The \textit{Try-Catch Description Prompt}.}
    \label{fig:trycatch-desc}
\end{figure}

\begin{figure}[H]
    \centering
    \includegraphics[width=0.95\textwidth, height=0.95\textheight, keepaspectratio]{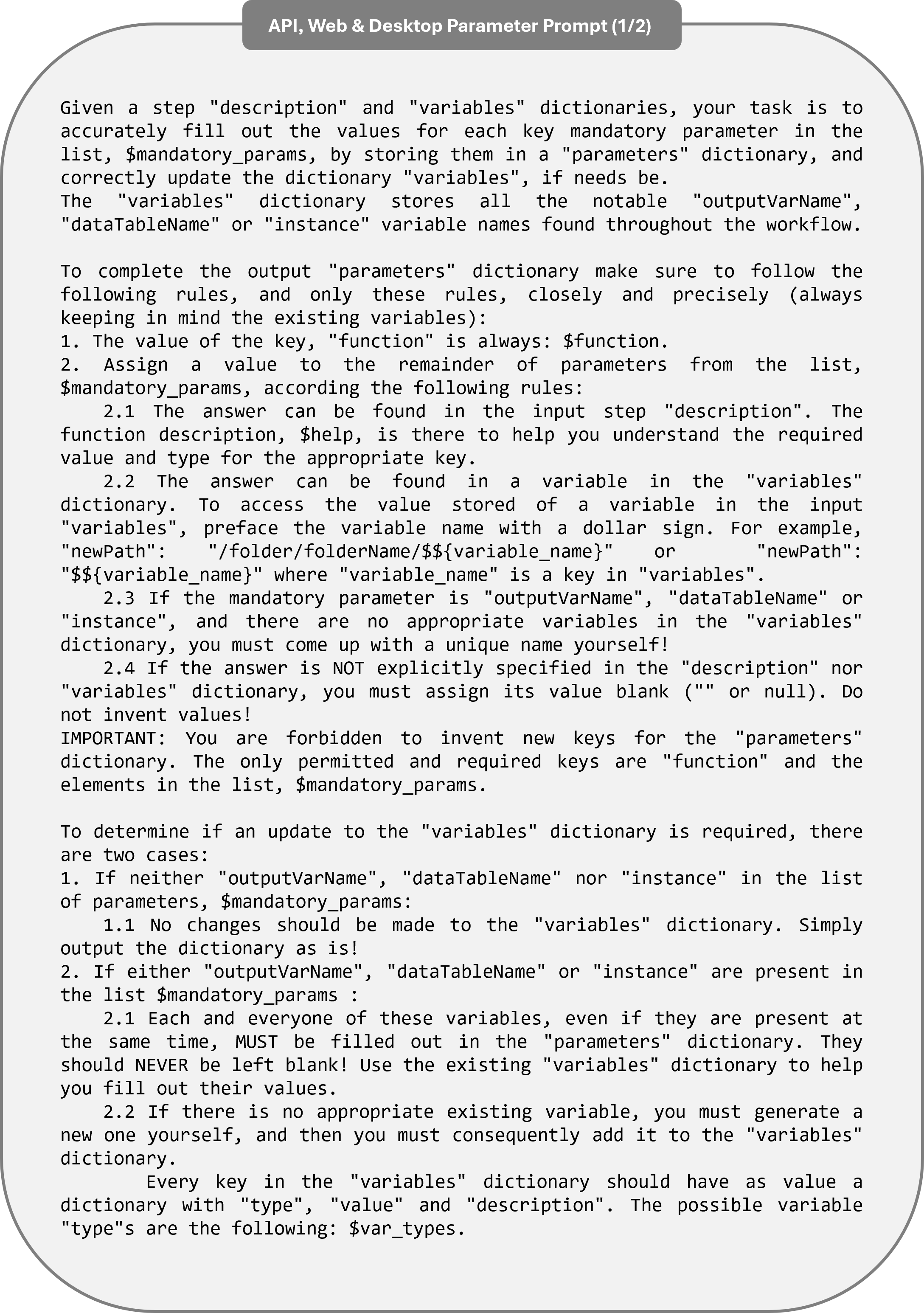}
    \caption{The \textit{API, Web and Desktop Parameters Prompt} (Part 1 of 2).}
    \label{fig:api-params-1}
\end{figure}

\begin{figure}[H]
    \centering
    \includegraphics[width=0.95\textwidth, height=0.95\textheight, keepaspectratio]{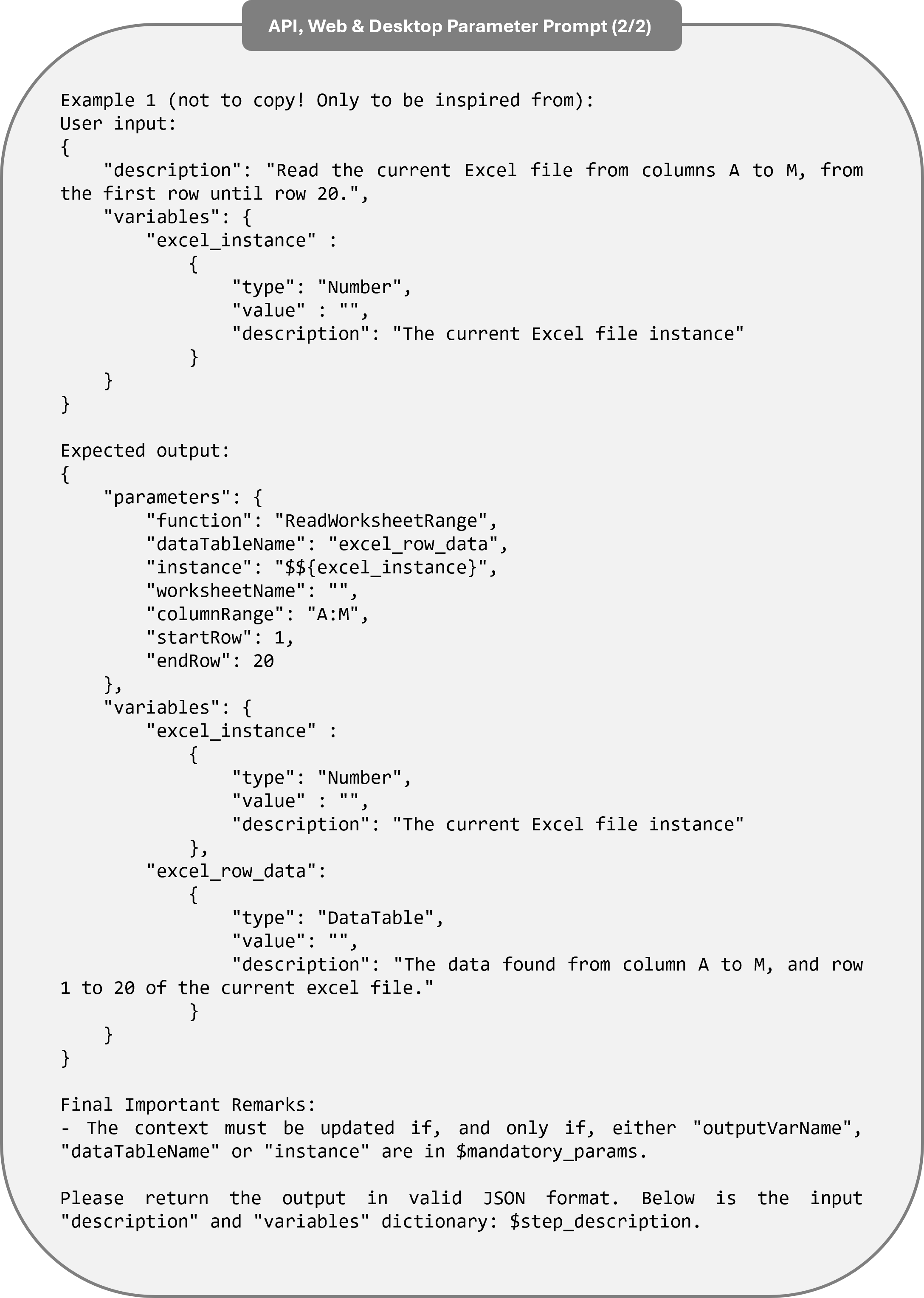}
    \caption{The \textit{API, Web and Desktop Parameters Prompt} (Part 2 of 2).}
    \label{fig:api-params-2}
\end{figure}

\begin{figure}[H]
    \centering
    \includegraphics[width=0.95\textwidth, height=0.95\textheight, keepaspectratio]{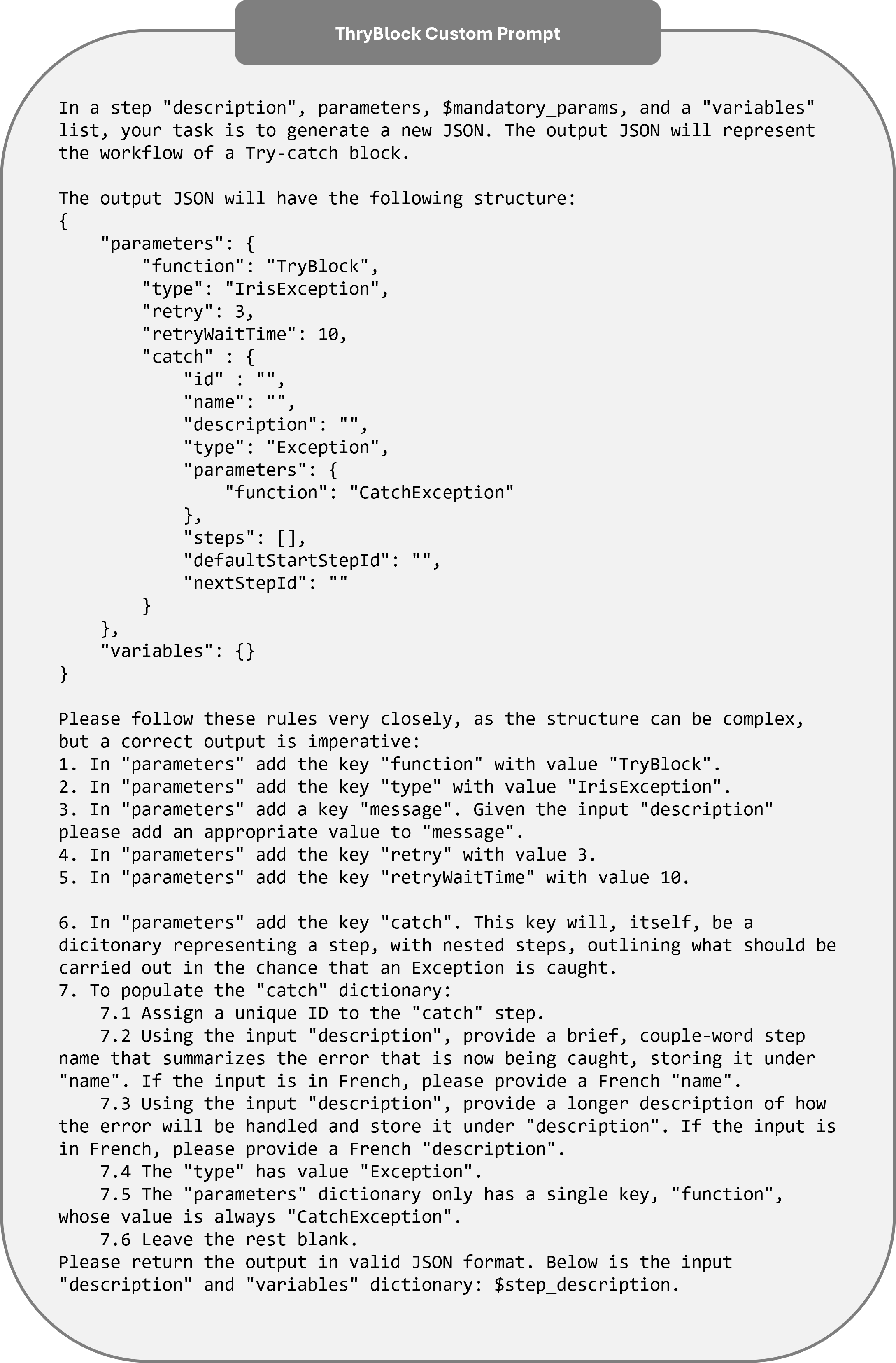}
    \caption{The \textit{Try-Block Parameter Prompt}.}
    \label{fig:tryblock-params}
\end{figure}

\begin{figure}[H]
    \centering
    \includegraphics[width=\textwidth, height=\textheight, keepaspectratio]{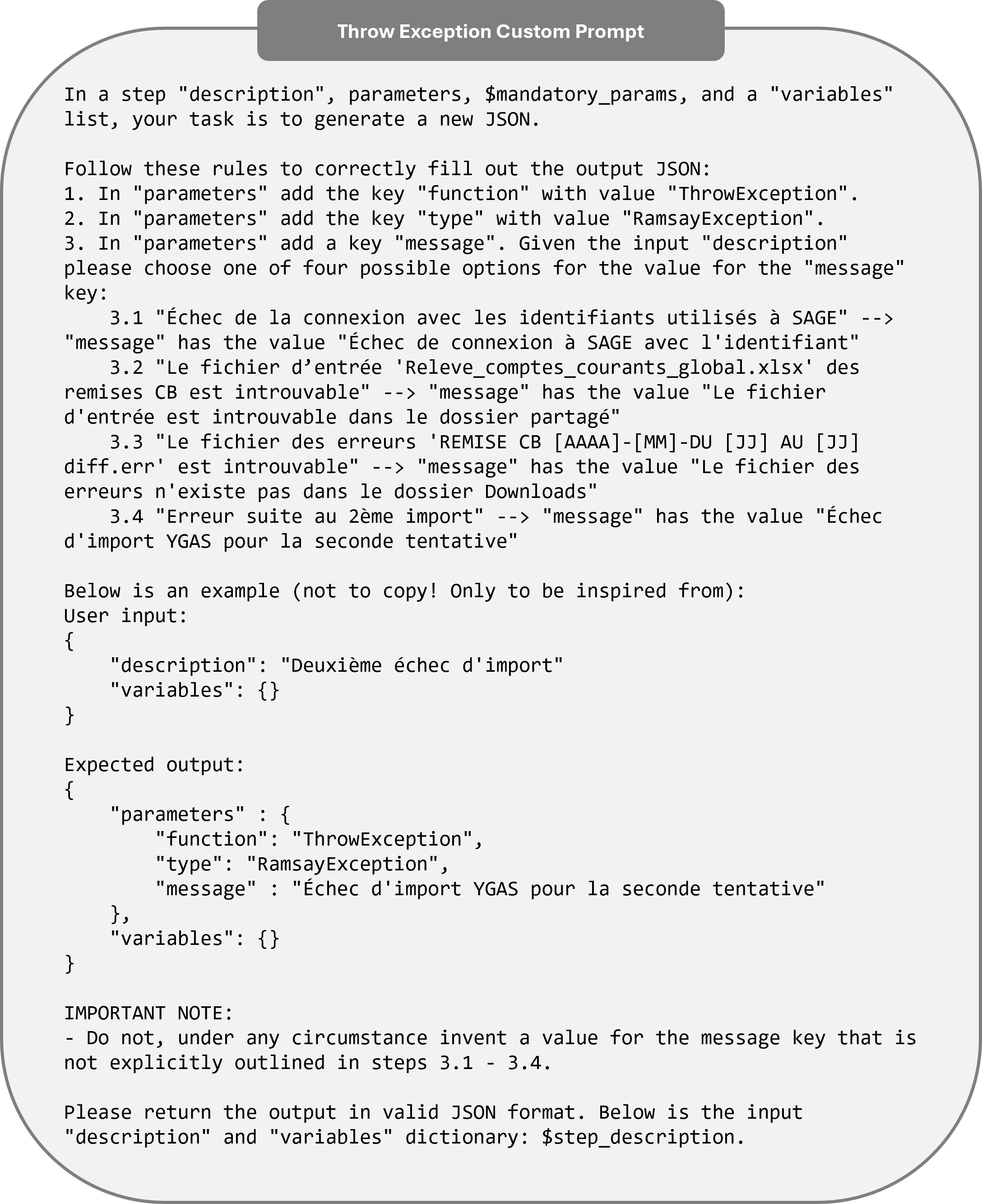}
    \caption{The \textit{Custom ThrowException Parameter Prompt}.}
    \label{fig:throw-except-params}
\end{figure}

\begin{figure}[H]
    \centering
    \includegraphics[width=\textwidth, height=\textheight, keepaspectratio]{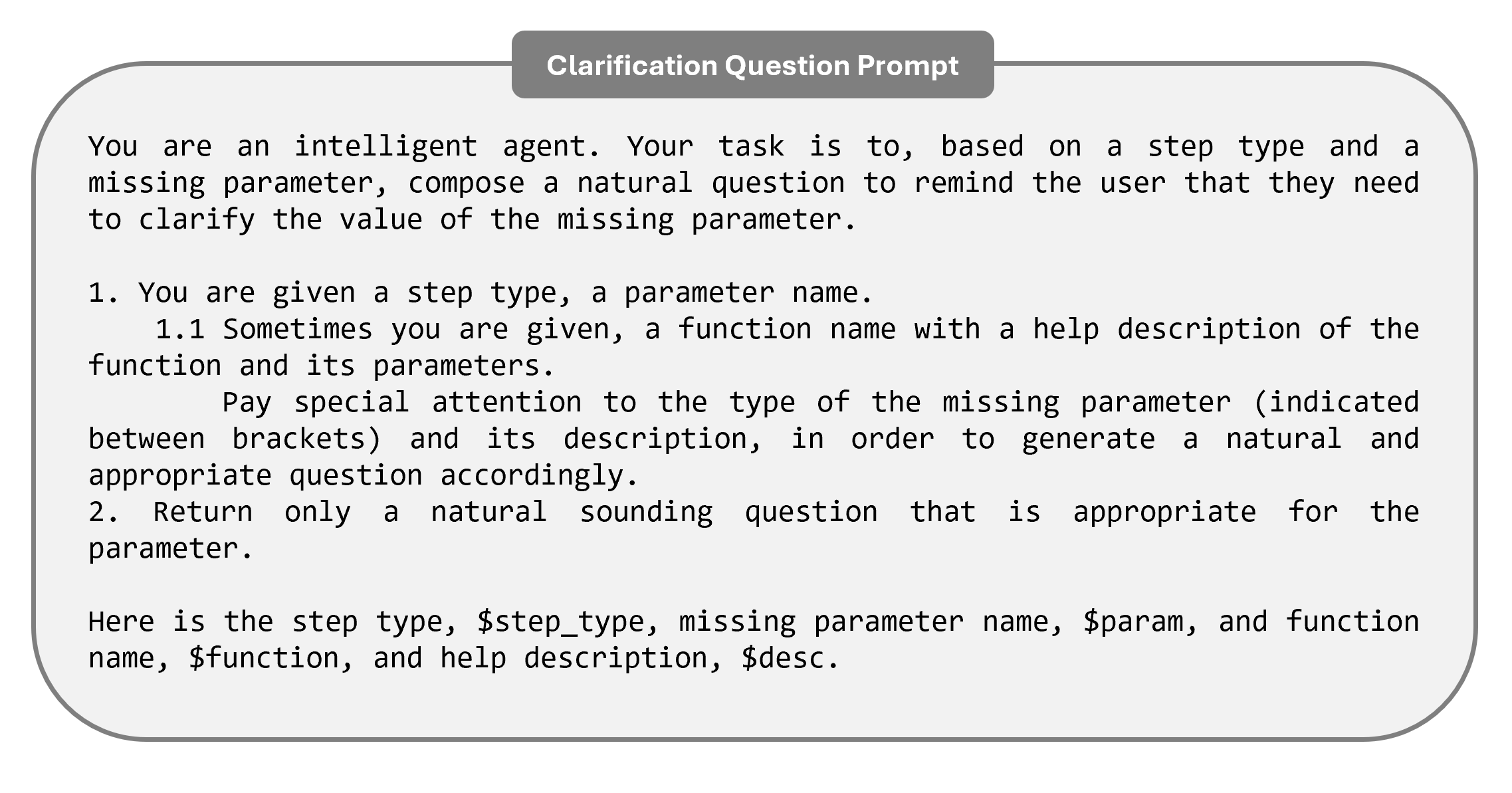}
    \caption{The \textit{Questions Prompt}.}
    \label{fig:questions-prompt}
\end{figure}

\begin{figure}[H]
    \centering
    \includegraphics[width=\textwidth, height=\textheight, keepaspectratio]{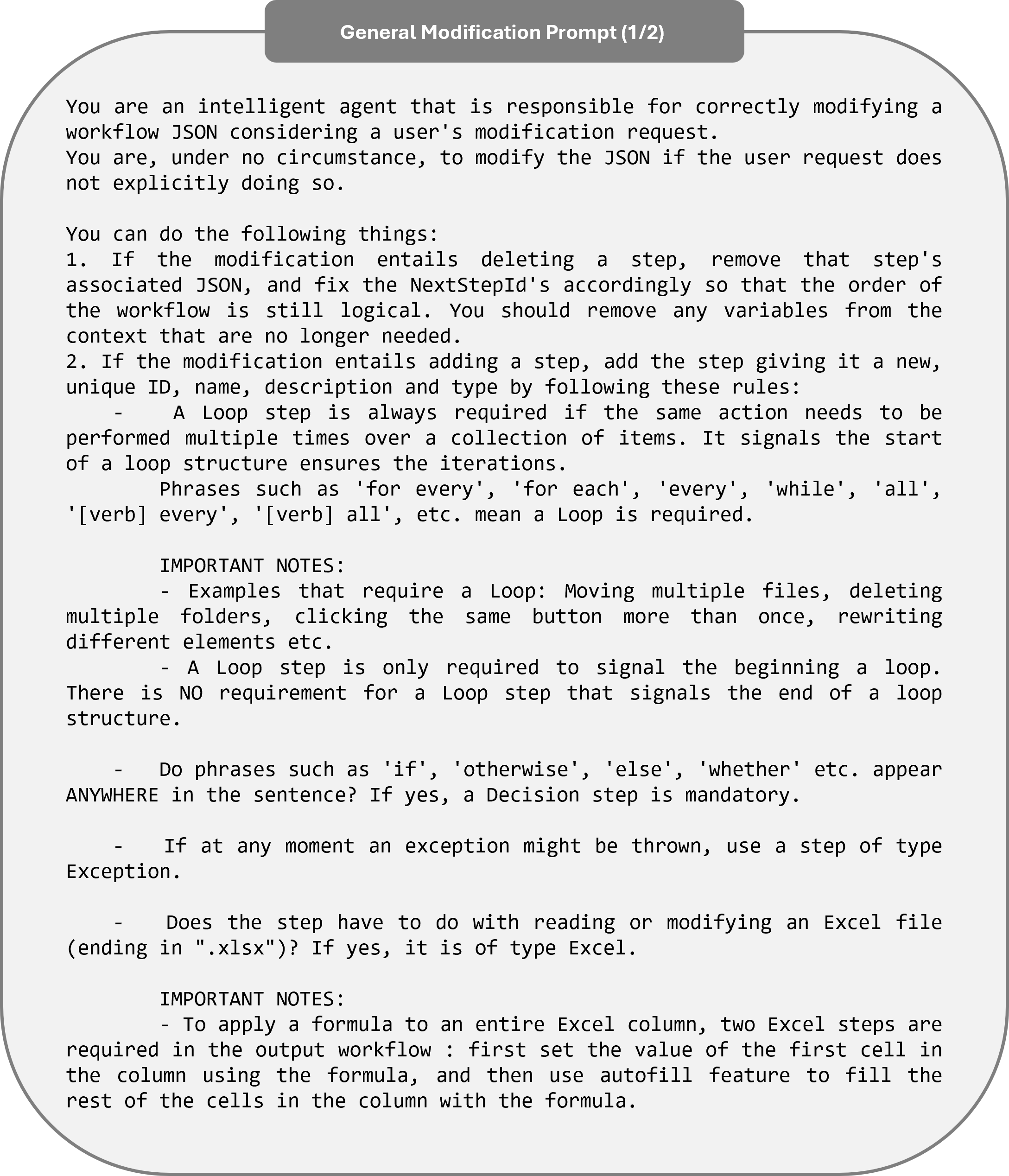}
    \caption{The \textit{Workflow Modification Prompt} (Part 1 of 2).}
    \label{fig:workflow-mod-prompt-1}
\end{figure}

\begin{figure}[H]
    \centering
    \includegraphics[width=0.9\textwidth, height=0.9\textheight, keepaspectratio]{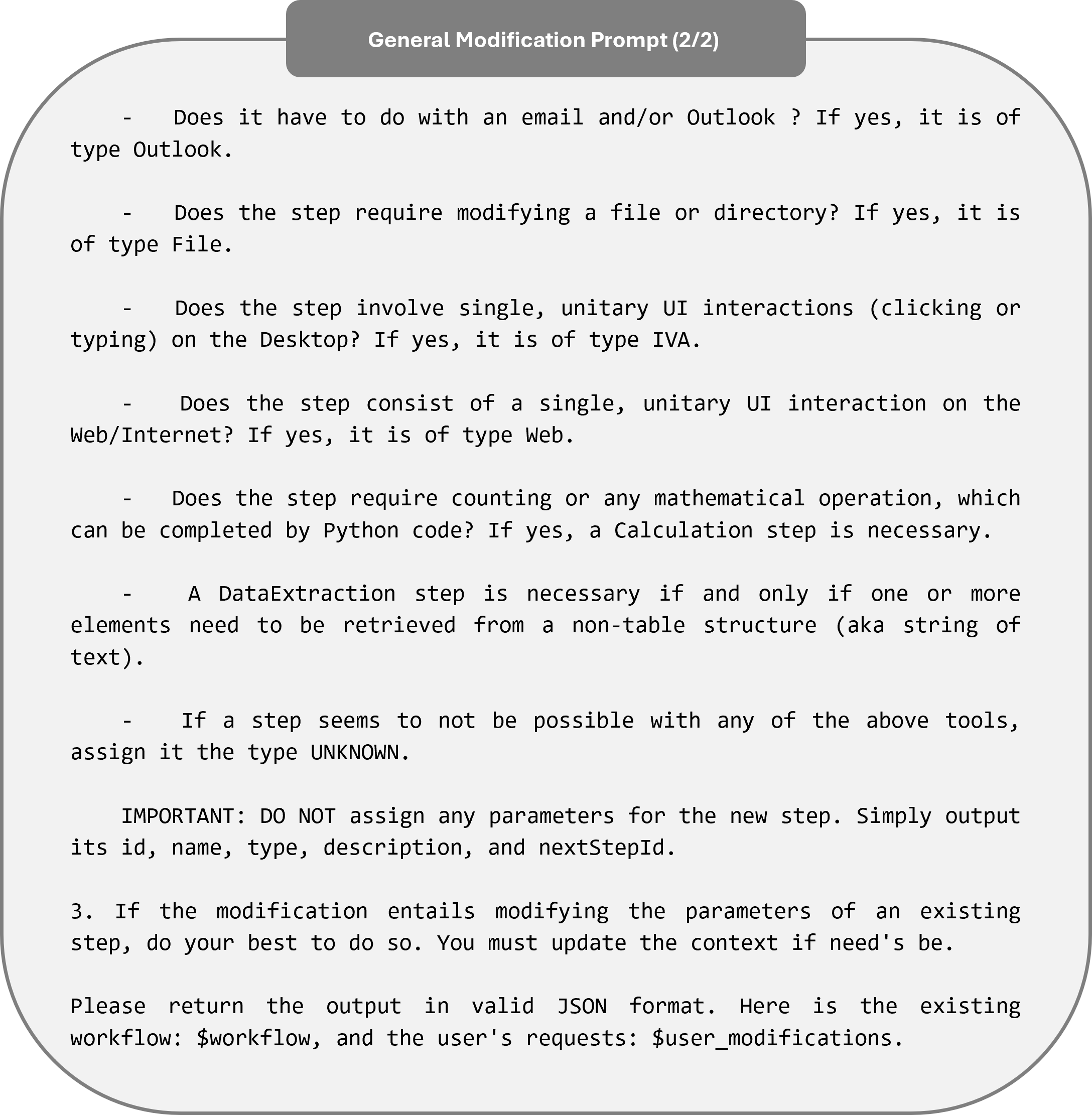}
    \caption{The \textit{Workflow Modification Prompt} (Part 2 of 2).}
    \label{fig:workflow-mod-prompt-2}
\end{figure}

    In this appendix, we describe the prompts used across the seven distinct layers of Text2Workflow: User Request Ambiguity Screening (\Cref{fig:logic-screening-prompt}), Building the Workflow Skeleton (\Cref{fig:gen-process-prompt,fig:master-prompt-1,fig:master-prompt-2,fig:master-prompt-3,fig:master-prompt-4,fig:master-prompt-5}), User Feedback Loop (\Cref{fig:brkdwn-summary}), Workflow Details (\Cref{fig:api-function-prompt,fig:writein-params,fig:click-select-params,fig:calculation-params-1,fig:calculation-params-2,fig:data-extract-params-1,fig:data-extract-params-2,fig:loop-params-1,fig:loop-params-2,fig:trycatch-desc}), Special Case – Further Workflow Details (\Cref{fig:api-params-1,fig:api-params-2,fig:tryblock-params,fig:throw-except-params}), Verifying Missing Parameters (\Cref{fig:questions-prompt}), and Final Modifications ( \Cref{fig:workflow-mod-prompt-1,fig:workflow-mod-prompt-2}).

\section{Baseline Prompt}
\label{app:baseline-prompt}

\begin{figure}[H]
    \centering
    \includegraphics[width=0.8\textwidth, height=0.8\textheight, keepaspectratio]{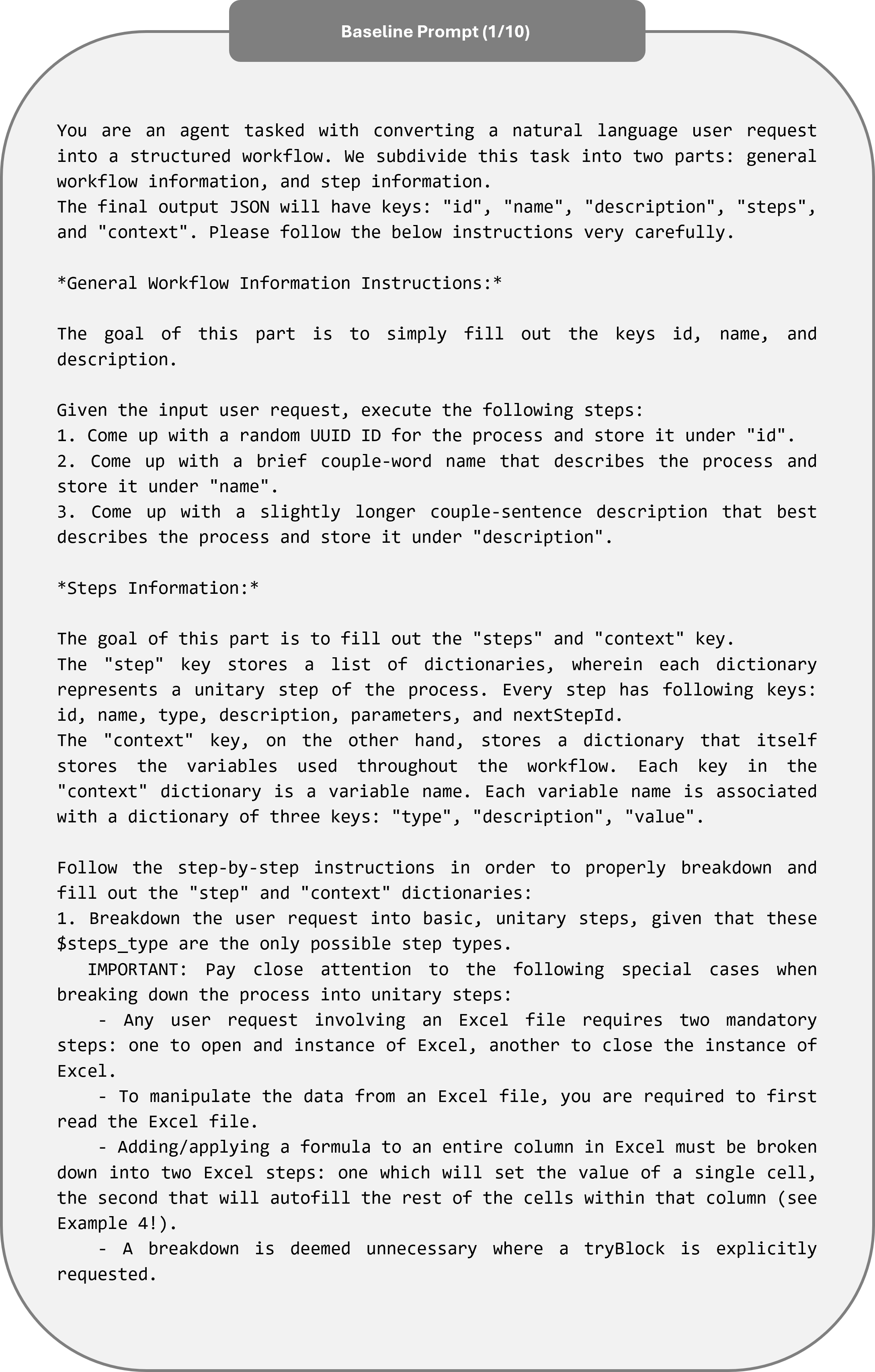}
    \caption{The \textit{Baseline Prompt} (Part 1 of 10).}
    \label{fig:baseline-prompt-1}
\end{figure}

\begin{figure}[H]
    \centering
    \includegraphics[width=0.9\textwidth, height=0.9\textheight, keepaspectratio]{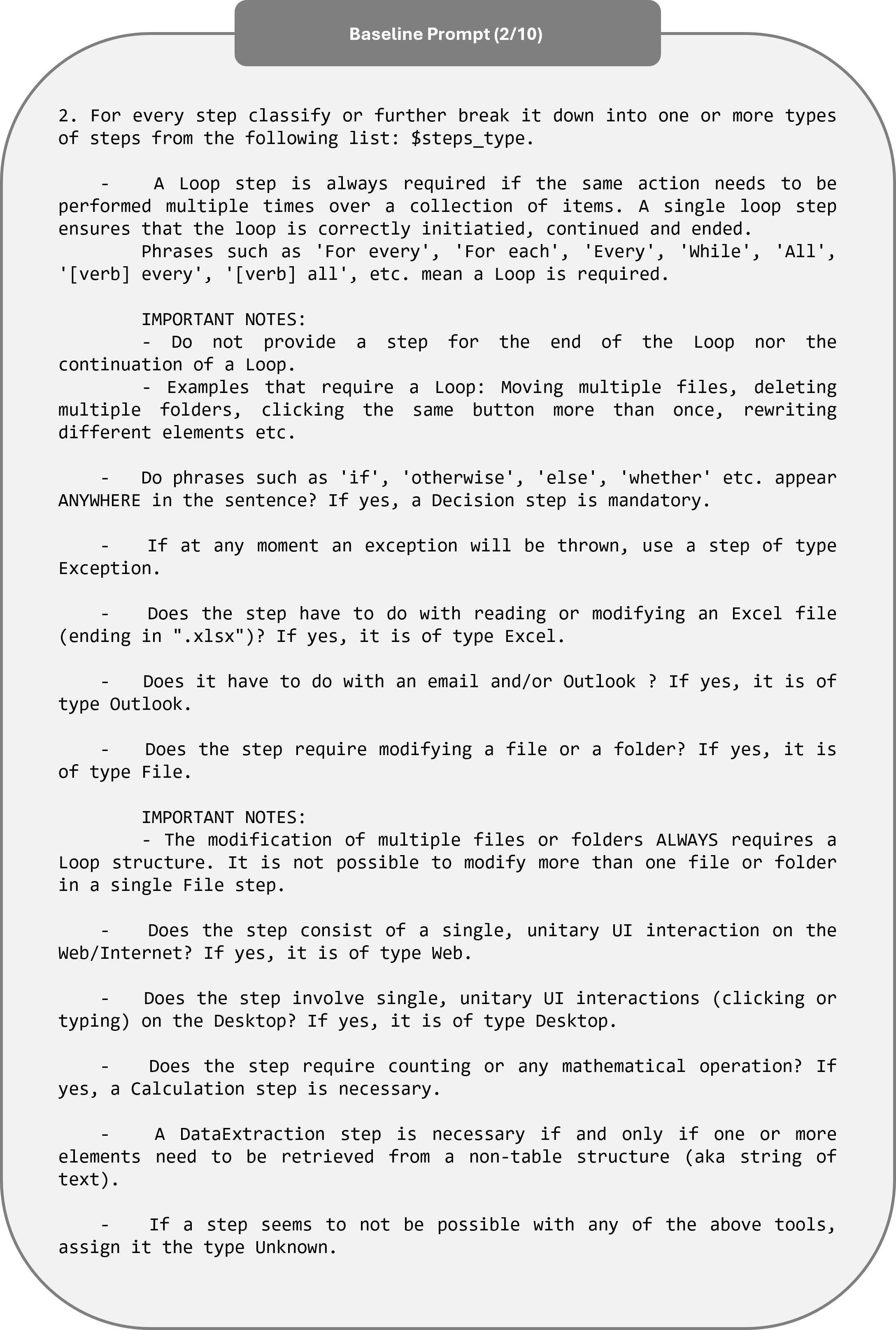}
    \caption{The \textit{Baseline Prompt} (Part 2 of 10).}
    \label{fig:baseline-prompt-2}
\end{figure}

\begin{figure}[H]
    \centering
    \includegraphics[width=\textwidth, height=\textheight, keepaspectratio]{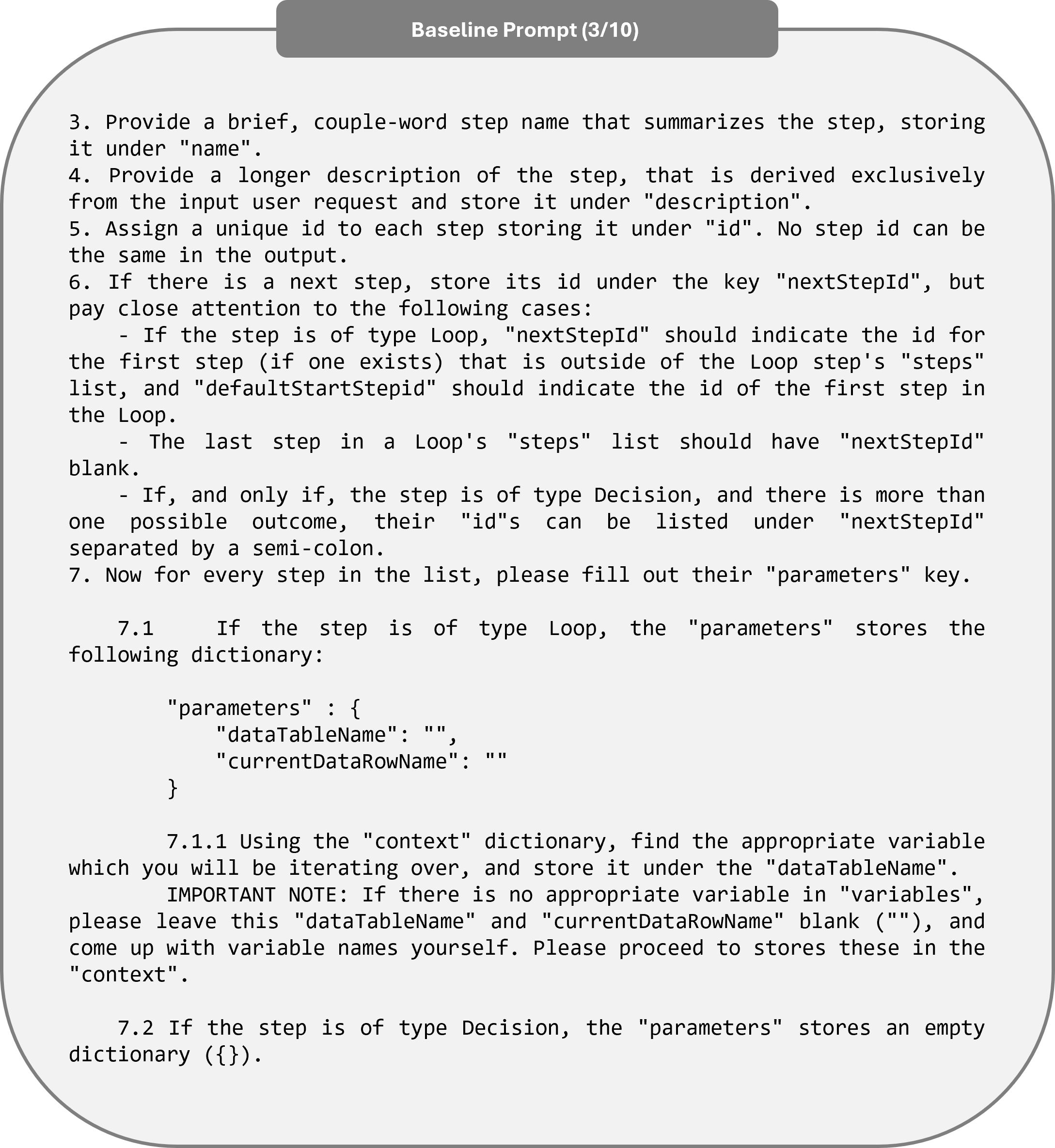}
    \caption{The \textit{Baseline Prompt} (Part 3 of 10).}
    \label{fig:baseline-prompt-3}
\end{figure}

\begin{figure}[H]
    \centering
    \includegraphics[width=\textwidth, height=\textheight, keepaspectratio]{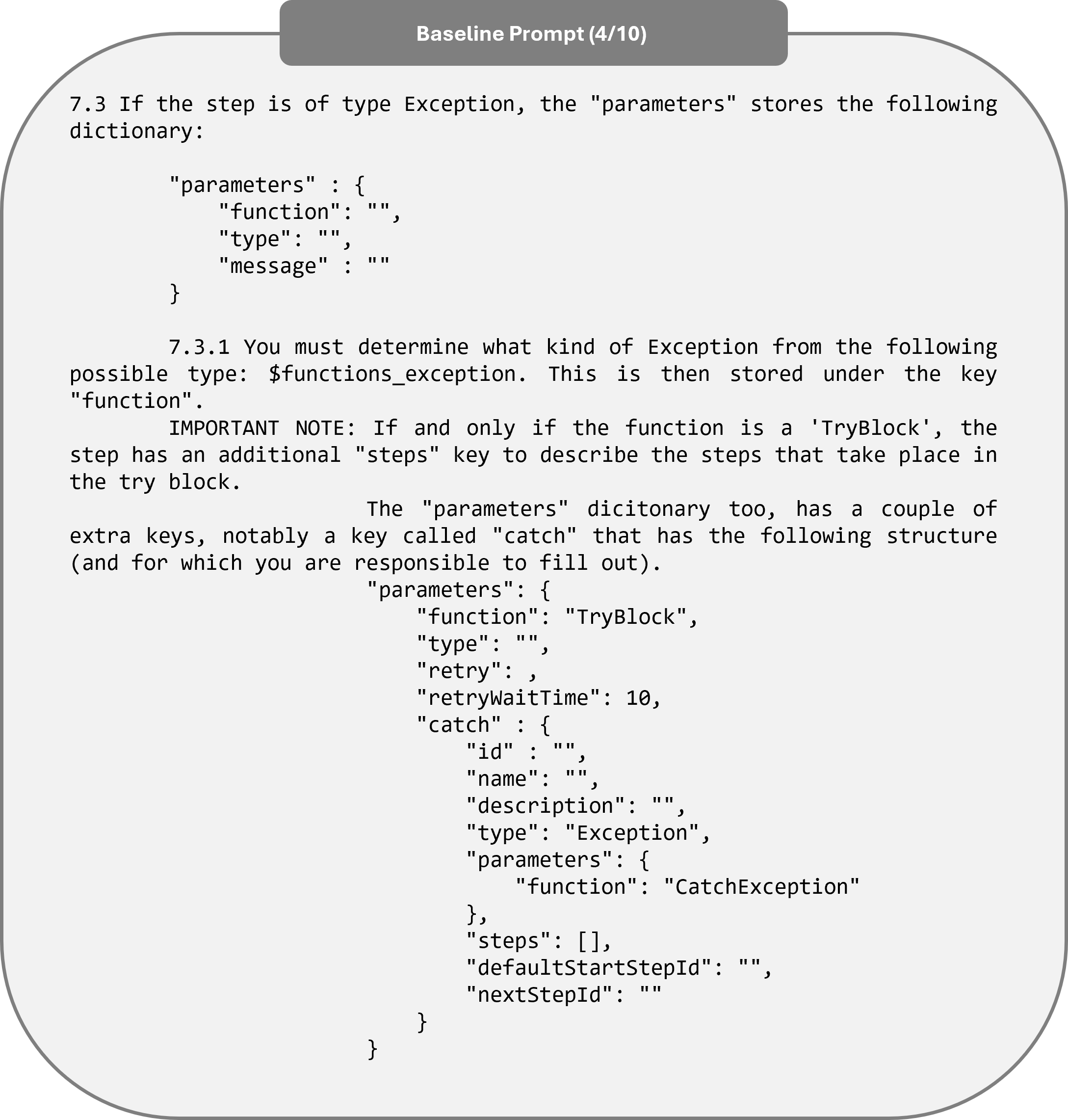}
    \caption{The \textit{Baseline Prompt} (Part 4 of 10).}
    \label{fig:baseline-prompt-4}
\end{figure}

\begin{figure}[H]
    \centering
    \includegraphics[width=\textwidth, height=\textheight, keepaspectratio]{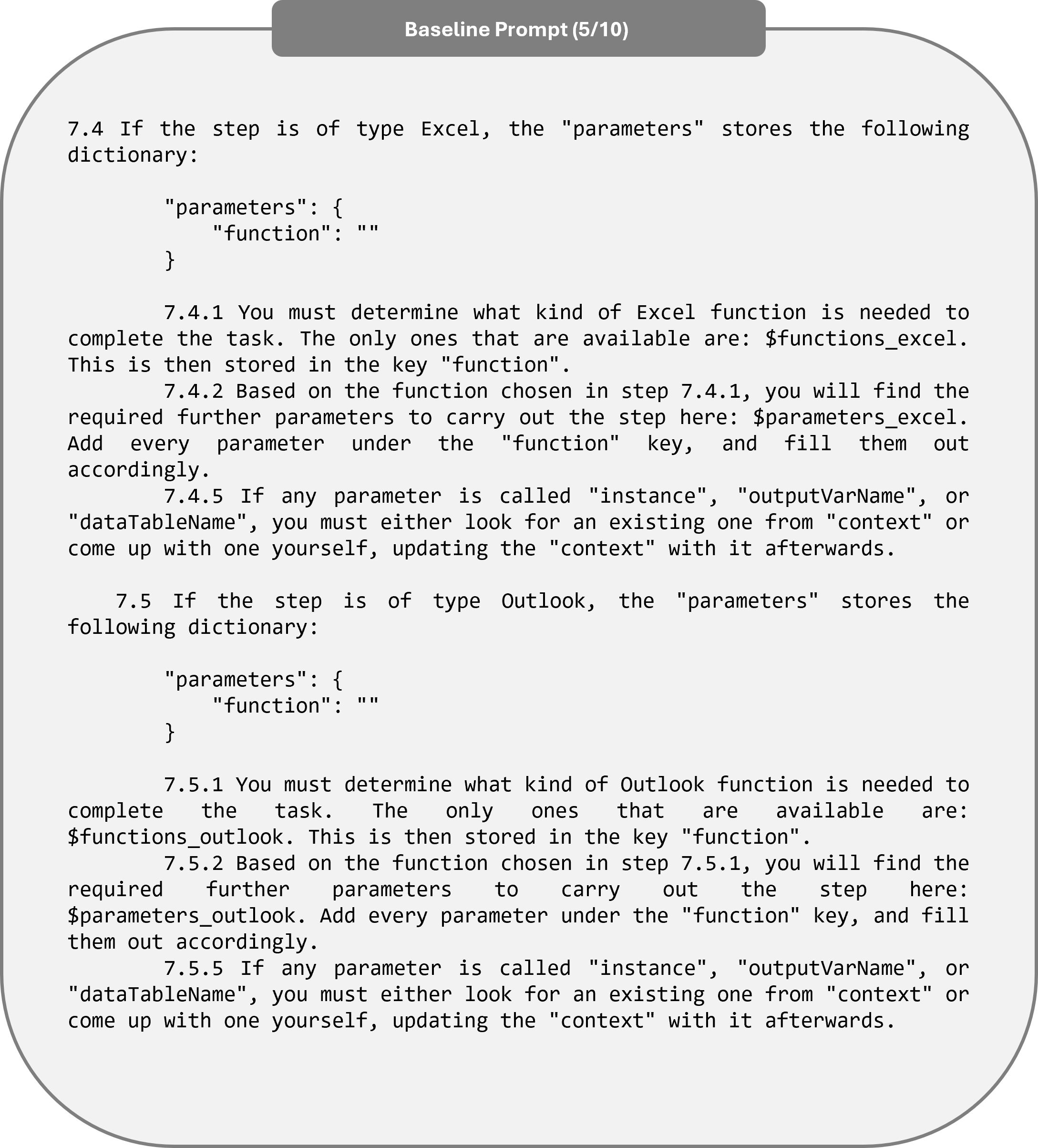}
    \caption{The \textit{Baseline Prompt} (Part 5 of 10).}
    \label{fig:baseline-prompt-5}
\end{figure}

\begin{figure}[H]
    \centering
    \includegraphics[width=\textwidth, height=\textheight, keepaspectratio]{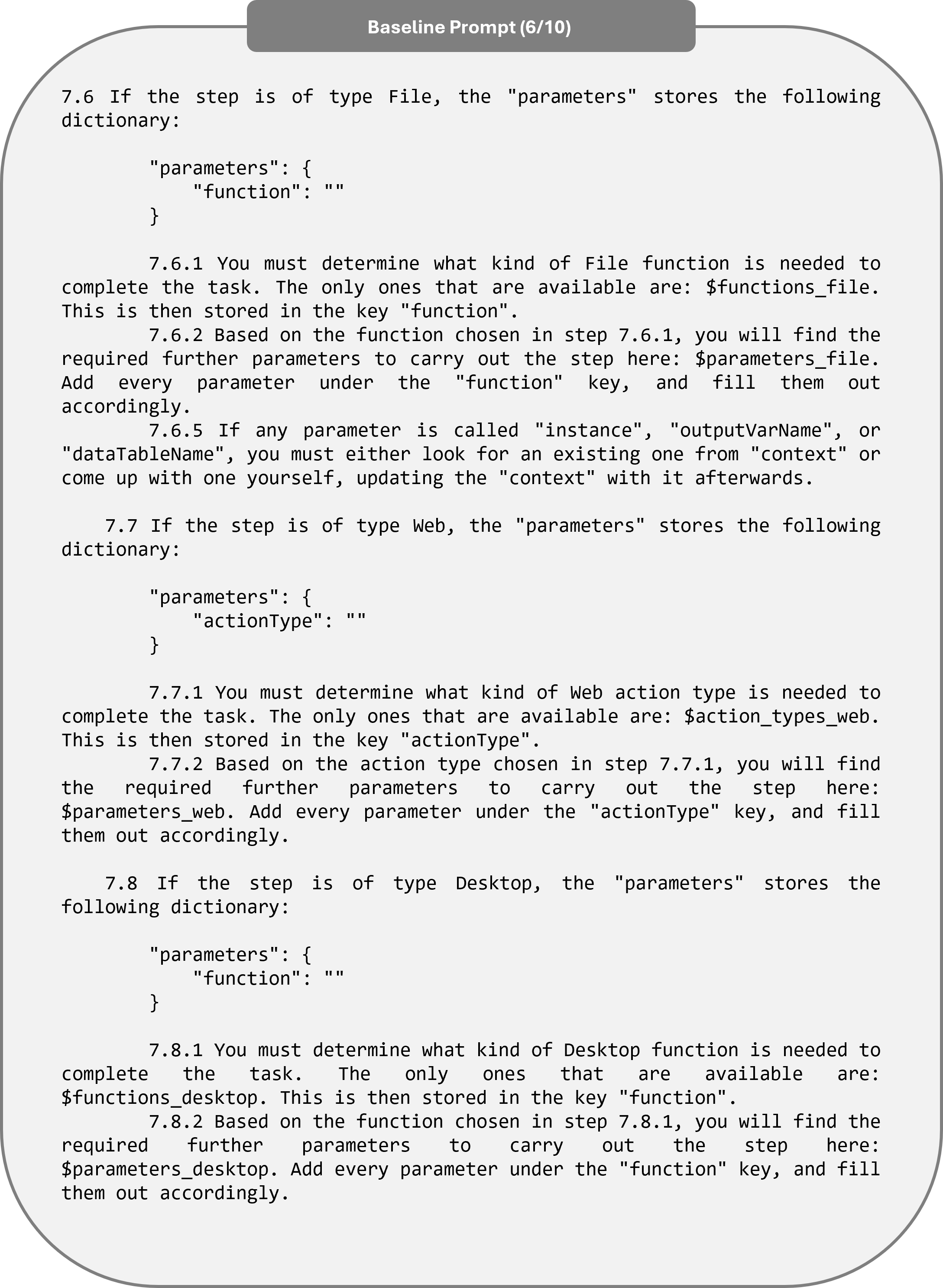}
    \caption{The \textit{Baseline Prompt} (Part 6 of 10).}
    \label{fig:baseline-prompt-6}
\end{figure}

\begin{figure}[H]
    \centering
    \includegraphics[width=\textwidth, height=\textheight, keepaspectratio]{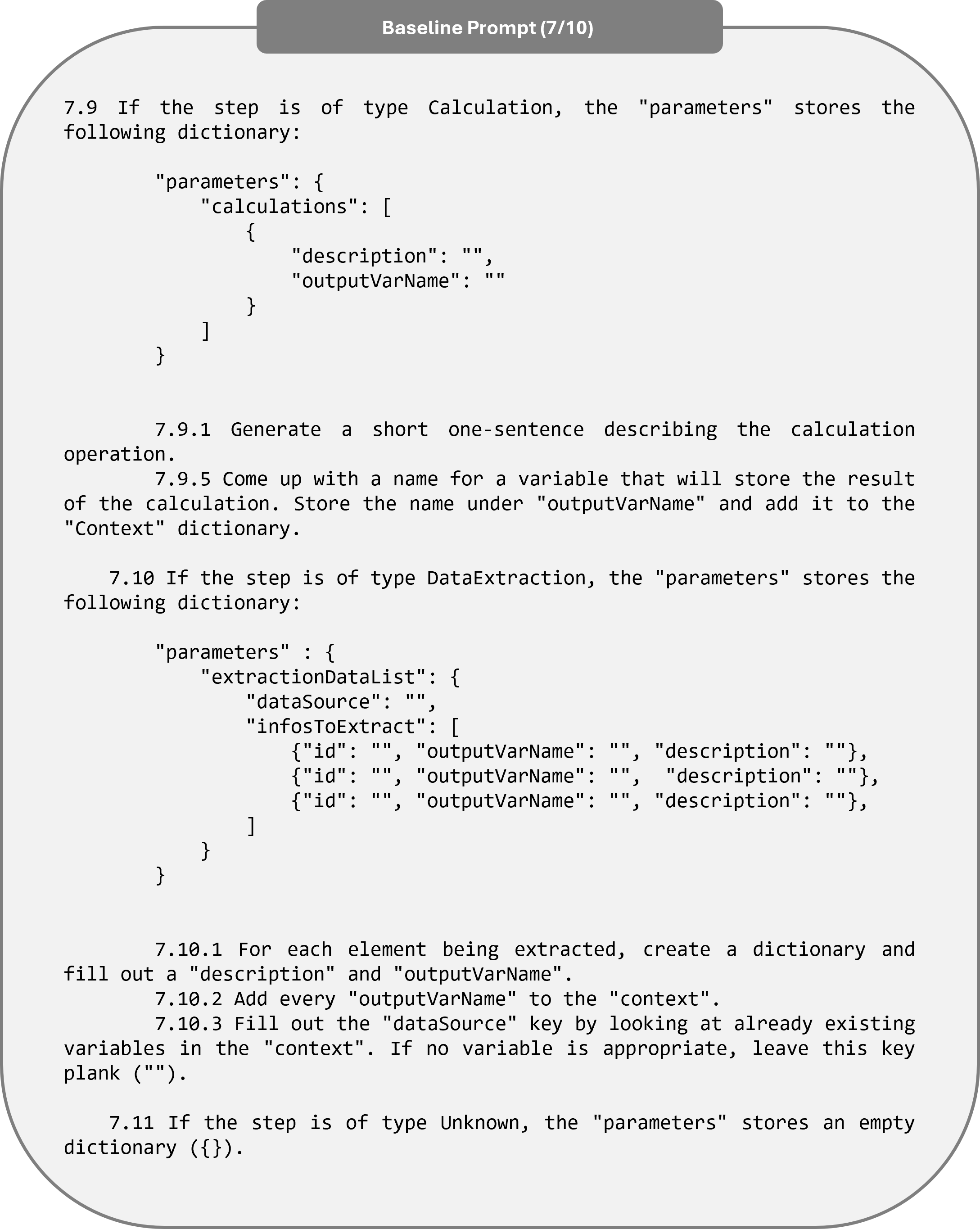}
    \caption{The \textit{Baseline Prompt} (Part 7 of 10).}
    \label{fig:baseline-prompt-7}
\end{figure}

\begin{figure}[H]
    \centering
    \includegraphics[width=\textwidth, height=\textheight, keepaspectratio]{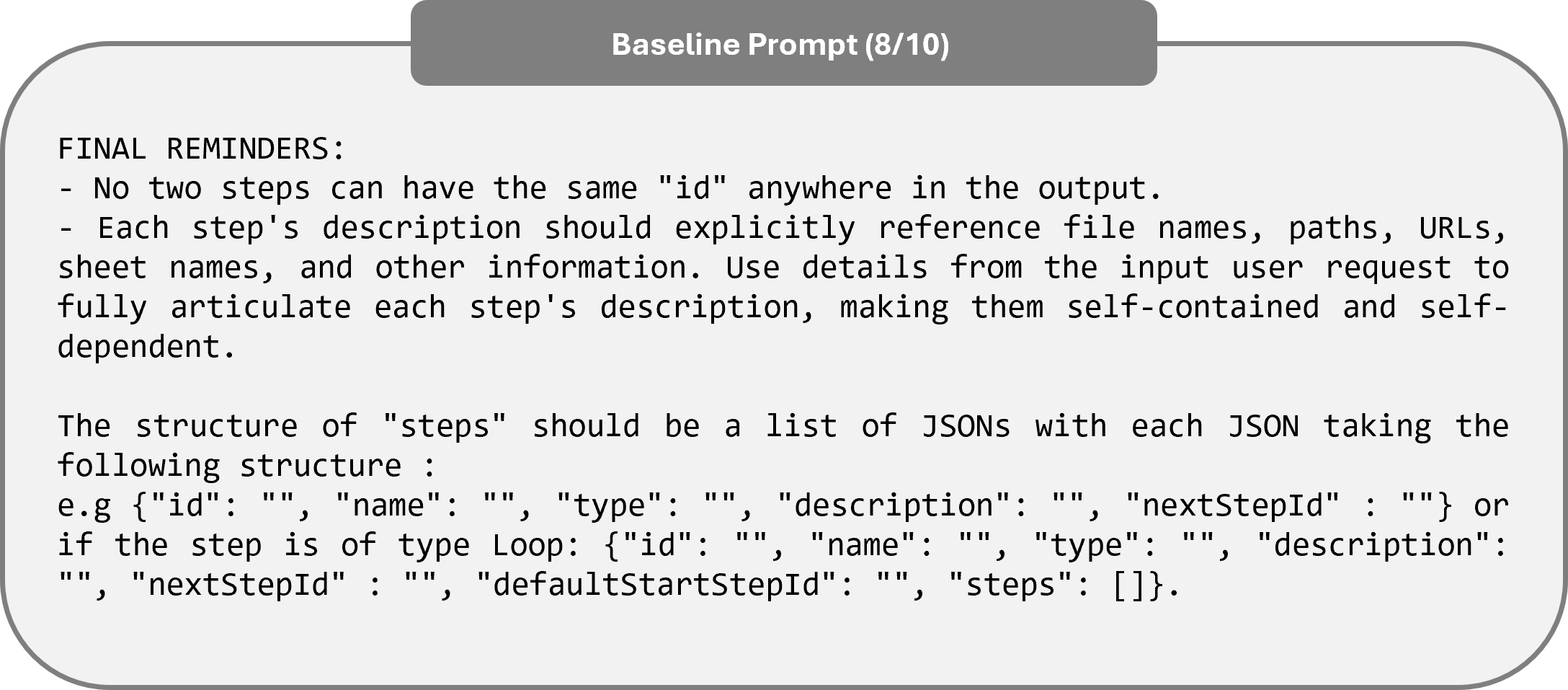}
    \caption{The \textit{Baseline Prompt} (Part 8 of 10).}
    \label{fig:baseline-prompt-8}
\end{figure}

\begin{figure}[H]
    \centering
    \includegraphics[width=0.95\textwidth, height=0.95\textheight, keepaspectratio]{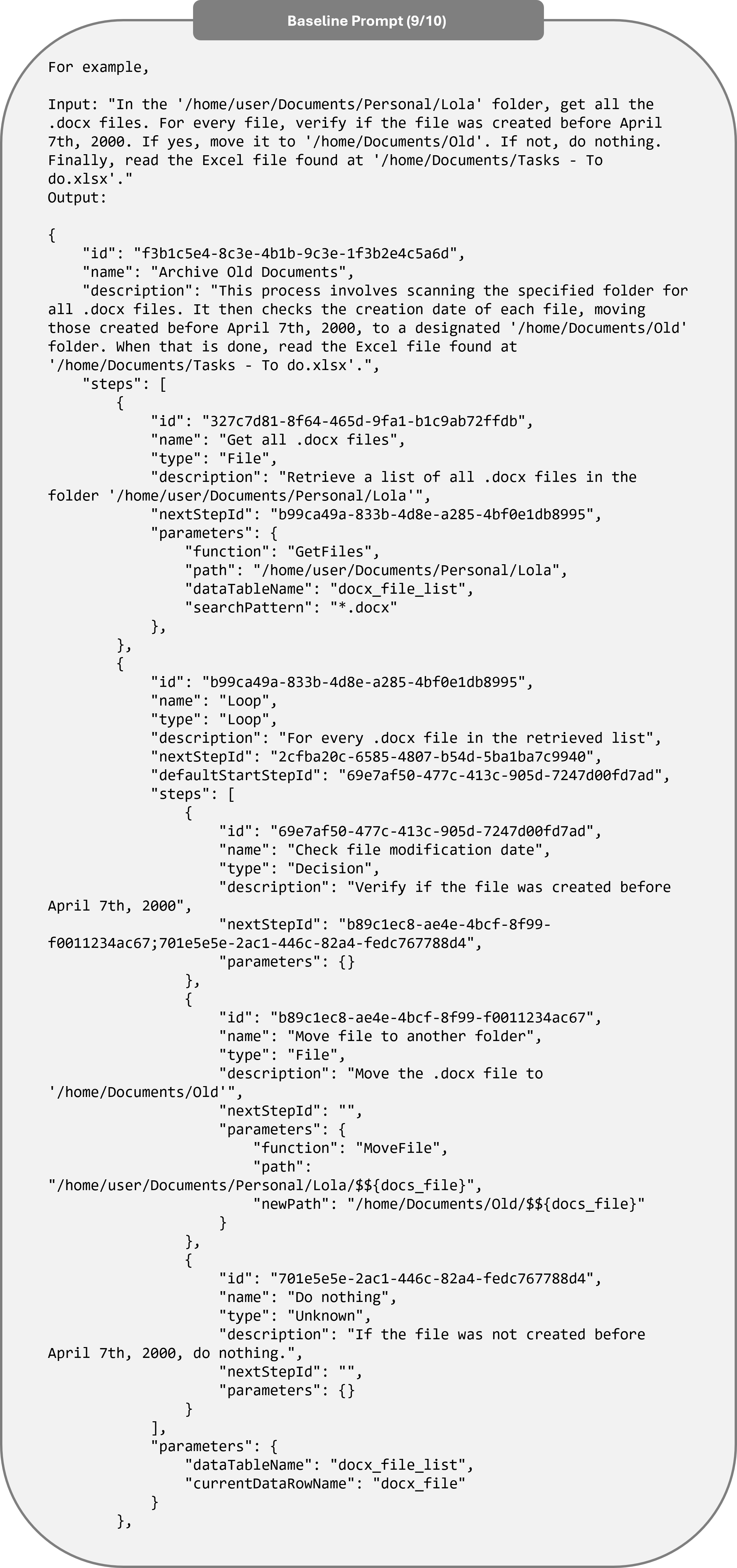}
    \caption{The \textit{Baseline Prompt} (Part 9 of 10).}
    \label{fig:baseline-prompt-9}
\end{figure}

\begin{figure}[H]
    \centering
    \includegraphics[width=0.95\textwidth, height=0.95\textheight, keepaspectratio]{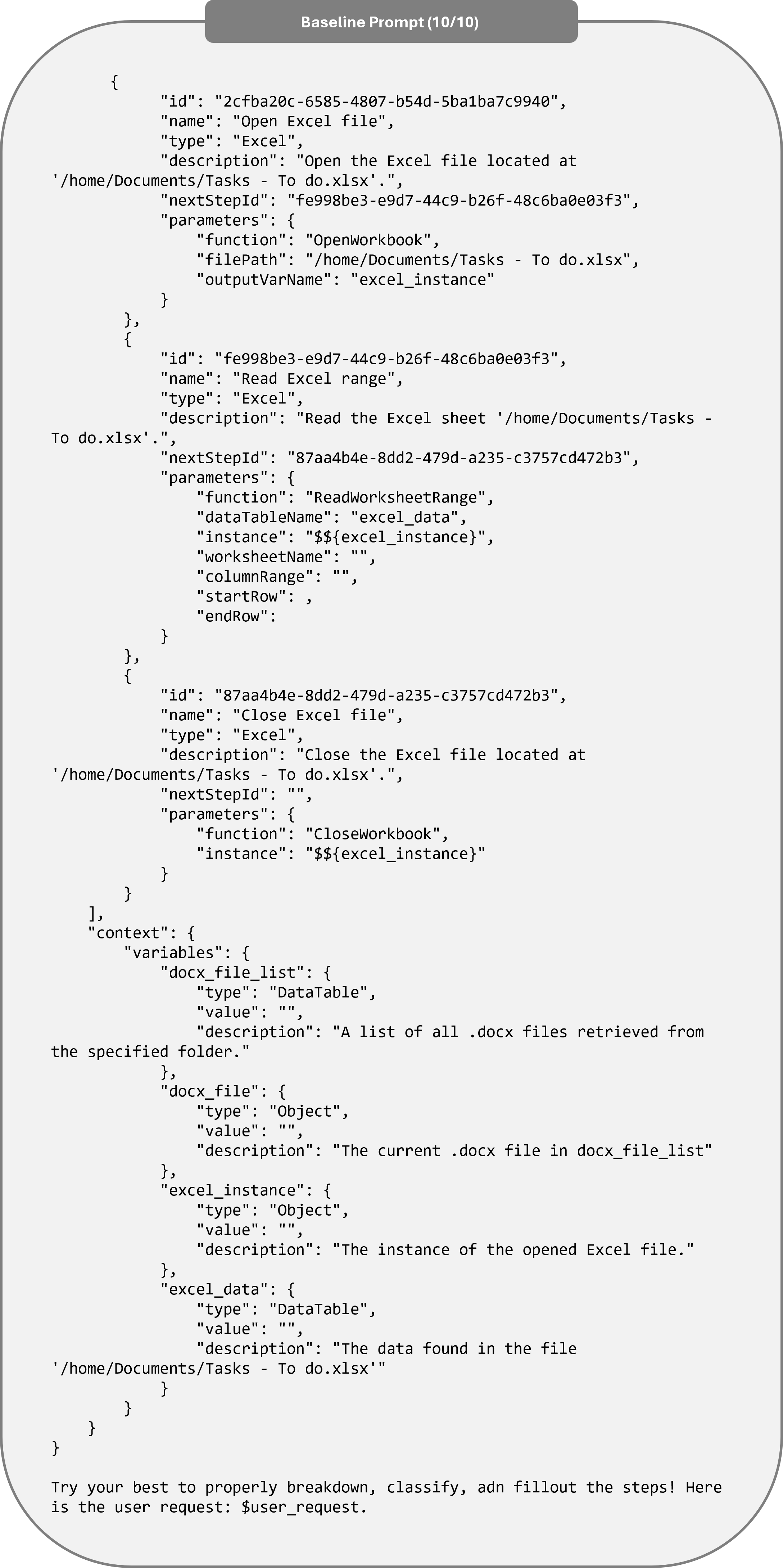}
    \caption{The \textit{Baseline Prompt} (Part 10 of 10).}
    \label{fig:baseline-prompt-10}
\end{figure}

\section{Example of JSON Workflow Generation Failures}
\label{app:example_failures}

Text2Workflow tends to reproduce the following common errors in its final output JSON: mistaking an evident step type for type Unknown, misusing the parameters from a Loop type step, incorrectly assigning nextStepIds, particularly with respect to Decision and Loop type steps, and finally, misidentification of Exception type steps, especially those with TryBlocks.

\begin{figure}[H]
    \centering
    \includegraphics[width=0.95\textwidth, height=0.95\textheight, keepaspectratio]{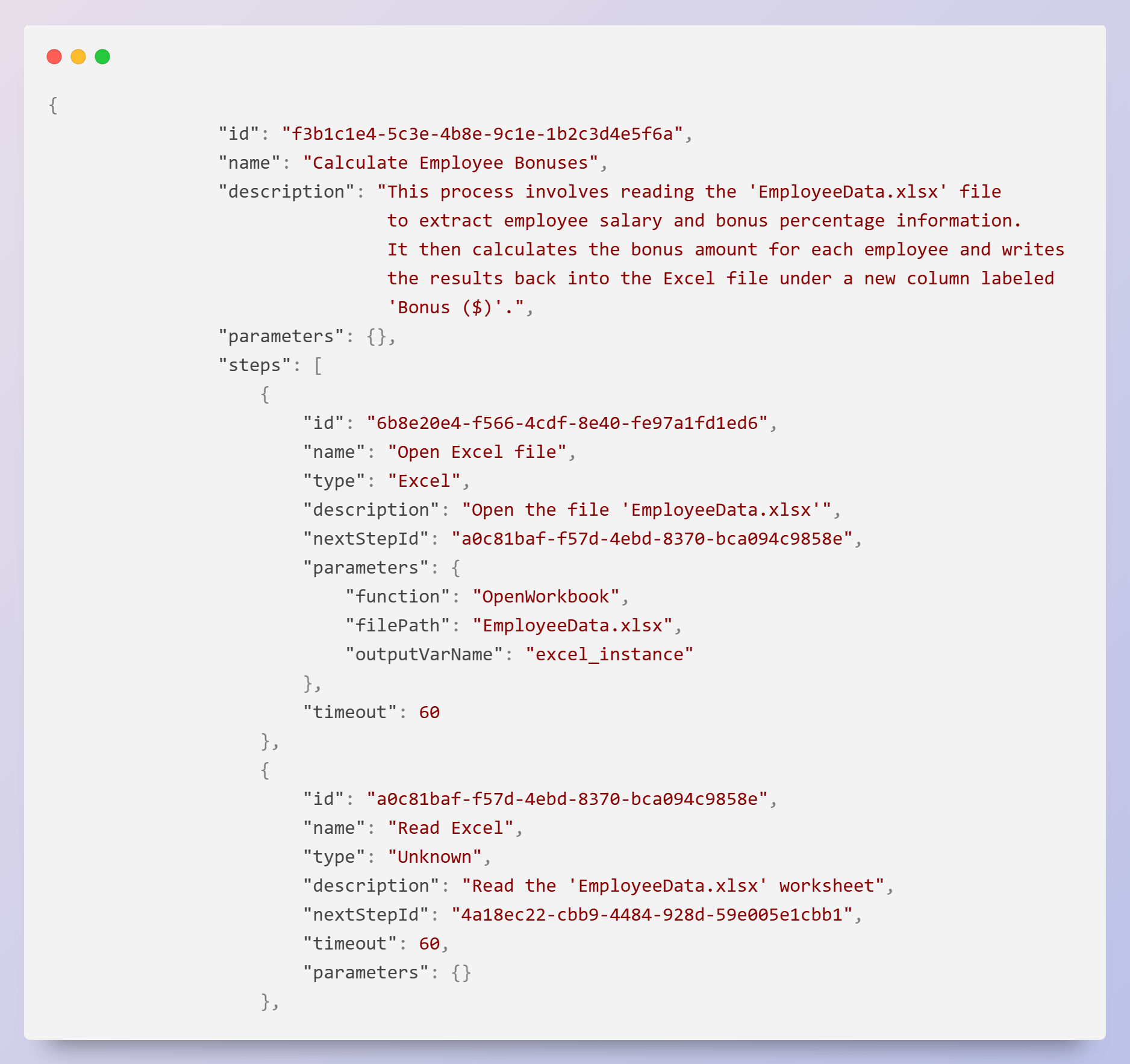}
    \label{fig:medium-example3-erroneous-1}
    \caption{Erroneous output from Text2Workflow for \protect\hyperlink{medium-example3}{Example 1} (1/3).}
\end{figure}

\begin{figure}[H]
    \centering
    \includegraphics[width=0.95\textwidth, height=0.95\textheight, keepaspectratio]{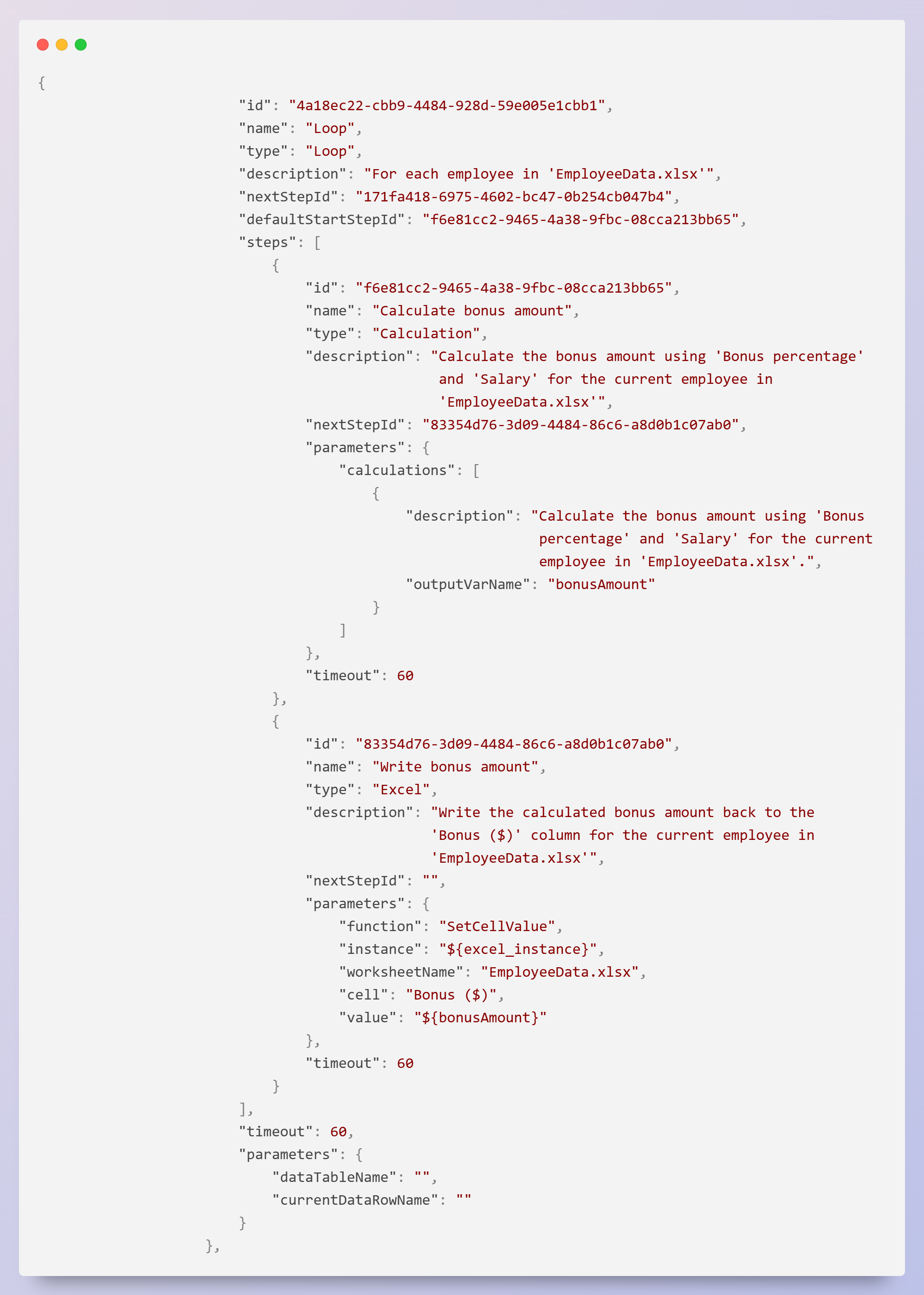}
    \caption{Erroneous output from Text2Workflow  for \protect\hyperlink{medium-example3}{Example 1} (2/3).}
    \label{fig:medium-example3-erroneous-2}
\end{figure}

\begin{figure}[H]
    \centering
    \includegraphics[width=0.95\textwidth, height=0.95\textheight, keepaspectratio]{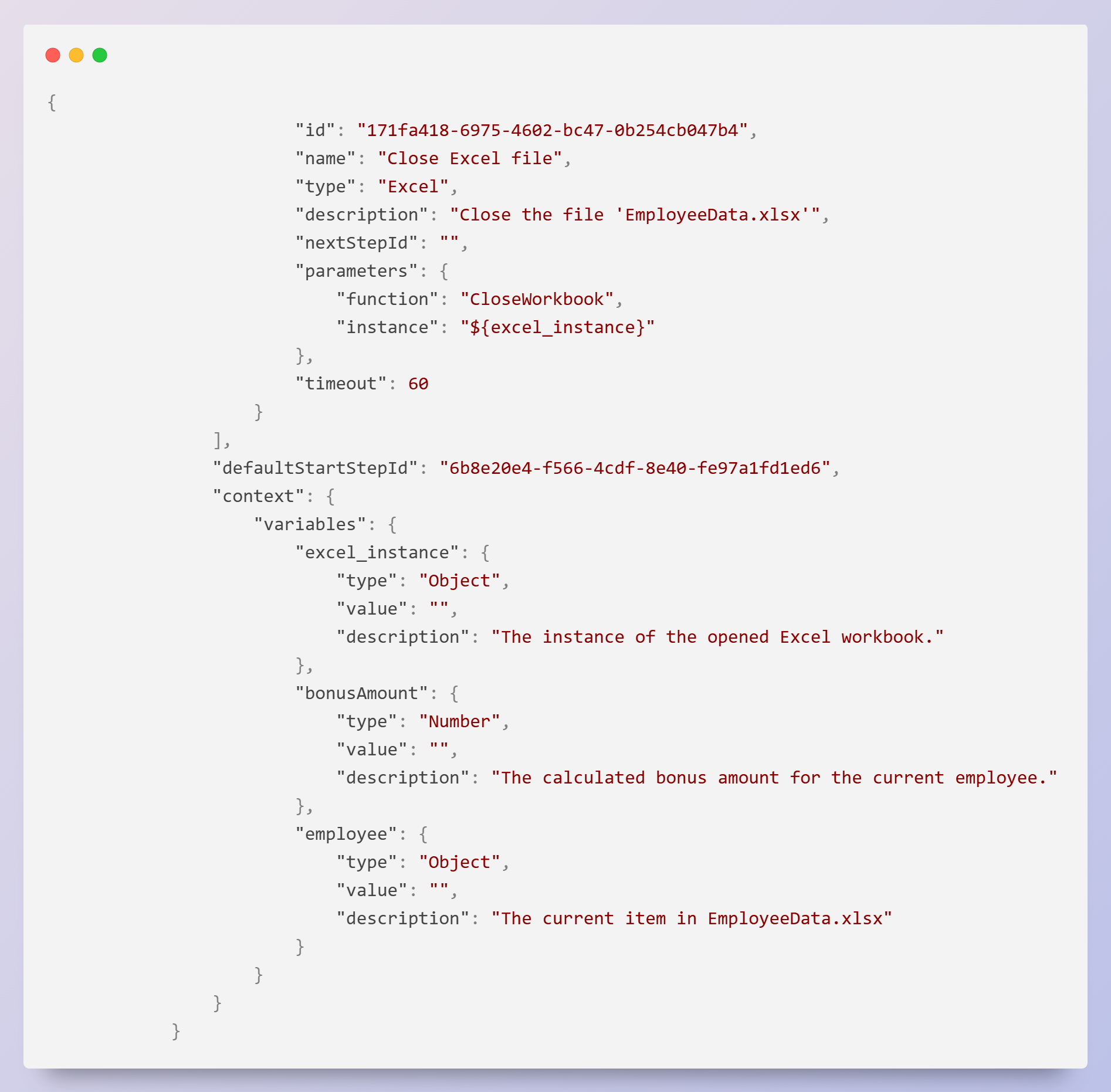}
    \caption{Erroneous output from Text2Workflow  for \protect\hyperlink{medium-example3}{Example 1} (3/3).}
    \label{fig:medium-example3-erroneous-3}
\end{figure}

Though there are several different instances that exemplify the above errors, in the interest of saving space, we provide a single example. The following \Cref{fig:medium-example3-erroneous-1,fig:medium-example3-erroneous-2,fig:medium-example3-erroneous-3} demonstrate an output JSON with the first two common mistakes - marking a step Unknown, and not filling out nor using the parameters from a Loop step. The original input user request can be seen below.

\hypertarget{medium-example3}{
\begin{quotation}
\textbf{Example 1 (Input)} : Read the 'EmployeeData.xlsx' file in Excel. For each employee, use the ‘Bonus percentage’ and ‘Salary’ information to calculate their bonus amount, and then write the bonus amount back to the Excel file in a column called ‘Bonus (\$)’.
\end{quotation}}


We observe that Text2Workflow's inability to identify the second step in the workflow, with name 'Read Excel', as an Excel step type with function 'ReadWorkSheetRange' creates a chain reaction that eventually affects its ability to identify the proper parameters for the Loop step, rendering the final workflow nonfunctional. 

\begin{figure}[H]
    \centering
    \includegraphics[width=0.9\textwidth, height=0.9\textheight, keepaspectratio]{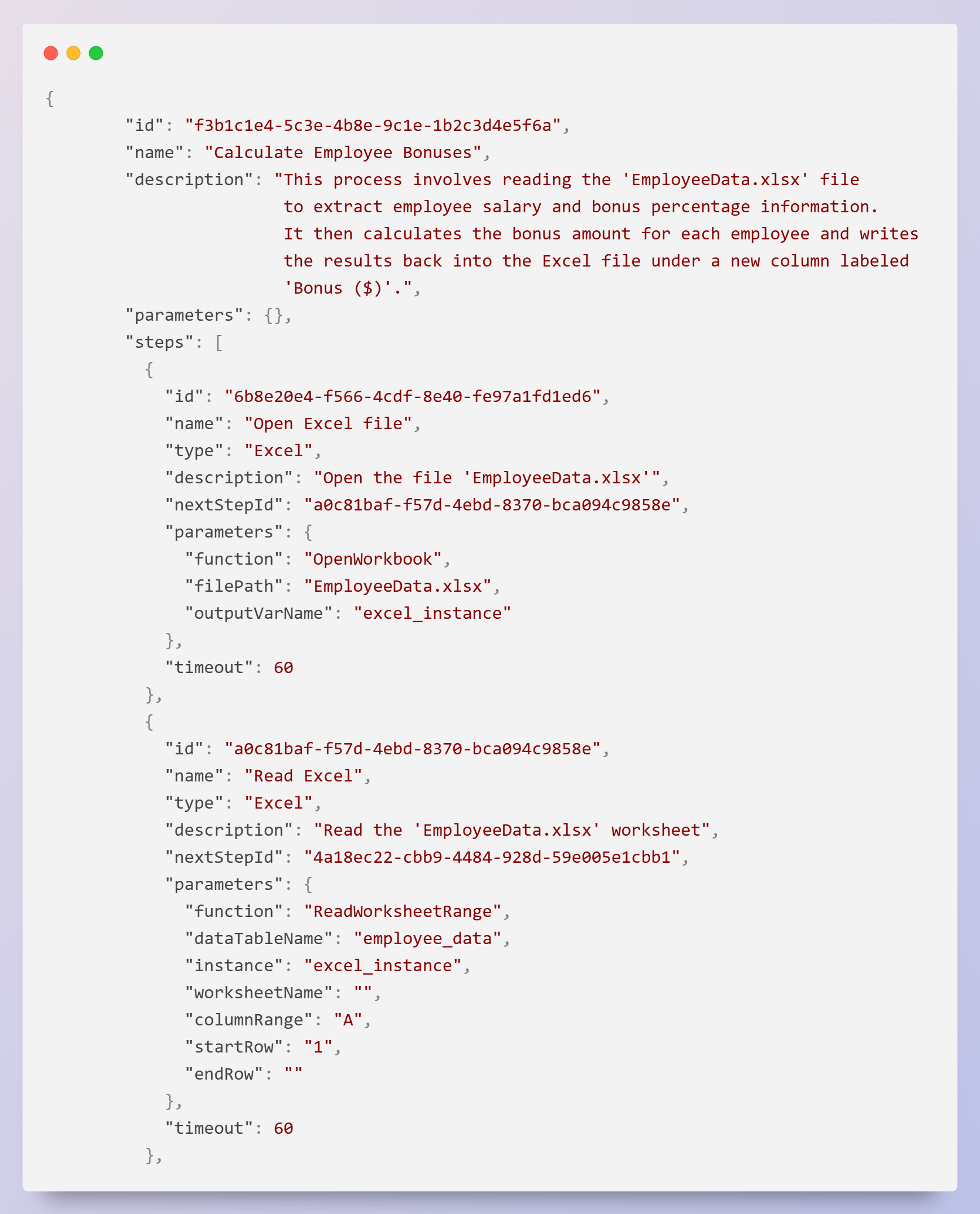}
    \caption{Expected output from Text2Workflow for \protect\hyperlink{medium-example3}{Example 1} (1/3).}
    \label{fig:medium-example3-perfect-1}
\end{figure}

\begin{figure}[H]
    \centering
    \includegraphics[width=0.95\textwidth, height=0.95\textheight, keepaspectratio]{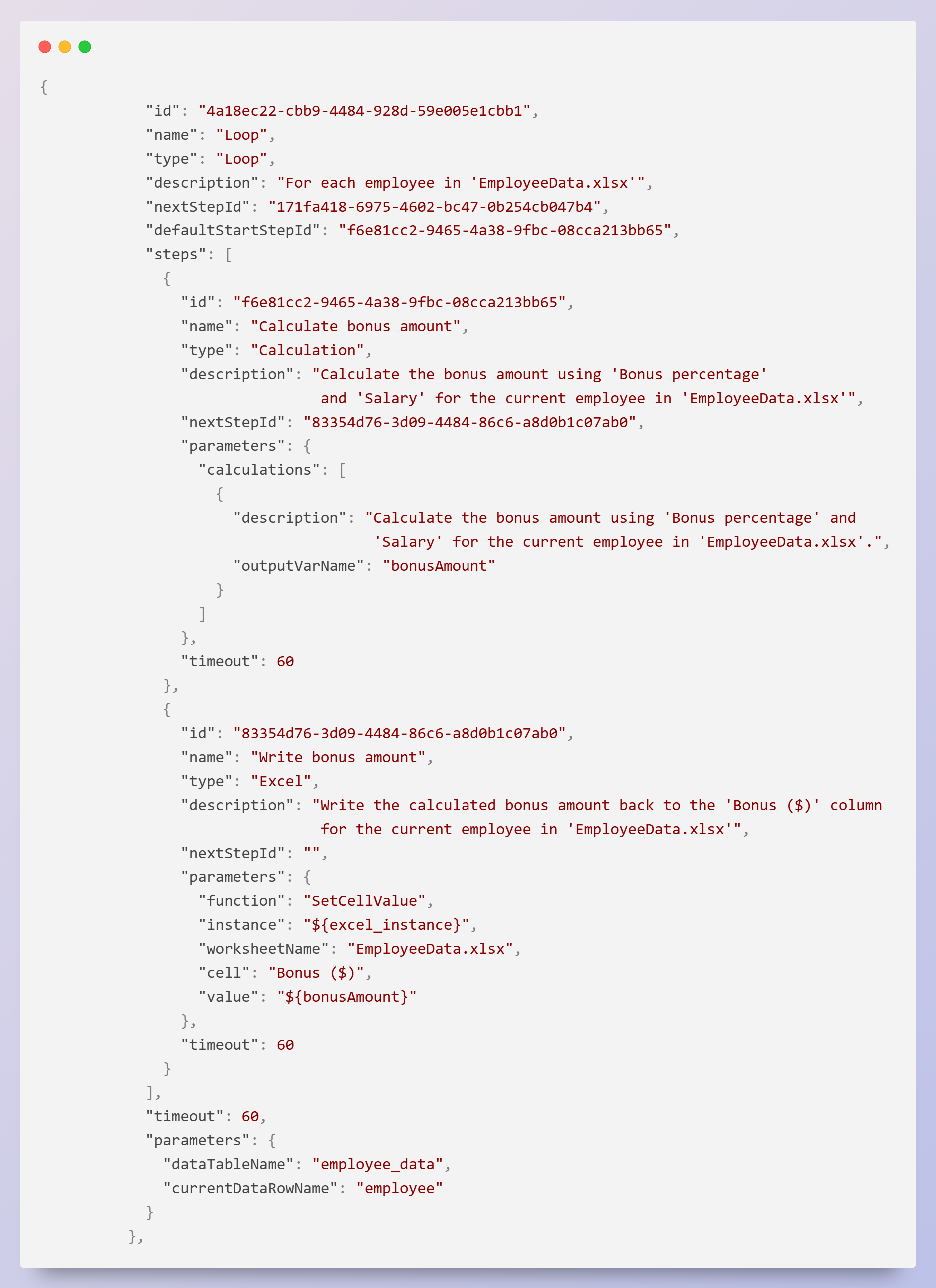}
    \caption{Expected output from Text2Workflow for \protect\hyperlink{medium-example3}{Example 1} (2/3).}
    \label{fig:medium-example3-perfect-2}
\end{figure}

\begin{figure}[H]
    \centering
    \includegraphics[width=0.95\textwidth, height=0.95\textheight, keepaspectratio]{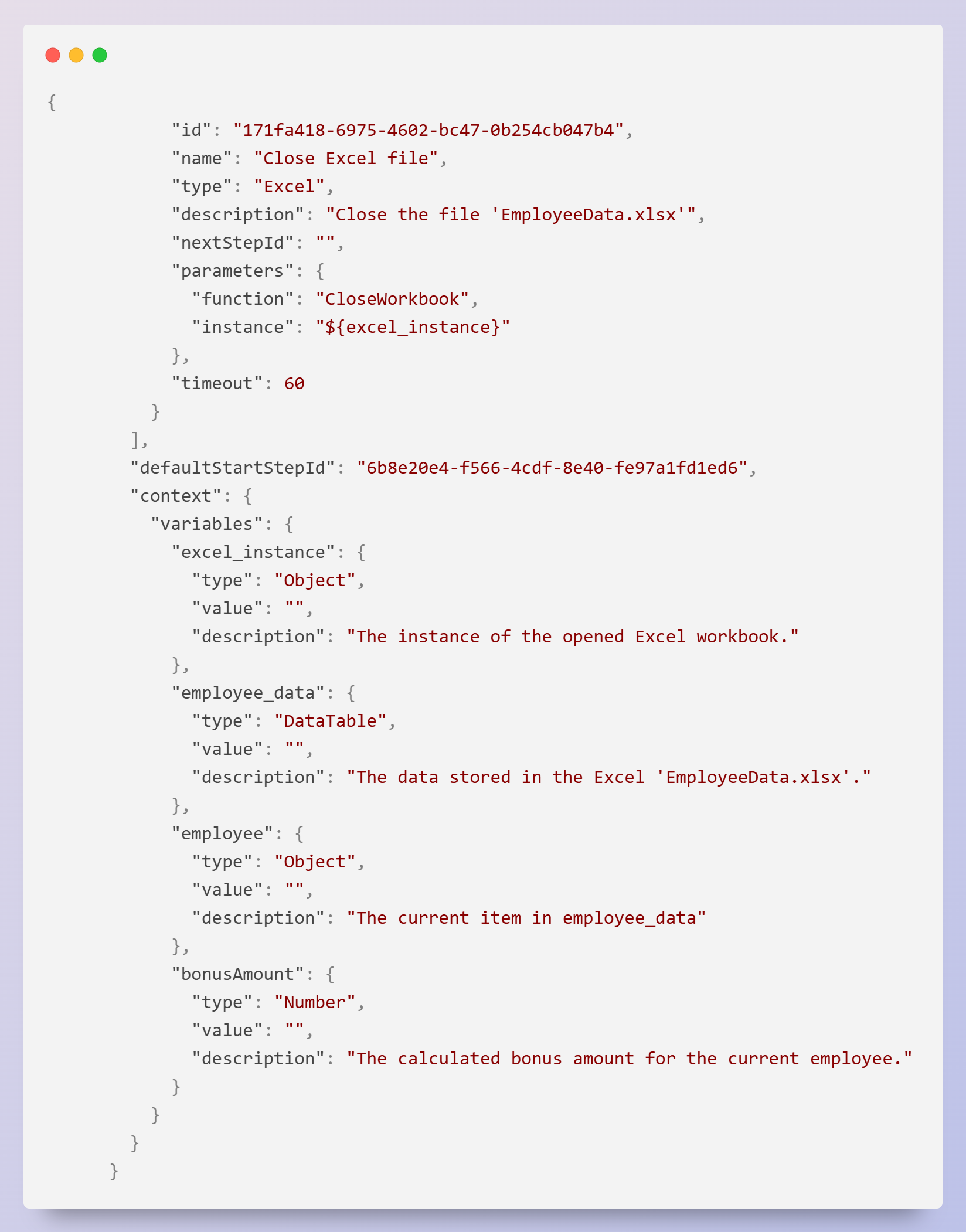}
    \caption{Expected output from Text2Workflow for \protect\hyperlink{medium-example3}{Example 1} (3/3).}
    \label{fig:medium-example3-perfect-3}
\end{figure}

\Cref{fig:medium-example3-perfect-1,fig:medium-example3-perfect-2,fig:medium-example3-perfect-3} below demonstrate the expected output for the aforementioned \hyperlink{medium-example3}{Example 1}. One can observe that the correct classification of the 'Read Excel' step leads to proper initialization of the Loop parameters, and a proper and complete context dictionary.

\end{document}